\documentclass[11pt]{article}

\usepackage[margin=1in]{geometry}
\usepackage{amsmath, amssymb}
\usepackage{graphicx}
\usepackage{xcolor} 
\usepackage{natbib}
\usepackage{soul}
\usepackage{comment}
\usepackage{xcolor}
\usepackage{booktabs}
\usepackage{amsfonts}
\usepackage{algorithmic}
\usepackage{array}
\usepackage{stfloats}
\usepackage{tabularx}
\usepackage{float}
\usepackage{subcaption}
\usepackage{setspace}
\usepackage{threeparttable}
\usepackage{rotating}
\usepackage{booktabs}  % For better table formatting
\usepackage{multirow}  % Allows multi-row cells
\usepackage{array}     % Allows better column formatting
\usepackage{pdflscape}
\usepackage{pifont}
\usepackage{caption}
\usepackage{longtable}
\usepackage[utf8]{inputenc}
\usepackage{lmodern}
\usepackage{hyperref}

\title{Reinforcement Learning for Multi-Objective Multi-Echelon Supply Chain Optimisation}

\author{
  Rifny Rachman\thanks{Corresponding author. Email: \texttt{rifny.rachman@postgrad.manchester.ac.uk}}\\
  The University of Manchester, United Kingdom
  \and
  Josh Tingey\\
  Peak AI, Ltd, United Kingdom\\
  \texttt{josh.tingey@peak.ai}
  \and
  Richard Allmendinger\\
  The University of Manchester, United Kingdom\\
  \texttt{richard.allmendinger@manchester.ac.uk}
  \and
  Pradyumn Shukla\\
  The University of Manchester, United Kingdom\\
  \texttt{pradyumn.shukla@manchester.ac.uk}
  \and
  Wei Pan\\
  The University of Manchester, United Kingdom\\
  \texttt{wei.pan@manchester.ac.uk}
}

\date{} % leave empty for arXiv

\begin{document}
\maketitle

\begin{abstract}
\noindent
This study develops a generalised multi-objective, multi-echelon supply chain optimisation model with non-stationary markets based on a Markov decision process, incorporating economic, environmental, and social considerations. The model is evaluated using a multi-objective reinforcement learning (RL) method, benchmarked against an originally single-objective RL algorithm modified with weighted sum using predefined weights, and a multi-objective evolutionary algorithm (MOEA)-based approach. We conduct experiments on varying network complexities, mimicking typical real-world challenges using a customisable simulator. The model determines production and delivery quantities across supply chain routes to achieve near-optimal trade-offs between competing objectives, approximating Pareto front sets. The results demonstrate that the primary approach provides the most balanced trade-off between optimality, diversity, and density, further enhanced with a shared experience buffer that allows knowledge transfer among policies. In complex settings, it achieves up to 75\% higher hypervolume than the MOEA-based method and generates solutions that are approximately eleven times denser, signifying better robustness, than those produced by the modified single-objective RL method. Moreover, it ensures stable production and inventory levels while minimising demand loss.
\end{abstract}


%\item 
%\item 
%\item 

% Keywords
\vspace{1em}

%\maketitle
\onehalfspacing
% Main text
\section{Introduction}\label{introduction}
In real-world decision-making scenarios, particularly in the context of optimisation problems, it is common to encounter multiple, often conflicting, objectives. In supply chain (SC) optimisation, prevalent objectives include profit maximisation or cost minimisation, with a growing focus on sustainability issues that cover economic, environmental, and social dimensions~\citep{becerra_green_2021}. Although studies covering economic and environmental aspects have been prevalent, social aspects still lack literature~\citep{asgharizadeh_sustainable_2019,jayarathna_multi-objective_2021}. In addition, SC network design (SCND) and inventory management problem (IMP) are particularly critical, comprising around 80\% of the total SC cost~\citep{watson_supply_2013}. SCND refers to defining the location, capacities, and product flow~\citep{autry_definitive_2013,badri_two-stage_2017}, while IMP involves daily product movement and stock level to meet demand~\citep{watson_supply_2013}. Optimising SC can be challenging due to the intricacies and uncertainties within a high-dimensional combinatorial problem that originates from multiple echelons, facilities, and their interconnections~\citep{bozarth_impact_2009}.

Traditional research in multi-objective SC optimisation has predominantly employed classical and metaheuristic approaches~\citep{abdolazimi_robust_2020, darestani_robust_2019, tautenhain_efficient_2021, farrokhi-asl_developing_2020}. While effective in exploring trade-offs, these methods typically neglect the dynamic interactions between an agent’s actions and the resulting responses from the SC environment. As a result, they often lack feedback sensitivity and may struggle to adapt to real-time system states or evolving operational conditions when identifying trade-off solutions. To address this, recent studies have explored reinforcement learning (RL) methods based on the Markov decision process (MDP)~\citep{sutton_reinforcement_2015}, which explicitly incorporate agent-environment interactions. This approach observes states and rewards to determine subsequent actions based on learnt policies. It enables online learning, where the model continuously updates its policy based on real-time interactions with the environment during the SC optimisation process. Most of the existing literature in the SC domain focuses on single-objective RL applications~\citep{kemmer_reinforcement_2018, oroojlooyjadid_deep_2017, abu_zwaida_optimization_2021, ganesan_adaptive_2019, pontrandolfo_global_2002}. A few recent studies have begun to apply multi-objective reinforcement learning (MORL) to SC problems~\citep{shar_multi-objective_2023, zhao_research_2025, singh_multi-objective_2025}, yet these are often confined to simple or narrowly defined networks and typically omit social objectives. Building on initial insights from a case-specific study~\citep{singh_multi-objective_2025}, this paper presents a generalised and customisable MORL framework for SC optimisation, capable of handling diverse network configurations, non-stationary demand, and integrated economic, environmental, and social trade-offs. This paper makes several key contributions as follows.
\begin{description}
    \item{\textit{1. Mathematical modelling:}} We design a comprehensive, generalised SC optimisation model as a multi-objective Markov decision process (MOMDP), incorporating multi-objective, multi-echelon networks, and fluctuating demand, emphasising economic, environmental, and social aspects. To the best of our knowledge, this is the first effort to craft a model of this breadth that can handle diverse network and parameter complexities (Sections~\ref{sec:MOMDP Framework} and \ref{sec:MOMDP-Based SC Optimisation}).
    \item{\textit{2. Simulation framework:}} We present \textit{Messiah (Multi-Echelon Supply SImulation AlgoritHm)}, a versatile simulation environment creator for SC problems designed for MDP-based models. Unlike traditional SC optimisation tools, \textit{Messiah} facilitates sequential decision-making amidst uncertainty and inherently integrates with RL frameworks, allowing for realistic and dynamic policy training and assessment (Section~\ref{sec:messiah}).
    % Users can configure different parameters, goals, and demands for the SC network. \textit{Messiah} enables creating SC environments reflecting various complexities of business in the real world
    \item{\textit{3. Algorithms:}} We encode solution representations using MOEA and MOMDP-based frameworks, highlighting differences in their mechanisms (Sections~\ref{sec:sol_nsgaII} and~\ref{sec:sol_momdp}). Then, we investigate the performance of two MOMDP-based and one multi-objective evolutionary algorithm (MOEA): MORL based on decomposition (MORL/D), proximal policy optimisation (PPO) with a weighted sum approach, and non-dominated sorting genetic algorithm II (NSGA-II), for a range of SC networks that vary in the number of facilities, markets, and echelons. We identify the potential best usage scenarios for these methods under specific SC circumstances (Section~\ref{sec:overall_performance}). %The objective of the proposed method and the baseline approaches is to generate a diverse set of trade-off solutions, allowing decision-makers to choose the most suitable option based on their priorities. 
    \item{\textit{4. Experimentation:}} We perform an analytical evaluation of how the solutions generated by the three methods perform and operate. Our study highlights the unique characteristics of the PF approximation sets and the varied impacts on manufacturing, inventory, and demand loss resulting from the implementation of solutions from each algorithm (Sections~\ref{sec:pf_solution_sets} and~\ref{sec:operational_behaviour}).
\end{description}

%To validate our framework, we conduct experimental case studies using three samples of SC optimisation case studies with simple, moderate, and complex network designs, emulating demand fluctuations, high-dimensional action (8, 21, and 59 dimensions), and observation spaces (20, 49, and 131 dimensions). 

Section~\ref{sec:related_work} provides context with relevant literature on multi-objective SC and RL applications. Section~\ref{sec:problem_definition} elucidates the formal definition of SC problems in the MOMDP framework and formulates SC problems in a generalised framework for both the optimisation and MDP-based models. Section~\ref{sec:methods} describes the methods, the customisable SC environment builder used in our experiments, and performance measurements for analytical comparison. Section~\ref{sec:experimentation} presents the experimental setup and analysis on various SC networks. The SC network configurations are based on the typical topologies and demand patterns of SC problems encountered by our industrial partner, Peak AI, Ltd, and all feature three objectives including profit maximisation, greenhouse (GHG) emission minimisation, and service level (SL) inequality (i.e., the cumulative gap of the demand fulfilment rate across markets~\citep{khodaee_humanitarian_2022}) minimisation. Lastly, Section~\ref{sec:conclusion} concludes the study and discusses potential future research areas.

\section{Related Work} \label{sec:related_work}
This section discusses retrospective studies in multi-objective SC optimisation, RL usage in SC optimisation, and the multi-objective framework in RL.

\subsection{Multi-Objective Sustainable SC}
SC activities include the procurement of raw materials, their conversion to finished goods and their distribution to markets~\citep{fahimnia_review_2013}. Sustainable SC optimisation involves managing the flow of products, services, and information between nodes, with an emphasis on economic, environmental, and social sustainability to address stakeholder concerns and ensure long-term success~\citep{ahi_comparative_2013}. Nearly 89\% of studies on sustainable SC focus on the economic aspect, measured through cost reduction, profit increase, or operational performance~\citep{jayarathna_multi-objective_2021}. Environmental sustainability metrics include minimising GHG emissions, driven by regulations like carbon taxes and emission limits~\citep{modak_using_2021,tenggren_climate_2020,turki_modelling_2018} and reducing environmental impact~\citep{mota_towards_2015,khalili_nasr_novel_2021}. Social sustainability in SC is more equivocal, with indicators such as social benefits, job creation, and reducing social impact~\citep{sharifi_novel_2020,allaoui_sustainable_2018,mota_towards_2015}. Distribution fairness, which gained prominence during the distribution of the COVID-19 vaccine, has been introduced as an alternative, with the aim of decreasing SL inequality~\citep{khodaee_humanitarian_2022}.

Optimisation of SC integrates functions such as procurement, manufacturing, warehousing, transportation, and market planning at the strategic, tactical, and operational levels~\citep{jayarathna_multi-objective_2021}. SCND at strategic and tactical levels involves the location of the facility, the capacity, and product flows, affecting competitiveness and investment~\citep{autry_definitive_2013,badri_two-stage_2017}. IMP at the tactical and operational levels balances stock levels to meet demand without inflating operational costs~\citep{singh_inventory_2018}. Environmental factors are frequently integrated into sustainable SCND, as evidenced by 96.8\% of the studies that examine this component. In contrast, social dimensions are addressed by approximately 31.7\% of the research~\citep{kumar_application_2024}. Economic and environmental integration is presented in 61.5\% of IMP studies, but only 15.4\% include the three sustainability pillars~\citep{becerra_green_2021}. Environmental assessments in SCND and IMP cover facilities, transport, and product disposal~\citep{pattnaik_recent_2021}. This study evaluates the environmental impact of facilities and transport measured by GHG emission, with an emphasis on the underexplored social dimension~\citep{asgharizadeh_sustainable_2019,becerra_green_2021} represented by the SL inequality.

\subsection{Solving Multi-Objective SC}\label{sec:solving_SC}
To solve multi-objective SC issues, various methods are applied, notably multi-objective optimisation techniques, which include classical, metaheuristic and hybrid approaches~\citep{jayarathna_multi-objective_2021}. Among classical methods, the $\epsilon$-constraint~\citep{abdolazimi_robust_2020, huang_waste_2020, banasik_closing_2017, mota_towards_2015} and weighted sum~\citep{chen_multiobjective_2014, darestani_robust_2019} are prevalent. The $\epsilon$-constraint method treats all objectives except the main one as constraints, while the weighted sum applies a vector of weights to the objective values, transforming the problem into a single-objective one. Nevertheless, classical approaches can oversimplify high-dimensional problems, leading researchers to favour metaheuristic methods such as simulated annealing~\citep{eskandarpour_sustainable_2015}, multi-objective particle swarm optimisation~\citep{wang_efficiency_2020}, Lagrangian-based heuristic~\citep{tautenhain_efficient_2021}, and the non-dominated sorting genetic algorithm II (NSGA-II)~\citep{farrokhi-asl_developing_2020, wang_efficiency_2020}. Simulated annealing explores solutions by allowing beneficial and some disadvantageous moves to avoid local optima, while particle swarm optimisation iteratively guides solutions towards convergence based on their own and their neighbours' best positions. The Lagrangian-based method uses Lagrange multipliers to include constraints within objective functions.

NSGA-II is a widely used MOEA-based method in SC problems, extending genetic algorithms with two key mechanisms: non-dominated ranking and crowding distance. These mechanisms measure solution optimality and diversity, respectively, enabling the algorithm to preserve the best solutions while maintaining a well-distributed Pareto front~\citep{deb_fast_2002}. NSGA-II is an elitist algorithm, which means that it retains the best solutions from generation to generation, improving convergence efficiency toward the Pareto-optimal front.~\citet{farrokhi-asl_developing_2020} suggested that NSGA-II represents better performance in a large-dimensional SC problem compared to the $\epsilon$-constraint method when applied in a waste vehicle routing problem. \citet{wang_efficiency_2020} compared four metaheuristic algorithms for the optimisation of multi-objective SC networks and revealed that NSGA-II shows the most efficient result. Its Pareto front (PF) approximation set performance is also superior in all corresponding objectives.~\citet{tirkolaee_multi-objective_2020} applied NSGA-II, $\epsilon$-constraint and multi-objective simulated annealing in a routing problem. Their results revealed that all PF solutions of NSGA-II are better than the others. They also exhibited that NSGA-II has the best efficiency. Hence, this paper employs the NSGA-II algorithm as a baseline method to solve our problem.

\subsection{MDP-Based Approaches}
Conventional optimisation methods work well for limited solution spaces but have low adaptability as they cannot learn from the environment~\citep{sutton_reinforcement_2015}. RL has become a promising approach that allows data-driven decisions through continuous interaction with the environment~\citep{boute_deep_2021, rolf_review_2022}. RL creates actions through policy and value functions to maximise cumulative long-term rewards, thus increasing adaptability~\citep{yan_reinforcement_2022}. By employing parameterised approximators, RL can overcome the 'curse of dimensionality' in extensive systems such as SC. In their practice, RL techniques must carefully balance exploration and exploitation. Its online learning paradigm provides immediate feedback on actions, thereby enabling continuous refinement of reward estimates. This sequential decision-making framework often allows RL to achieve a more effective equilibrium between exploring new strategies and exploiting proven ones than conventional metaheuristic methods.

To address real-world problems involving multiple objectives, RL has expanded to MORL. Most MORL algorithms use utility-based optimisation, where utility functions quantify the desirability of each objective, typically scaling from 0 to 1. This facilitates explicit interpretation of trade-offs between objectives based on their relative importance. Recent studies have clarified the MORL framework and its practical applications.~\citet{alegre_sample-efficient_2023} introduced an approach linking optimal policy transfer with successor features of MORL, ensuring optimal policies for any given preference. They also developed a sample-efficient MORL method that uses generalised policy improvement to detect significant preferences.~\citet{tran_two-stage_2023} proposed a novel multi-objective proximal distilled evolutionary RL algorithm, merging evolutionary algorithms with RL to optimise policies and improve learning. \citet{yang_generalized_2019} presented an envelope Q-learning algorithm that learns across preference spaces to optimise a single policy network, converging multi-objective results using linear preferences. In contrast, \citet{reymond_pareto_2022} highlighted the challenge of exploring expansive state spaces with additional objectives and introduced Pareto-conditioned networks, which efficiently manage multiple objectives using a single neural network. However, not all algorithms are suitable for high-dimensional action spaces, such as SC environments. MORL/D~\citep{felten_multi-objective_2023} is well-suited for such scenarios, combining RL with multi-objective optimisation concepts. It decomposes multi-objective problems into subproblems, using RL to generate diverse trade-off solutions within a modular framework that accommodates continuous policy representation.

\subsection{The Implementation of RL in SC}
RL methods have been applied in the SC domain over the past few decades. Table~\ref{tab:rl_implementation} summarises key SC and RL properties from several notable studies in RL-based SC optimisation to highlight existing research gaps.~\citet{stockheim_reinforcement_2003} applied RL for job scheduling in a decentralised SC management system, demonstrating superior performance compared to heuristic approaches.~\citet{oroojlooyjadid_deep_2017} utilised cooperative multi-agent RL for IMP in a partially observable environment.~\citet{kemmer_reinforcement_2018} found that RL agents outperformed the conventional static $(\mathit{r}, \mathit{Q})$-policy, where an order of size $\mathit{Q}$ is placed whenever the stock level drops to the reorder point $\mathit{r}$.~\citet{ganesan_adaptive_2019} implemented RL for inventory replenishment, reducing costs by 30\%. Similarly,~\citet{lalla-ruiz_deep_2020} showed that RL-based approaches achieved higher profitability than traditional linear programming techniques. In the healthcare sector,~\citet{abu_zwaida_optimization_2021} applied deep RL for drug replenishment, demonstrating better performance than traditional algorithms. Meanwhile, ~\citet{perez_algorithmic_2021} compared RL and optimisation-based methods, revealing that while optimisation approaches yield higher profits, RL-based techniques provide a more stable and resilient SC network. More recently,~\citet{dehaybe_deep_2024} employed RL for lot-sizing problems, optimising order quantities to minimise total costs under backorder and lost sales scenarios.

Although RL has been extensively employed in SC challenges, MORL for sustainable SC optimisation is still underexplored. Previous research, including works by \citet{shar_multi-objective_2023}, \citet{zhao_research_2025}, and our earlier study \citep{singh_multi-objective_2025}, have investigated MORL in simplified SC contexts. \citet{shar_multi-objective_2023} utilised MORL in a two-echelon serial SC network focusing on cost and GHG emissions. \citet{zhao_research_2025} examined single-echelon systems with deep Q-networks (DQN), while \citet{singh_multi-objective_2025} addressed economic, environmental, and social sustainability pillars in a specific case study. However, these studies did not tackle customisable, multi-echelon networks or scalable frameworks for strategic to tactical-level, multi-objective SC decision-making. Our study addresses this limitation by introducing a generalised MOMDP-based SC model that integrates IMP and SCND across various network complexities. It simultaneously captures economic, environmental, and social trade-offs, expanding MORL applicability to dynamic, multi-level SC contexts. This approach provides a holistic view on sustainable decision-making with insights at both strategic and operational levels.

\begin{landscape}
\begin{table}
\small
    \caption{\small Summary of several highlighted studies in RL implementation on SC domain, specifically in IMP. Most studies address IMP on simple networks without incorporating SCND. The most employed RL techniques are Q-learning (including DQN) and PPO. Single-objective SC dominates the literature with cost minimisation as the most common objective. Only an article involves flexible echelons. Our study addressed the literature gap by incorporating SCND in the multi-objective SC domain and enabling generalised SC models with flexible echelons.}
    \label{tab:rl_implementation}

    \begin{threeparttable}
    \begin{tabularx}{\linewidth}{p{1.9cm} p{1.2cm} p{1.3cm} p{1.2cm} p{2.7cm} p{2 cm}p{4.4 cm} p{2.8cm} p{1.7cm}}  % RL Algorithm
        \toprule %firsttable
        \multirow{2}{*}{\textbf{Articles}} & \multirow{2}{*}{\textbf{SCND}} & \multirow{2}{*}{\textbf{Multi-}} & \multirow{2}{*}{\textbf{Flexible}} & \multirow{2}{*}{\textbf{Network}} & \multicolumn{3}{c}{\textbf{RL Properties}} & \multirow{2}{*}{\textbf{RL Algo-}}\\
        \cmidrule(lr){6-8}
        & \textbf{included} & \textbf{objective}& \textbf{echelons} & \textbf{complexities} & Reward & State & Action &\textbf{rithm}\\
        \midrule

        \citet{oroojlooyjadid_deep_2017} & \ding{55} & \ding{55} & \ding{55} & 4 serial echelons & min cost & inventory level, on-going order, demand, size of shipment, action taken by other agent (multi-agent scenario) & order quantity & DQN\\

        \citet{kemmer_reinforcement_2018} & \ding{55} & \ding{55} & \ding{55} & 2 echelons: a factory, multiple warehouses & min cost & inventory level, demand & delivery amount & Approximate SARSA, REINFORCE\\

        \citet{ganesan_adaptive_2019} & \ding{55} & \ding{55} & \ding{55} & single facility & min cost & Inventory level, demand pattern, surplus stock, shortages, lead time variability & Inventory replenishment policy selection, order quantity, reorder timing & Q-Learning\\

        \citet{lee_application_2019} & \ding{55} & \ding{55} &\ding{55} & 3 echelons: suppliers, a factory, customers & max profit & inventory level, demand forecast, bidding history, production capacity, procurement cost, supplier offer & inventory threshold, bidding decision, procurement decision, product scheduling & Q-Learning, softmax, $\epsilon$-greedy\\

        \citet{peng_comprehensive_2020} & \ding{55} & \ding{55} & \ding{55} & 3 echelons: a factory, a warehouse, retailers & max profit & inventory level & production target, delivery amount & vanilla policy gradient\\

        \citet{tariq_afridi_deep_2020} & \ding{55} & \ding{55} & \ding{55} & 2 echelons: a supplier and a customer & target stock level & predicted stock position, actual demand, predicted target stock level & replenishment amount & DQN\\

        \citet{lalla-ruiz_deep_2020} & \ding{51} & \ding{55} & \ding{55} & 4 echelons: 2 suppliers, 2 factories, 2 wholesalers, 2 retailers & min cost & demand for next period, current stock level, material in transport, number of remaining period & delivery amount & PPO\\

        \citet{abu_zwaida_optimization_2021} & \ding{55} & \ding{55} & \ding{55} & single facility & min cost & inventory level, demand, buying cost, inventory cost, purchase budget, storage capacity, maximum amount to refill & refilling decision & DQN\\

        \bottomrule

    \end{tabularx}
    \end{threeparttable}
        
\end{table}
\end{landscape}

\begin{landscape}
\begin{table}
    \ContinuedFloat
    \small
    \caption{\small \textit{(continued)}}
    %\label{tab:rl_implementation}
    \renewcommand{\arraystretch}{1.2}
    \begin{threeparttable}
    \begin{tabularx}{\linewidth}{p{1.9cm} p{1.2cm} p{1.3cm} p{1.2cm} p{2.7cm} p{2 cm}p{4.4 cm} p{2.8cm} p{1.7cm}}  % RL Algorithm
        \toprule %2ndtable
        \multirow{2}{*}{\textbf{Articles}} & \multirow{2}{*}{\textbf{SCND}} & \multirow{2}{*}{\textbf{Multi-}} & \multirow{2}{*}{\textbf{Flexible}} & \multirow{2}{*}{\textbf{Network}} & \multicolumn{3}{c}{\textbf{RL Properties}} & \multirow{2}{*}{\textbf{RL Algo-}}\\
        \cmidrule(lr){6-8}
        & \textbf{included} & \textbf{objective}& \textbf{echelons} & \textbf{complexities} & Reward & State & Action & \textbf{rithm}\\
        \midrule

        \citet{perez_algorithmic_2021} & \ding{51} & \ding{55} & \ding{55} & 4 echelons: 2 suppliers, 3 factories, 2 warehouses, a retailer & max profit & inventory level, service level & delivery quantity & PPO\\

        \citet{shar_multi-objective_2023} & \ding{55} & \ding{51} & \ding{55} & 3 echelons: a factory, a warehouse, a retailer & min cost, min GHG emission & inventory level, outstanding order, cumulative carbon credit, carbon credit price & order quantity, carbon buy/sell decision & MONES, NES, EMOQL\tnote{1}\\

        \citet{rizqi_neuroevolution_2024} & \ding{55} & \ding{55} & \ding{51}\tnote{2} & 2 and 3 echelons & min cost & inventory position, supplier discount, unfulfilled demand, lead time variability & order quantity, delivery mode selection, supplier selection & Neuroevolution RL\\

        %\citet{wang_optimisation_2024} & \ding{51} & \ding{51} & \ding{55} & 5 echelons: multiple facilities at each echelon & min cost, min emission, max investment return & inventory levels, recycling volumes, carbon emission, government intervention status, transportation network status & recycling channel selection, inventory and production decision, transport route selection, policy response & Improved Multi-Objective Deep RL (IMDRL)\\

        \citet{dehaybe_deep_2024} & \ding{55} & \ding{55} & \ding{55} & single facility & min cost & forecasted demand, inventory level & order quantity & PPO\\

        \citet{zhao_research_2025} & \ding{55} & \ding{51} & \ding{55} & single echelon: factories and customers & min cost, max time efficiency, min carbon emission & inventory level, order status, production capacity, and transportation status & procurement action, transportation choice & DQN with scalarisation \\

        \citet{singh_multi-objective_2025} & \ding{51} & \ding{51} & \ding{55} & 4 echelons: 2 suppliers, 3 factories, 2 warehouses, 3 markets & max profit, min emission, min SL inequality & inventory level, order status, cumulative emission, average SL inequality & manufacturing quantity, delivery quantity & MORL/D, PPO\\
        
        \midrule
        \textbf{This study} & \ding{51} & \ding{51} & \ding{51} & \textbf{generalised} multiple echelons with multiple, \textbf{customisable} facilities, and customers & max profit, min emission, min SL inequality & inventory level, order status, cumulative emission, average SL inequality & manufacturing quantity, delivery quantity & MORL/D, PPO\\

        \bottomrule
    \end{tabularx}

    \begin{tablenotes}
    \footnotesize\item[1] Multi-objective natural evolution strategy (MONES), natural evolution strategy (NES), envelope multi-objective Q-learning (EMOQL).
    \footnotesize \item[2] Each model is individually crafted, limited to 2 echelon types only.
    \end{tablenotes}
    \end{threeparttable}
\end{table}
\end{landscape}

\section{Problem Definition}\label{sec:problem_definition}

\begin{figure}[h]
    \centering
    \includegraphics[width=12 cm]{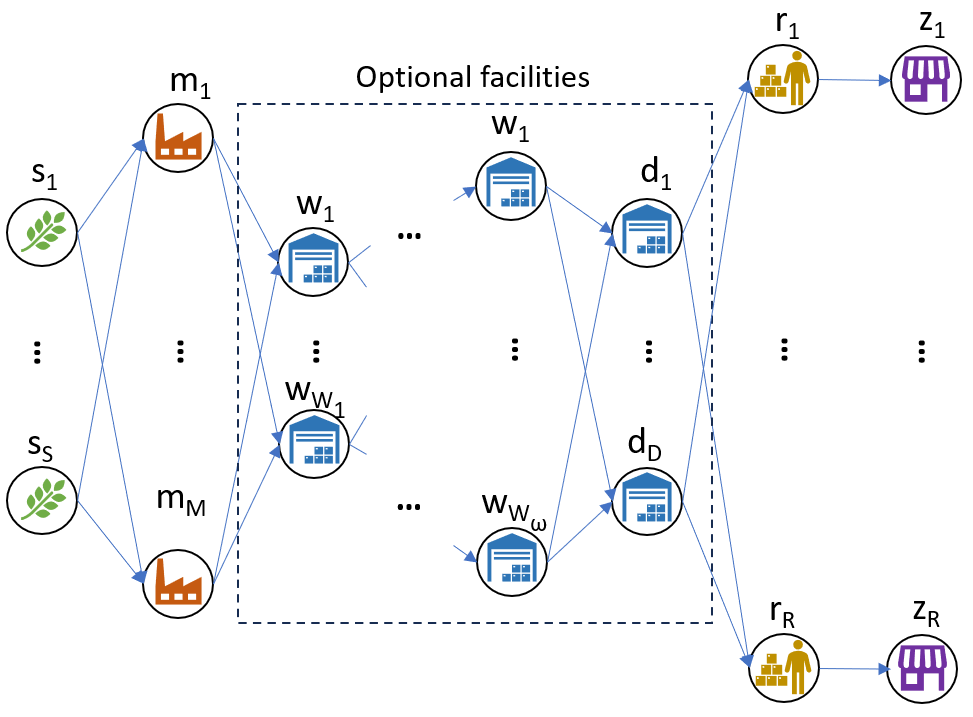}
    \caption{Generalised SC network consisting of $S$ number of suppliers, $M$ number of manufacturers, $\omega$ echelons of warehouses with customisable number at each echelon, $D$ number of distribution centre, and $R$ number of retailers and markets. The warehouses and distribution centres are optional, defining the number of the whole network's echelon number.}
    \label{fig:generalised_sc}
\end{figure}

This section presents the SC challenges addressed in this study. Our formulation stems from the same fundamental SC problem but is expressed through two distinct frameworks: an optimisation-based approach, defined as a multi-objective nonlinear programming problem~\citep{miettinen_nonlinear_1998}, and a MOMDP framework. Building on our prior work, which examined a specific scenario, we have broadened our research to establish a general framework suitable for addressing multi-objective SCND and IMP, which is one of our main contributions. This framework is capable of handling systems that involve multiple echelons, facilities, and markets with various demand patterns, as depicted in Figure~\ref{fig:generalised_sc}. Table~\ref{tab:parameter_definition} enumerates the variables, parameters, as well as some sets and a family of sets to differentiate models between SC networks with varying numbers of echelons.

\subsection{Optimisation-Based SC Problem} \label{sec:optimisation_based}
This part formulates the problem based on the traditional optimisation approach that determines the solution as cumulative values over a specified time period.\\

\setlength{\textfloatsep}{5pt}
\begin{table}[H]
\small
    \centering
    \caption{\small Sets, parameters, and variables used in the model. Nodes $i$, $j$, $k$ ($i < j < k$) represent supply chain facilities. Indices $t$ denotes time period , while $\varsigma$, $\psi$, and $\tau$ correspond to inventory, production, and transportation activities, respectively.}
    \label{tab:parameter_definition}
    \begin{threeparttable}
    \begin{tabularx}{15cm}{ll}
        \toprule
        Notations & Definition\\
        \midrule
        \textbf{Sets}\\
        $N^s$ & suppliers\\
        $N^m$ & manufacturers\\
        $N^{w1},\dots, N^{w\omega}$ & mid-level warehouses that deliver to lower-level warehouses\tnote{*}\\
        $N^d$ & distribution centres that deliver to retailers\tnote{*}\\
        $N^r$ & retailers\\
        $N^z$ & markets\\
        \midrule
        \textbf{Family of sets}\\
        $N^{From}$ & origin nodes ($|N^{From}| = n_1$)\\
        $N^{To}$ & destination nodes($|N^{To}| = n_2$)\\
        $N^{Inv}$ & nodes that store inventories without producing ($|N^{Inv}| = n_3$)\\
        $N^{Invf}$ & nodes that ship goods for next node inventories ($|N^{Invf}| = n_4$)\\
        $N^{Invt}$ & nodes that receive inventories from previous nodes ($|N^{Invt}| = n_5$)\\
        \midrule
        \textbf{Parameters}\\
        $T$ & Total period horizon\\
        $Profit$ & Total profit of all periods\\
        $E$ & Total emission of all periods\\
        $F$ & Total SL inequality measure of all periods\\
        $SL_y$ & SL at node $y$ of all periods\\
        $Rev_t$ & Revenue at period $t$\\
        $PC_t$ & Production cost at period $t$\\
        $TC_t$ & Transport cost at period $t$\\
        $IC_t$ & Inventory cost at period $t$\\
        $L$ & Transport lead time\\
        $I_{tj}$ & Inventory unit at period $t$, node $j$\\
        $c_j^\varsigma$ & Inventory cost per unit at node $j$\\
        $e^{\varsigma}_j$ & Emission per unit resulted from inventory at node $j$\\
        $c_k^\psi$ & Manufacturing cost per unit at node $k$\\
        $v_k^\psi$ & Yield ratio from manufacturing process at node $k$\\
        $e^{\psi}_k$ & Emission per unit resulted from the manufacturing process at node $k$\\
        $c_{ij}^\tau$ & Transport cost per unit per day from node $i$ to $j$\\
        $e^{\tau}_{ij}$ & Emission per unit per day resulted from product transport from node $i$ to $j$\\
        $\mathbf{Q_{tj}}$ & The outstanding order at node $j$\\
        $CE_t$ & The accumulated emission at period $t$\\
        $AF_t$ & The average service level inequality in period $t$\\
        $d_{tj}$ & Demand at period $t$, retailer $j$\\
        $Cap$ & Transport capacity\\
        $price_j$ & Product selling price at retailer $j$\\
        $\mathcal{M}$ & Very large number (big-M)\\
        \midrule
        \textbf{Variables}\\
        $q^\tau_{tij}$ & Transport quantity at period $t$, from node $i$ to $j$\\
        $q^\psi_{tk}$ & Manufacturing quantity at period $t$, at manufacturer $k$\\
        \bottomrule     
    \end{tabularx}

    \begin{tablenotes}
        \footnotesize
        \item[*] Optional nodes in the SC. Mid-level warehouses, which may span multiple echelons, exist only if distribution centres (i.e., the lowest-level warehouses) are present. Without warehouses, manufacturers act as distribution centres. %Mid-level warehouses may span multiple echelons.
    \end{tablenotes}
    \end{threeparttable}
\end{table}

\noindent\textbf{Objective Functions.}\quad The multiple objectives are given in separate functions as shown in Equations~\eqref{eq:ga_profit},~\eqref{eq:ga_emission},~\eqref{eq:ga_ineq}, encompassing profit maximisation, emission minimisation, and SL inequality minimisation, respectively.

\paragraph{Objective-1: Profit.}

Equation~\eqref{eq:ga_profit} shows that $Profit$ is calculated from $Rev_t$ subtracted by $PC_t$, $TC_t$ and $IC_t$ for the entire period, as follows: 
\begin{equation}
    \label{eq:ga_profit}
    Profit=\sum^{T}_{t=1}{\left(Rev_t-\left(PC_t+TC_t+IC_t\right)\right)}.
\end{equation}

\noindent Each element in the profit calculation is broken down into several aspects. Equation~\eqref{eq:ga_revenue} shows that $Rev_t$ is obtained from the total product absorbed in the markets, multiplied by the unit price, as follows:
\begin{equation}
\label{eq:ga_revenue}
    Rev_t= \sum_{i \in N^d} {\sum_{j \in N^r}{\left(q_{\left(t-L\right)ij}^\tau \right)}} \cdot price_j.
\end{equation}

\noindent $PC_t$ is affected by the operational cost and yield ratio, as exhibited in Equation~\eqref{eq:ga_production_cost} below:
\begin{equation}
\label{eq:ga_production_cost}
    PC_t = \sum_{k \in N^m} \left(\frac{ q_{tk}^\psi \cdot c^\psi_k}{v^\psi_k}\right).
\end{equation}

\noindent $TC_t$ consists of all unit transport costs incurred during the transport lead time, as given in Equation~\eqref{eq:ga_transport_cost} below:
\begin{equation}
\label{eq:ga_transport_cost}
    T C_t = \sum_{i \in N^{From}}\sum_{j \in N^{To}}\left(q^\tau_{tij} \cdot c^\tau_{ij} \cdot L \right).
\end{equation}

\noindent Equation~\eqref{eq:ga_inventory_cost} defines $IC_t$, calculated from the total inventory on hand multiplied by the unit holding cost below:
\begin{equation}
\label{eq:ga_inventory_cost}
    IC_t=\sum_{j \in N^{To}}\left(I_{tj} \cdot c^\varsigma_j\right).
\end{equation}

\paragraph{Objective-2: GHG emission.}

The second objective alludes to the environmental aspect represented by the GHG emissions~\citep{shar_multi-objective_2023}. Equation~\eqref{eq:ga_emission} illustrates the formulation of total GHG emissions $E$, including those that arise during inventory, manufacturing, and transportation, as given by the following:
\begin{equation}
\label{eq:ga_emission}
    \begin{split}
        E =& \sum^{T}_{t=1}\left(\sum_{j \in N^{To}}\left(I_{tj} \cdot e^\varsigma_j\right)+\sum_{k \in N^{m}}\left(q^\psi_{tk} \cdot e^\psi_k\right)+ \sum_{i \in N^{From}}\sum_{j \in N^{To}}\left(q_{ij}^\tau \cdot e_{ij}^\tau \cdot L\right)\right).
    \end{split}
\end{equation}

\paragraph{Objective-3: SL inequality.}

The third objective is the social aspect, represented by SL inequality $F$, calculated in Equation~\eqref{eq:ga_ineq}, which is the total discrepancies between each pair of SLs in the markets~\citep{khodaee_humanitarian_2022}, as follows:
\begin{equation}
\label{eq:ga_ineq}
    \begin{split}
    F =& \frac{1}{2} \sum_{j\in N^r} \sum_{\substack{j' \in N^r \\ j' \neq j}} \left| SL_{j} - SL_{j'} \right|.
    \end{split}
\end{equation}
We maximise the objective vector $\mathbf{R} = \{Prof, -E, -F\}$. These objectives are subject to several constraints that are formulated by Equation~\eqref{eq:ga_mfg_qty} to~\eqref{eq:ga_mfg}. The constraint~\eqref{eq:ga_mfg_qty} guarantees that all materials received from the suppliers are passed on to the manufacturing stage:

\begin{equation}
    \label{eq:ga_mfg_qty}
    q_{tj}^{\psi} = \sum_{i \in N^s} q_{(t-L)ij}^{\tau} \quad \forall t, \forall j \in N^m.
\end{equation}

The constraint~\eqref{eq:ga_inv_all} computes the inventory of the next period based on previous inventory, goods received (i.e., inflow), and deliveries dispatched (i.e., outflow). In Equation~\eqref{eq:ga_inv_all}, $I_{tj}$ is defined for $j$ in the set $N^m$ and $N^{Inv}$, where $N^{Inv}_1$ represents the first set in $N^{Inv}$ and $N^{Invt}_{\{2,\dots,n_5\}}$ represents the second set to the last set in $N^{Invt}$, as follows:

\begin{align}
\label{eq:ga_inv_all}
    I_{tj} &= I_{(t-1)j} + \text{inflow}_{tj} - \text{outflow}_{tj}, \quad \forall t, \forall j \in N^{Inv} \cup N^m\\
    \text{where} \quad
    \text{inflow}_{tj} &=
    \begin{cases}
        q_{tj}^{\psi} & \text{if } j \in N^m \\
        \sum\limits_{i \in N^{Invf}} q_{(t-L)ij}^{\tau} & \text{if } j \in N^{Inv}
    \end{cases} \notag \\
    \text{and} \quad
    \text{outflow}_{tj} &=
    \begin{cases}
        \sum_{k \in N^{Inv}_1} q_{tjk}^{\tau} & \text{if } j \in N^m\\
        \sum_{k \in N^{Invt}_{\{2,\dots,n_5\}}}q_{tjk}^\tau
&\text{if } j \in N^{Inv}.
    \end{cases} \notag
\end{align}

%\begin{equation}
%    \label{eq:ga_inv_all}
%    I_{tj} =
%   \begin{cases}
%    I_{(t-1)j} + q_{tj}^{\psi} - \sum_{k \in N^{Inv}_1} q_{tjk}^{\tau} &,\quad \forall t, \forall j \in N^m\\[10pt]
%    I_{(t-1)j} + \sum_{i \in N^{Invf}} \left(q_{(t-L)ij}^\tau\right)-
%    \sum_{k \in N^{Invt}_{\{|N^{Invf}|+1,\dots,|N^{Invt}|\}}}q_{tjk}^\tau
%    &,\quad \forall t, \forall j \in N^{Inv}
%    \end{cases}
%    ,\quad\text{where}\quad i<j<k
%\end{equation}

\noindent The constraint~\eqref{eq:ga_inv_ieq} ensures that shipments do not exceed inventory levels and keep inventory non-negative since delivery amounts cannot be negative:
\begin{equation}
\label{eq:ga_inv_ieq}
    \begin{split}
    &\sum_{k \in N^{Invt}} q^\tau_{(t+1)jk} \leq I_{tj} \quad \forall t, \forall j \in N^{To}.
    \end{split}
\end{equation}

\noindent The constraint~\eqref{eq:ga_sl} gauges the SL in each market as the percentage of demand met, ranging from 0\% to 100\%:

\begin{equation}
\label{eq:ga_sl}
    \begin{split}
        SL_{j} = \min \left(\frac{\sum_{t=(t-L)}^T\sum_{i \in N^d} q^\tau_{tij}}{\sum_t d_{tj}}, 100\% \right)\quad \forall j \in N^r,
    \end{split}
\end{equation}
\noindent where $d_{tj}$ is associated demand for each retailer $j$. Constraint~\eqref{eq:ga_transport} restricts delivery amounts to between zero and the transport capacity, as follows:

\begin{equation}
    \label{eq:ga_transport}
    0 \leq q_{tij}^{\tau} \leq Cap \quad \forall t, \forall{i} \in N^{From}, \forall j \in N^{To}.
\end{equation}

\noindent Finally, constraint~\eqref{eq:ga_mfg} enforces non-negative manufacturing quantities:

\begin{equation}
    \label{eq:ga_mfg}
    0 \leq q_{tk}^{\psi} \quad \forall t, \forall k \in N^{m}.
\end{equation}
In optimising combinatorial problems like SCND, the dimensionality rises along with the SC network complexity. To illustrate, each defined SC network contains $|\text{decision variables}|=(\text{number of delivery routes} + \text{number of manufacturers}) \times \text{period horizon}$.

\subsection{MOMDP Framework for SC Problems} \label{sec:MOMDP Framework}
Sequential decision-making reduces the effective dimensionality of combinatorial problems by decomposing the overall decision into a sequence of lower-dimensional subproblems. At each decision point, the current state, which captures all relevant past decisions, conditions the available actions. This mitigates the combinatorial explosion that arises when decisions for all periods are made simultaneously. In our SC environment, A MORL agent trained operates on the basis of sequential decision-making modelled as a finite MOMDP, with period $t=1, 2, 3,\dots, T$, which is a day-to-day sequence in this SC problem. A MOMDP problem is defined as a tuple $\langle S,A,ST,\gamma,\mu,\mathbf{R}\rangle$, as explained by~\citet{hayes_practical_2022}. At each period $t$, it perceives the state of the environment $S_t=\mathcal{S}$ based on its observation in the previous state, where $\mathcal{S}$ is a set of possible states. In an SC environment, the state typically represents the level of inventory at each facility, the number of orders being processed on each route, and the demand at each market. Furthermore, in multi-objective SC, the observation can be extended to cumulative emission value at each period and the average SL inequality of all markets.

Given the state, the agent decides on an action, $A_t=\mathcal{A}(S_t)$, where $\mathcal{A}(S_t)$ is the set of available actions in the state $S_t$. The actions, which typically constitute manufacturing and delivery quantities, are selected based on the probability distribution generated by the agent, namely the agent policy $\pi_t$. The $\pi_t=(a\mid s)$ is the probability of $A_t=a$ if $S_t=s$. Meanwhile, $ST_t$ represents the probability function of the state transition, where $ST_t(s, a, s') = P(s_{t+1} = s' | s_t = s, a_t = a)$. Moreover, $\mu$ denotes the probability distribution in the initial state, given by $\mu_t(s) = P(s_t = s)$.

As an action set is applied, where the agent determines a set of manufacturing and delivery quantities and releases them to the SC environment, the latter updates the state based on the state transition functions. Afterwards, the reward is calculated based on the reward function, and the agent receives it as a consequence of its selected action. In MORL, the environment provides multiple rewards based on several conflicting objectives that are optimised simultaneously (see Section~\ref{MORL_Mechanism}). The key distinction between MOMDP and MDP with a single objective is signified in the representation of the reward functions. In MOMDP, the reward functions are given in a vector value, where $\mathbf{R}\subset\mathbb{R}^\textit{d}$. This vector could consist of total profit, negative GHG emissions, and negative SL inequality in the SC environment. Negative values are set for the reward minimisation scenario, as the agent will maximise the cumulative return that is depicted in a vector of value functions that corresponds to a policy $\pi$~\citep{roijers_survey_2013}, as follows:
\begin{equation}
\label{eq:vector_value}
\mathbf{V}^\pi=\mathbb{E}\left[\sum_{k=0}^{\infty} \gamma^k \mathbf{r}_{k+1} \mid \pi, \mu\right].
\end{equation}
Without prior information about users' utility functions, several optimal value vectors could exist in the multi-objective cases. These vectors are associated with a set of non-dominated policies, namely Pareto front approximation $PF(\mathbf{\Pi})$, where $\mathbf{\Pi}$ is a set of all possible policies. MORL cases assume that the utility function $u$ is increasing monotonically (that is, if a policy improves one or more objectives without compromising the rest, the scalarised value will also rise). Referring to~\citet{hayes_practical_2022} let $b$ be the objective index, the PF approximation is given by:
\begin{equation}
\label{eq:pareto_front}
PF(\mathbf{\Pi})=\{\pi\in\mathbf{\Pi}\mid \nexists \pi'\in\mathbf{\Pi}:\mathbf{V}^{\pi'}\succ_P\mathbf{V}^{\pi}\},
\end{equation}
where
\begin{equation}
\label{eq:domination}
\mathbf{V}^{\pi} \succ_P \mathbf{V}^{\pi'} \iff \left( \forall b : \mathbf{V}_b^{\pi} \geq \mathbf{V}_b^{\pi'} \right) \land \left( \exists b : \mathbf{V}_b^{\pi} > \mathbf{V}_b^{\pi'} \right).
\end{equation}

\subsection{MOMDP-Based SC Problem} \label{sec:MOMDP-Based SC Optimisation}
Multi-echelon SC problems with multiple facilities, markets, and non-stationary demand are formulated based on the MOMDP framework.

\subsubsection{State Transition Functions}
This represents the functions when moving from one state at period $(t-1)$ to the next state at period $t$. The state transition functions are given in Equations~\eqref{eq:rl_mfg_qty} to~\eqref{eq:rl_avg_ineq}. Equation~\eqref{eq:rl_mfg_qty} ensures that all raw materials provided by suppliers are manufactured into finished products, as follows:

\begin{equation}
    \label{eq:rl_mfg_qty}
    q_{tj}^{\psi} = \sum_{i \in N^s} q_{(t-L)ij}^{\tau} \quad \forall t, \forall j \in N^m.
\end{equation}

\noindent Equation~\eqref{eq:rl_inv_all} is similar to equation~\eqref{eq:ga_inv_all}, which calculates the inventory levels considering the remainder of previous periods, additional goods (i.e., inflow), and goods shipped (i.e., outflow):
\begin{align}
\label{eq:rl_inv_all}
    I_{tj} &= I_{(t-1)j} + \text{inflow}_{tj} - \text{outflow}_{tj}, \quad \forall t, \forall j \in N^{Inv} \cup N^m\\
    \text{where} \quad
    \text{inflow}_{tj} &=
    \begin{cases}
        q_{tj}^{\psi} & \text{if } j \in N^m \\
        \sum\limits_{i \in N^{Invf}} q_{(t-L)ij}^{\tau} & \text{if } j \in N^{Inv}
    \end{cases} \notag \\
    \text{and} \quad
    \text{outflow}_{tj} &=
    \begin{cases}
        \sum_{k \in N^{Inv}_1} q_{tjk}^{\tau} & \text{if } j \in N^m\\
        \sum_{k \in N^{Invt}_{\{2,\dots,n_5\}}}q_{tjk}^\tau
&\text{if } j \in N^{Inv}.
    \end{cases} \notag
\end{align}

%\begin{equation}
%    \label{eq:rl_inv_all2}
%    I_{tj} =
%   \begin{cases}
%    I_{(t-1)j} + q_{tj}^{\psi} - \sum_{k \in N^{Inv}_1} q_{tjk}^{\tau} &,\quad \forall t, \forall j \in N^m\\[10pt]
%    I_{(t-1)j} + \sum_{i \in N^{Invf}} \left(q_{(t-L)ij}^\tau\right)-
%    \sum_{k \in N^{Invt}_{\{|N^{Invf}|+1,\dots,|N^{Invt}|\}}}q_{tjk}^\tau
%    &,\quad \forall t, \forall j \in N^{Inv}
%    \end{cases}
%    ,\quad\text{where}\quad i<j<k
%\end{equation}

\noindent Equation~\eqref{eq:rl_order_vector} defines the total order at each node, which is the cumulative incoming deliveries:
\begin{equation}
    %\begin{split}
    \label{eq:rl_order_vector}
    \mathbf{Q}_{tj} = \big\{ q^\tau_{(t-L+1)ij}, \ldots, q^\tau_{tij}, \quad \forall i \in N^{From},\forall j \in N^{To} \big\}.
    %\end{split}
\end{equation}

\noindent Equation~\eqref{eq:rl_cum_emission} calculates the cumulative emission on each day, which is:
\begin{equation}
    \label{eq:rl_cum_emission}
    CE_t=CE_{(t-1)}+E_t.
\end{equation}

\noindent Equation~\eqref{eq:rl_avg_ineq} calculates average SL inequality on each day:
\begin{equation}
    \label{eq:rl_avg_ineq}
    AF_t=\frac{\left(AF_{t-1} \cdot(t-1)+F_t\right)}{t}.
\end{equation}

\noindent In our SC problem, the state transition is assumed to be deterministic, where no inherent randomness is introduced (e.g., in lead time fluctuation and defective product disposal that engender uncertainty in the state transition, specifically in inventory calculation at the next period).

\subsubsection{State Space (S)}
The state of the system represents the observation of the learnt environment by the agents. Our SC model includes inventory levels for raw materials and products, outstanding orders on all routes for a certain lead time, cumulative emission, and average SL inequality, where $0 \leq I_{(t-1)j}$. The state at period $t$ is defined as follows:
\begin{equation}
    \label{eq:rl_state_vector}
    \mathbf{S_t}=\{I_{(t-1)j}, \mathbf{Q_{(t-1)j}}, CE_{t-1}, AF_{t-1},\quad \forall{j} \in N^{To}\}.
\end{equation}

\subsubsection{Action Space (A)}
In our SC environment, the action covers manufacturing quantities of all manufacturers and delivery quantities of all routes. The actions at period $t$ are given by:
\begin{equation}
    \label{eq:rl_action_vector}
    \mathbf{A_t}=\{q_{tij}^{\tau}, q_{tk}^{\psi}, \quad 0\leq q_{tij}^{\tau} \leq Cap, \quad 0\leq q_{tk}^{\psi}, \quad \forall{i} \in N^{From},
    j \in N^{To}, k \in N^{m}\}.
\end{equation}

\subsubsection{Rewards (R)}
Three objectives are incorporated into this problem: profit maximisation, GHG emission minimisation, and SL inequality minimisation, which constitutes a three-dimensional vector reward. It is emphasised that the MORL agents are set to maximise the expected cumulative rewards. Objectives 2 and 3 should have negative reward values. Thus, the reward vector in this problem is defined as $\mathbf{R_t}=\{\mathit{Prof_t},-\mathit{E_t},-\mathit{F_t}\}\cdot\rho_t$, where $\rho$ represents penalty should any inventory falls below zero as given in the Equation~\eqref{eq:penalty}:
\begin{equation}
\label{eq:penalty}
    \rho_t = \sum_{j \in N^{To}} \min(I_{tj},0) \cdot \mathcal{M}.
\end{equation}
\noindent In RL, the reward values are derived from objective functions at each period as given in Equations~\eqref{eq:rl_profit}, ~\eqref{eq:rl_emission}, and ~\eqref{eq:rl_sl_ineq} below:

\begin{equation}
\label{eq:rl_profit}
\begin{aligned}
Profit_t =\;& 
\underbrace{\sum_{i \in N^d} \sum_{j \in N^r} q_{(t-L)ij}^\tau \cdot price_j}_{\text{Revenue from delivered products}} 
\; - \;
\underbrace{\sum_{k \in N^m} \left( \frac{q_{tk}^\psi \cdot c_k^\psi}{v_k^\psi} \right)}_{\text{Manufacturing cost}}
 - \underbrace{\sum_{i \in N^{From}} \sum_{j \in N^{To}} q_{tij}^\tau \cdot c_{ij}^\tau \cdot L}_{\text{Transport cost over lead time}}\\ 
&\; - \;
\underbrace{\sum_{j \in N^{To}} I_{tj} \cdot c_j^\varsigma}_{\text{Inventory holding cost}}
\end{aligned}
\end{equation}

\begin{equation}
\label{eq:rl_emission}
\begin{aligned}
E_t =\;& 
\underbrace{\sum_{j \in N^{To}} I_{tj} \cdot e^\varsigma_j}_{\text{Inventory-related emissions}} 
\; + \;
\underbrace{\sum_{k \in N^m} q_{tk}^\psi \cdot e_k^\psi}_{\text{Manufacturing emissions}} 
\; + \;
\underbrace{\sum_{i \in N^{From}} \sum_{j \in N^{To}} q_{ij}^\tau \cdot e_{ij}^\tau \cdot L}_{\text{Transport emissions over lead time}}
\end{aligned}
\end{equation}

%\begin{equation}
%\label{eq:rl_emission}
%    \text{Minimise}\quad E_t =& \Bigg(\sum_{j \in N^{To}}\left(I_{tj} \cdot e^\varsigma_j\right)+\sum_{k \in N^{m}}\left(q^\psi_{tk} \cdot e^\psi_k\right)
%    +\sum_{i \in N^{From}}\sum_{j \in N^{To}}\left(q_{ij}^\tau \cdot e_{ij}^\tau \cdot L\right)\Bigg)
%\end{equation}

\begin{equation}
\label{eq:rl_sl_ineq}
\begin{aligned}
F_t =\;& 
\frac{1}{2} \sum_{j \in N^r} \sum_{\substack{j' \in N^r \\ j' \neq j}} 
\left| SL_{tj} - SL_{tj'} \right|,
\end{aligned}
\end{equation}

%\begin{equation}
%\label{eq:rl_sl_ineq}
%    \begin{split}
%    \text{Minimise}\quad F_t =& \frac{1}{2} \sum_{y\in N^r} \sum_{\substack{y' \in N^r \\ y' \neq y}} \left| SL_{ty} - SL_{ty'} \right|,
%    \end{split}
%\end{equation}

\noindent{where:}

\begin{equation}
\label{eq:rl_sl}
    \begin{split}
        SL_{tj} = \min \left(\frac{\sum_{i \in N^d} q^\tau_{(t-L)ij}}{d_{tj}}, 100\% \right),\quad \forall j \in N^r,\\
    \end{split}
\end{equation}
\noindent where $d_{tj}$ is associated demand for each retailer $j$. Developing a model within the MDP framework can be less straightforward than crafting an optimisation model, due to the unique elements each approach defines. Although most formulations based on optimisation and MDP are equivalent, they are not always identical. In particular, the MDP-based approach often represents equations sequentially. Table~\eqref{tab:comp_equations} summarises the equivalence between the equations given in the models based on optimisation and MDP.

\begin{table}[h]
    \centering
    \caption{The overview of the equivalence between optimisation-based and MDP-based models encompasses the representation of formulations and their respective definitions within each framework. The constraints present in the optimisation model are entirely addressed by the MDP approach, furthermore, the latter includes additional formulations concerning observations.}
    \label{tab:comp_equations}
    \begin{tabularx}{16cm}{@{}XXXX@{}}
    \toprule
    Equations & Representation & Optimisation & MDP\\
    \midrule
    ~\eqref{eq:ga_profit} \&~\eqref{eq:rl_profit} & total profit & objective & reward\\
    ~\eqref{eq:ga_emission} \&~\eqref{eq:rl_emission} & total GHG emission & objective & reward\\
    ~\eqref{eq:ga_ineq} \&~\eqref{eq:rl_sl_ineq} & total SL inequality & objective & reward\\
    ~\eqref{eq:ga_mfg_qty} \&~\eqref{eq:rl_mfg_qty} & manufacturing quantity calculation & constraint & state transition\\
    ~\eqref{eq:ga_inv_all} \&~\eqref{eq:rl_inv_all} & inventory calculation & constraint & state transition\\
    ~\eqref{eq:ga_inv_ieq} \&~\eqref{eq:penalty} & inventory non-negativity enforcement & constraint & reward penalty\\
    ~\eqref{eq:ga_sl} \&~\eqref{eq:rl_sl} & SL calculation & constraint & reward (part of calculation)\\
    ~\eqref{eq:ga_transport} \&~\eqref{eq:rl_action_vector} & capacity limitation & constraint & action space\\
    ~\eqref{eq:ga_mfg} \&~\eqref{eq:rl_action_vector} & delivery non-negativity enforcement & constraint & action space\\
    \bottomrule
    \end{tabularx}
\end{table}

%These objective functions in RL (Equations~\eqref{eq:rl_profit}, ~\eqref{eq:rl_emission}, and ~\eqref{eq:rl_sl_ineq}) are equivalent to equations~\eqref{eq:ga_profit}, ~\eqref{eq:ga_emission}, and ~\eqref{eq:ga_ineq} in the optimisation model. Formulating constraints in RL can be less straightforward than the optimisation model due to the distinct elements specified by each method. Equations~\eqref{eq:rl_mfg_qty} and~\eqref{eq:rl_inv_all} of RL are equivalent with constraints~\eqref{eq:ga_mfg_qty} and ~\eqref{eq:ga_inv_all} respectively. The constraint~\eqref{eq:ga_inv_ieq} corresponds to the reward penalty given in Equation~\eqref{eq:penalty}, which prevents negative inventories. The constraint~\eqref{eq:ga_sl} becomes part of the calculation of the SL inequality reward in Equation~\eqref{eq:rl_sl}. The constraints~\eqref{eq:ga_transport} and~\eqref{eq:ga_mfg} are defined in the RL action space part (see Equation~\eqref{eq:rl_action_vector}).

Finally, some MDP model observations are not captured in the optimisation-based model as environment feedback. These components are not considered constraints in the model. Yet, they are valuable for more informed decision-making, as this signifies the interaction between the agent and the environment during the decision-making process. In this case, RL agents are informed about the vector of quantity of ongoing delivery, cumulative GHG emissions, and the average SL inequality at each period (see Equations~\eqref{eq:rl_order_vector},~\eqref{eq:rl_cum_emission},~\eqref{eq:rl_avg_ineq}) as a result of their actions, while NSGA-II is not. This is considered a perk of implementing RL rather than optimisation models.

\section{Methods}\label{sec:methods}
We address the SC optimisation problem using two distinct approaches: a MOEA-based approach for the optimisation-based formulation and RL methods for the MDP-based formulation. In SCND practice, facility candidates such as suppliers, manufacturers, warehouses, and retailers are specified prior to the optimisation process. Based on established facility candidates, the available transport routes are determined. Subsequently, the algorithms identify which facilities and routes are used and to what extent they are employed. The simulation model runs on the \textit{Messiah} simulator, a customisable SC environment built on a sequential decision-making framework.  This section describes the mechanism of MOEAs, originally a single-objective algorithm adapted with weights for multiple objectives, and validates multi-objective RL, the \textit{Messiah} simulator, and the performance metrics used for result comparison.

\subsection{MOEA-Based Approach}

We evaluate a metaheuristic optimisation framework to serve as an additional baseline for the MORL methodology. We select NSGA-II among metaheuristic methods due to its superior performance (see Section~\ref{sec:solving_SC}) and its effectiveness in solution search due to its elitism strategy, which is beneficial in a large solution search space. Considered a MOEA-based approach, NSGA-II performs a non-dominated sorting of solutions combined with crowding distance to break the ties of solutions in the same front~\citep{deb_fast_2002}. Each objective is evaluated separately, thus circumventing the requirement for a scalarisation process.

\begin{figure}[h]
    \centering
    %\vspace{-10pt}
    \includegraphics[width= 9 cm]{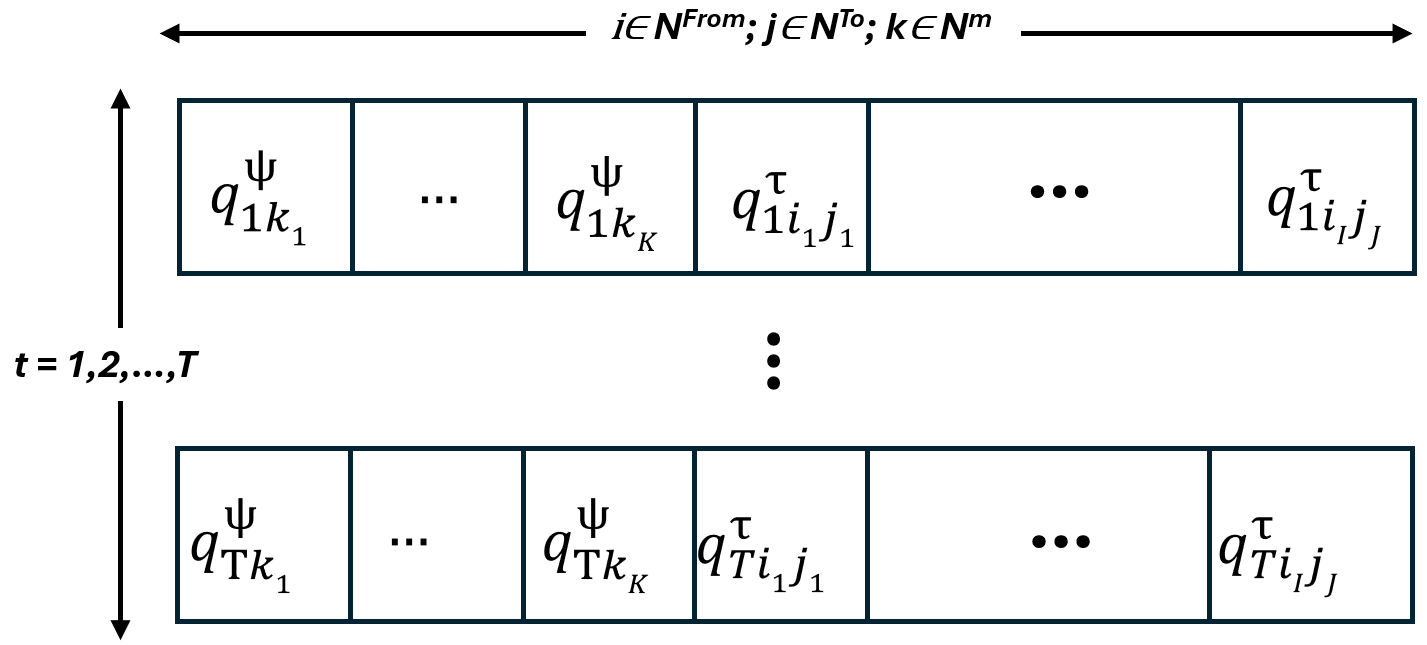}
    \caption{In NSGA-II, an individual solution is depicted by a vector with the size of total manufacturers and routes in the network for all period $T$ (i.e., $\text{decision variable vector size}=(T\cdot (|N^m|+|N^s|\cdot|N^m|+\dots+|N^d|\cdot|N^r|))$. Each chromosome gene signifies the production and delivery amounts $q_{tk}^{\psi}$ and $q_{tij}^{\tau}$, where $i=i_1, i_2,\dots,i_I, j=j_1,j_2,\dots,j_J, \text{and } k=k_1,k_2,\dots,k_K$ for each period $t$. Mutation and crossover apply to these solutions.}
    \label{fig:nsga2_encoding}
\end{figure}

\paragraph{\textbf{NSGA-II solution representation.}} \label{sec:sol_nsgaII} This study encodes the decision variables for NSGA-II as a collection of decisions encompassing the manufacturing and delivery quantities in the entire SC network throughout the period horizon, illustrated in Figure~\ref{fig:nsga2_encoding}. The algorithm commences by randomly initialising a parent population comprising potential solutions. Each solution is evaluated on the basis of multiple objective functions. Subsequently, NSGA-II performs non-dominated sorting based on two aspects: the number of solutions dominating it and the set of solutions it dominates. This step groups individual solutions into several fronts and ranks the fronts based on proximity to the PF. To ensure the diversity of solutions, the crowding distance is calculated within each front. Solutions with higher crowding distance values are assigned higher rankings. Genetic operations such as crossover and mutation are also executed to generate an offspring population. This offspring population is then merged into the initial parent population, resulting in a combined population with the size of both added populations. Non-dominated sorting and crowding distance calculations are performed on the combined population to guide truncation. The truncation reduces the combined population size to the original size of the parent population (as the population size remains constant throughout the optimisation). Then, this population is set as the new parent population. Finally, these procedures are iterated until the algorithm terminates (typically after a preset number of iterations). 

\subsection{MOMDP-Based Methods}

\begin{figure}[h]
    \centering
    \includegraphics[width=9 cm]{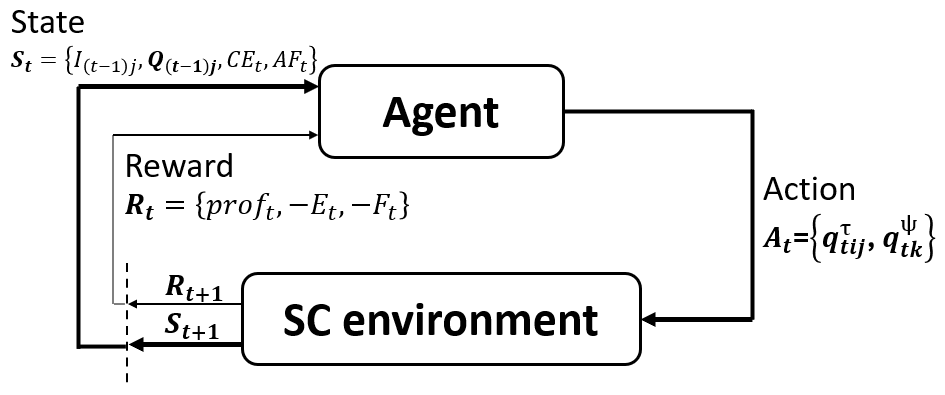}
    \caption{Agent-SC environment interaction along the time steps enabling more informed sequential decision making. The agent executes an action at state $t$, and in response, the environment provides both an observable state and a reward, serving as feedback to the agent. The agent then uses this information to decide its next action.}
    \label{fig:agent-env_interaction}
\end{figure}

In this paper, the multi-objective SC problem is constructed based on the MOMDP taxonomy to train the MORL/D~\citep{felten_multi-objective_2023} and weighted sum PPO algorithms~\citep{schulman_proximal_2017}. Unlike MOEA-based methods, MOMDP-based approaches support agent-environment interaction, enabling online learning and enhanced exploration, as shown in Figure~\ref{fig:agent-env_interaction}.

\begin{figure}[h]
    \centering
    \includegraphics[width= 9 cm]{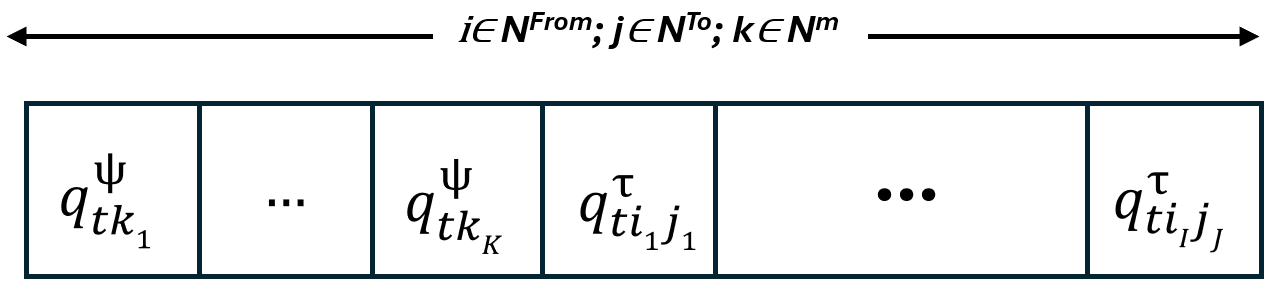}
    \caption{In a MOMDP, an action is represented by a vector whose size corresponds to the total number of manufacturers and routes in the network for each period $t$ (i.e., the size of the action space is given by $(|N^m|+|N^s|\cdot|N^m|+\dots+|N^d|\cdot|N^r|)$). The action is determined sequentially for each stage according to the policies.}
    \label{fig:momdp_encoding}
\end{figure}

\paragraph{\textbf{MOMDP-based solution representation.}} \label{sec:sol_momdp} The MOMDP framework simplifies solution representation by eliminating the time dimension due to its sequential nature. Actions, analogous to decision variables in optimisation contexts, encompass a range of manufacturing and delivery quantities throughout the supply chain network for each period, as illustrated in Figure~\ref{fig:momdp_encoding}. This action representation is similar in both weighted sum PPO and MORL/D.

\subsubsection{Weighted sum PPO}
\label{PPO}
%The reward function is a weighted sum of the reward functions used by MORL/D (with 20 different weight combinations being considered in this work). It will be interesting to understand how a population-based approach combined with the concept of non-dominance for dealing with multiple objectives (rewards) fares relative to a more standard RL approach that uses a single (weighted sum) reward function.

We modify a popular single-objective RL algorithm called PPO~\citep{schulman_proximal_2017} into a multi-objective method using scalarisation as a MOMDP-based baseline. This particular algorithm was selected from single-objective RLs because it offers both stability and efficiency, essential traits for training in intricate environments. This algorithm operates by optimising a surrogate objective function through stochastic gradient descent. It simplifies the predecessor method, Trust Region Policy Optimisation~\citep{pmlr-v37-schulman15}, which utilises second-order optimisation and hard constraints on the policy update step size. Instead, it clips the surrogate objective function to confine policy updates, ensuring stability in the policy search. PPO utilises an actor-critic architecture to maintain its stability in policy search. The actor maps the policies to define actions given the states, whereas the critic estimates the value functions. This algorithm exhibits more efficient sample use among the single-objective algorithms, referring to less interaction between the agent and the environment to learn an effective policy. This efficiency is desirable, particularly when training an agent in a complex environment, such as a multi-echelon, multi-facility SC network with non-stationary markets, where the agent training is time and resource-consuming.

Tailoring with the architecture of the algorithm, we scalarise the multiple objectives into a single-objective function by utilising 21 pre-defined weight vectors to produce adequate candidates for the PF approximation set, whose sum values range from 0 to 1. It is determined by balancing between the potential diversity and coverage of the set of solutions and the computational resources needed. Furthermore, the earlier multi-objective SC research conducted by~\citet{wang_efficiency_2020}, which matched the complexity levels (that is, the number of hierarchy and facilities are similar to this study), produced approximately 12 to 20 solutions in their PF approximation sets. These pre-defined weights are generated using the Das-Dennis method~\citep{das_normal-boundary_1998}, which is grounded upon uniform distribution to encourage diversity and even density in the resulting PF approximation set. Let $w_b$ be the weight of the objective $b$ ($b=1, 2, 3$), and let $R_{bt}$ be the reward of profit, negative GHG emission, and negative SL inequality sequentially, the single reward function that is a weighted sum value of multiple objectives is:
\begin{equation}
    R_t = \sum_b{w_{b} \cdot R_{bt}} \cdot \rho.
    \label{eq:rl_reward_vector}
\end{equation}
We normalise the rewards in advance to ensure a balanced preference among them, using the maximum and minimum possible values of each objective obtained by running every single objective separately (i.e., similar to when we set each weight to 1, while the rest are set to 0). Then, the algorithm runs and iterates at each weight combination, resulting in 21 solutions. Afterwards, the PF approximation is constructed considering all the solutions generated using Equation~\eqref{eq:pareto_front} where $\mathbf{\Pi}$ is defined as all the possible policies associated with the pre-defined weights.

\subsubsection{MORL/D algorithm}\label{MORL_Mechanism}
The MORL/D algorithm~\citep{felten_multi-objective_2023} is selected as our main method for its compatibility with the multi-dimensional action space in this study, whereas most MORL algorithms are designed for single-action settings. Inspired by multi-objective optimisation based on the decomposition concept (MOO/D)~\citep{felten_multi-objective_2023}, MORL/D decomposes multi-objective problems into scalarised subproblems, solving them as single-objective tasks. For simplicity, the scalarisation function is assumed to be linear. Its parallel problem-solving mechanism enables efficient operation in complex systems. MORL/D follows a multiple-policy approach, which is designed for scenarios where user preferences are unknown, aiming to generate trade-off solutions rather than a single decision. The algorithm initialises a population of policies, each assigned weight vectors and reference points, and evaluates their performance. The best-performing policies are retained in an external archive, while dominated solutions are pruned. At each iteration, a subset of policies is selected, interacts with the environment, and updates its experience buffer. These experiences are then used for policy improvement via gradient-based RL updates. The PF approximation is updated dynamically based on dominance criteria, ensuring non-dominated solutions are preserved, while weight vectors and reference points adapt to guide exploration toward under-represented regions.

To facilitate policy improvement, MORL/D leverages the multi-objective soft actor-critic (MOSAC) algorithm~\citep{chen_combining_2020}, which enables efficient learning in continuous action spaces. MOSAC extends the soft actor-critic framework to multi-objective settings by incorporating entropy-regularised exploration, which encourages policy diversity by preventing premature convergence to suboptimal solutions. This mechanism balances exploitation and exploration, ensuring that the agent effectively navigates trade-offs among competing objectives. In addition, MOSAC improves sample efficiency and stabilises learning through an off-policy approach, leading to more reliable policy updates. By integrating MOSAC, MORL/D efficiently optimises policies while maintaining a well-distributed PF approximation set, supporting effective multi-objective decision-making.

To enhance learning efficiency, MORL/D incorporates knowledge exchange between neighbouring subproblems. The Shared Buffer (SB) mechanism enables experience sharing across policies, improving sample efficiency and generalisation. Policies interact with the environment, storing state-action-reward tuples in a centralised buffer, which facilitates learning across subproblems. Policies are evaluated using scalarised returns, Pareto dominance, or hypervolume contribution, and the best-performing policies are retained in the external archive. The Pareto archive is iteratively updated, removing dominated solutions while ensuring diverse trade-offs. Cooperation mechanisms further enhance learning by enabling experience sharing among policies, increasing sample efficiency and performance. This iterative process continues until the convergence criteria are met, ensuring that the final PF approximation is diverse and optimal. By leveraging MOO/D decomposition, MORL/D effectively balances exploration and exploitation, producing well-distributed policies for multi-objective decision-making.

Unlike a weighted sum PPO, MORL/D does not require manually predefined weight vectors. Instead, the agent generates the weights dynamically during iterations. One of the weight adaptation methods is Pareto simulated annealing (PSA). PSA adjusts weights based on evaluations and proximity to PF approximation solutions. When the value in an objective vector of a solution $f_j(x)$ is equal to or lower than its nearest neighbour $f_j(x')$, the weight $\lambda^x_j$ increases slightly (multiplied by $\delta$, for example, 1.05). In contrast, when $f_j(x)>f_j(x')$, the weight $\lambda^x_j$ is reduced (divided by $\delta$). It encourages the decomposed single-objective values to move away from their neighbourhood solutions for broader exploration. Hence, this mechanism promotes diversity in the set of solutions~\citep{czyzzak_pareto_1998}.

\subsubsection{\textit{Messiah}: A Customisable SC Simulator} \label{sec:messiah}
We train our RL-based algorithms using \textit{Messiah}, a Python-based simulation framework tailored for SC optimisation. \textit{Messiah} has been developed in-house by our industrial partner, Peak AI Ltd., and is specifically designed to support sequential decision making under uncertainty within SC networks. Built on top of the Gymnasium interface~\citep{Towers_Gymnasium}, \textit{Messiah} ensures compatibility with widely used RL libraries and training pipelines, thereby facilitating reproducible experimentation and comparative evaluation of RL algorithms in SC contexts.

The framework offers a high degree of configurability and scalability, allowing users to construct SC environments of varying structural and operational complexity. Users can define detailed network topologies that include multiple echelons (e.g. suppliers, manufacturers, warehouses, and retailers) by specifying nodes, edges, and associated parameters. Each node can be designated as either internally controlled or externally governed. For controlled nodes, users are required to initialise inventory levels and define relevant attributes such as cost and emission coefficients. Demand at terminal nodes (e.g., markets) can be instantiated via statistical distributions or imported from external datasets. Similarly, edge attributes must be specified, including control status, processing logic (with optional input–output relationships), lead times, and per-time-step operational costs. Based on these specifications, \textit{Messiah} dynamically generates the SC network environment, which serves as the simulation ground for training and evaluating RL agents.

Unlike traditional SC optimisation tools, which typically rely on static or aggregate decision models, \textit{Messiah} supports dynamic, agent-based interactions over time, enabling more realistic modelling of decentralised and adaptive decision making. To support experimentation, the framework includes a library of utility functions and SC baseline policies, allowing researchers to implement, modify, and benchmark policy variants efficiently. Additionally, \textit{Messiah} exposes internal state histories, such as time-series records of node-level inventory levels, enabling comprehensive performance diagnostics. This capability is particularly valuable for conducting "what-if" scenario analyses and testing the operational robustness of learnt policies under various demand and supply perturbations, all within a controlled simulation environment that poses no risk to real-world operations.

\textit{Messiah} is actively versioned, with recent releases reflecting major performance and feature improvements. The current release (v2.2.0) builds on a substantial redesign introduced in version 2.0, which significantly improved runtime performance by restructuring core data into NumPy~\citep{harris_array_2020} arrays and applying vectorised operations along with Numba~\citep{lam_numba_2015} just-in-time compilation. These changes improved simulation efficiency and scalability. The framework is currently deployed in live use cases by several large B2B manufacturers, reinforcing its industrial relevance and maturity. Looking ahead, planned extensions include simulating market and pricing dynamics and enabling agent-based scenario planning, broadening its applicability to strategic decision-making contexts.

\subsection{Performance Measurements}\label{sec:performance_measurement}
After conducting the experiments on \textit{Messiah}, we employ various metrics to evaluate and compare the performance of different algorithms. In multi-objective optimisation, multiple optimal solutions are generated rather than a single value. The quality of the solution is assessed by convergence and diversity~\citep{branke_multiobjective_2008}. Convergence measures proximity to the true optimum, while diversity assesses the coverage of the solution space. Thus, in this research, several performance metrics are chosen to work in tandem to assess both quality dimensions, comprising:
\begin{itemize}
    \item Hypervolume~\citep{goos_multiobjective_1998}: This measures the volume between each solution on the PF approximation and a reference point. Hypervolume reflects both convergence and diversity and is valued for being monotonic (better solution sets are valued higher). Moreover, it always respects the Pareto dominance relation, which affirms it to be a reliable performance indicator~\citep{obayashi_hypervolume_2007}.
    \item Expected utility metric (EUM)~\citep{zintgraf_quality_2015}: EUM calculates the expected value of the highest utility on all weights in the set $\Lambda$, as shown in Equation~\eqref{eq:EUM}. This indicator is used for utility-based algorithms and considers user preferences. NSGA-II does not use this measure, as it is based on crowding distance. Given that the utility function is a function of the value vector $\mathbf{v^{\pi}}$ and the weight vector $\mathbf{w}$, the EUM of the front $\mathcal{F}$ is:
    \begin{equation}
    \label{eq:EUM}        EUM(\mathcal{F})=\underset{\mathbf{w}\in\Lambda}{\mathbb{E}}\left[\underset{\mathbf{v^\pi}\in\mathcal{F}}{\text{max }}u(\mathbf{v^\pi},\mathbf{w})\right].
    \end{equation}

    \item Sparsity~\citep{pmlr-v119-xu20h}: This calculates the average squared distance between adjacent solutions $\widetilde{P}_j(i)$ and $\widetilde{P}_j(i+1)$ in the front set $\mathcal{F}$ as given in Equation~\eqref{eq:sparsity}, where $j$ represents the objective index and $i$ denotes the solution index. Sparsity indicates the density of the solution coverage area and could be a complementary measurement to the hypervolume to examine the robustness of the model, as formulated below:
     \begin{equation}
        \label{eq:sparsity}
        S(\mathcal{F})=\frac{1}{|\mathcal{F}|-1}\sum^{m}_{j=1}\sum^{|\mathcal{F}|-1}_{i=1}\left(\widetilde{P}_j(i)-\widetilde{P}_j\left(i+1\right)\right)^2.
    \end{equation}
    \item Average Hausdorff Distance (AHD)~\citep{schutze_using_2012}: It denotes the maximum value obtained from the average generational distance ($GD_p$) and the average inverted generational distance ($IGD_p$). The number of solutions in the PF approximation set influences both the hypervolume and sparsity. AHD provides a more equitable comparison between sets with varying solution numbers by utilising average distances. Each distance is computed between each solution of the resultant PF approximation sets and the true PF set. In this research, the true PF is estimated by merging all resultant solution sets across all algorithm runs and then selecting the best PF approximation set using Equation~\eqref{eq:pareto_front}. Let $N$ be the solution number of the front set $X$ and $M$ be the solution number of the true PF set $Y$, while $p$ is the norm parameter (typically set to 2 for the Euclidean distance), hence the AHD between both fronts ($\Delta_p(X,Y)$) is given in Equation~\eqref{eq:AHD} below:
    \begin{equation}
        \label{eq:AHD}
        \begin{split}
        \Delta_p(X,Y)&=max(GD_p(X,Y),IGD_p(Y,X))\\
        &=max\Bigg(\bigg(\frac{1}{N}\sum^N_{i=1}{dist(x_i,Y)^p}\bigg)^{\frac{1}{p}},
        \bigg(\frac{1}{M}\sum^M_{i=1}{dist(y_i,X)^p}\bigg)^{\frac{1}{p}}\Bigg).
        \end{split}
    \end{equation}
\end{itemize}
In addition, we also evaluate other indicators, such as computational time, visual assessment of the PF approximation sets, and operational performance, which is observation of the state and behaviour when implementing the decisions.

\section{Experiments and Results}\label{sec:experimentation}
This section discusses the preliminary implementation that has been done so far.
\subsection{Implementation Details}
All experiments are carried out using Python 3.11.2 on the JupyterLab platform with 128GB RAM and a 16-core CPU. There have been 810 total runs overall. This section provides details of the setup used during the experiments.

\subsubsection{SC Environment Setup} \label{sec:SC Environmental Setup}
\begin{figure}[h]
    \centering
     
    \begin{subfigure}{0.45\textwidth}
    \includegraphics[width=\textwidth]{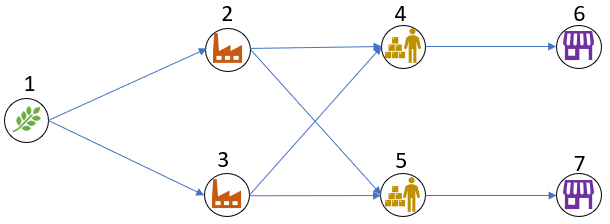}
    \caption{  Simple SC network}
    \end{subfigure}
    \hspace{0.05\textwidth}
    \begin{subfigure}{0.45\textwidth}
    \includegraphics[width=\textwidth]{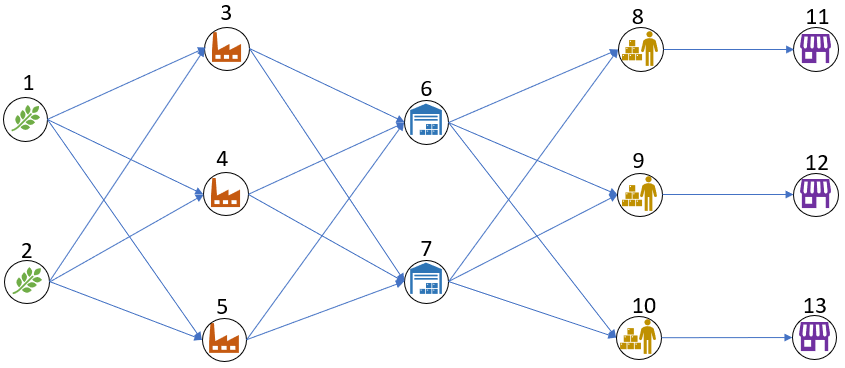}
    \caption{  Moderate SC network}
    \end{subfigure}
    
    \par\bigskip
    
    \begin{subfigure}{0.6\textwidth}
    \includegraphics[width=\textwidth]{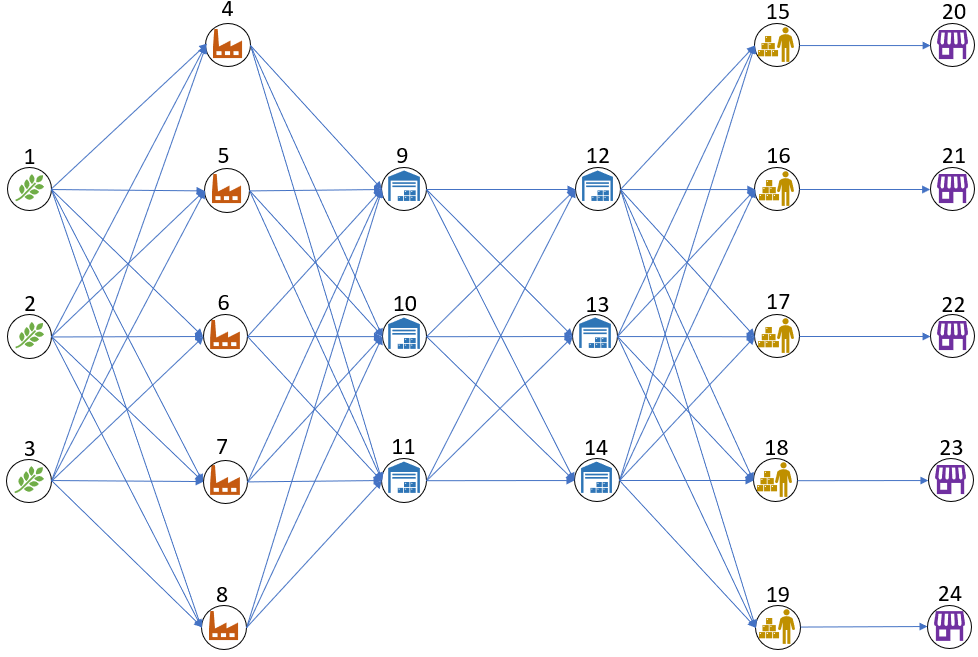}
    \caption{  Complex SC network}
    \end{subfigure}
  \caption{  The three examples of multi-echelon SC networks simulated in this work. The network starts with sourcing raw materials from suppliers $N^s$, which are transported to manufacturers $N^m$, which in turn manufacture and store the products before delivering them to either the distributors $N^w, N^d$ or directly to the retailers $N^r$. The retailers $N^r$ then will receive the products and sell them on to different markets $N^z$. The demand is subject to non-stationarity.}
  \label{fig:sc_network}
\end{figure}

To validate our generalised MOMDP model formulated in Section~\ref{sec:MOMDP-Based SC Optimisation}, we simulate experimental case studies through three SC problems, which are simple, moderate, and complex problems. Three example networks are presented in Figure~\ref{fig:sc_network} for improved clarity. To illustrate the complexities, in the optimisation-based model (see Section~\ref{sec:optimisation_based}), our problems result in 800, 2100, and 5900 decision variables. Meanwhile, in the MDP-based structure (see Section~\ref{sec:MOMDP-Based SC Optimisation}), our SC environment employs 20, 49, and 131-dimensional observation spaces and 8, 21, and 59-dimensional action spaces for simple, moderate, and complex SC environments consecutively. The SC echelon-related sets are given in the Table~\ref{tab:sets}. This table distinguishes the network difference between our problems.

The SC problems presented in this paper follow some pre-defined parameter setups. All networks are 'fully connected', meaning that every node in one layer is linked to every node in the subsequent layer. The supply of material from suppliers is assumed to be unlimited. The demand from all markets is non-stationary, whose fluctuation from the first and second markets are given in the normal distribution ($\mu=150, \sigma=60$ and $\mu=100, \sigma=40$, respectively), while the third, fourth, fifth markets follow the Poisson distribution ($\lambda=200, 100, 150$ respectively). All demand distributions are multiplied by the sinusoidal distribution to express seasonal fluctuation. Markets 1 and 2 are covered by a simple SC, a moderate SC includes markets 1, 2, and 3, and a complex SC encompasses all 5 markets.

A fixed price of $20$ is adopted for all simple SC markets, whereas variable pricing is implemented in moderate and complex SC markets: $\{20, 21, 20.5\}$ and $\{100, 101, 105, 103, 104\}$. Unfulfilled demand is considered a demand loss. This problem adheres to the MOMDP framework with a finite time horizon $T=100$ and a lead time $L=2$. The capacity for transportation ($Cap$) is 200, while the amount of manufacturing is not limited. Moreover, the values of the detailed parameters are given in the supplementary material. All methods use similar reference points, constituting slightly lower points than the minimum possible points in the hypervolume calculation, which are: $\{0,-2\times10^5,-100\}$, $\{0,-4\times10^5,-200\}$, $\{0,-1\times10^6,-500\}$, for simple, moderate, and complex SC, respectively.

\begin{table}[H]
\centering
\caption{ Sets and family of sets used in the formulation to distinguish between SC network problems. The following table lists the nodes covered in the sets and the sets covered in the family of sets in the three scenarios.}
\label{tab:sets}
    \begin{tabularx}{12cm}{llll}
        \toprule
        Notation & Simple SC & Moderate SC & Complex SC\\
        \midrule
        \textbf{Sets}\\
        $N^s$ & 1 & 1, 2&   1, 2, 3\\
        $N^m$ & 2, 3 & 3, 4, 5 & 4, 5, 6, 7, 8\\
        $N^{w1}$ & NA & 6, 7 & 9, 10, 11\\
        $N^d$ & NA & NA & 12, 13, 14\\
        $N^r$ & 4, 5 & 8, 9, 10 & 15, 16, 17, 18, 19\\
        $N^z$ & 6, 7 & 11, 12, 13 & 20, 21, 22, 23, 24\\
        \midrule
        \textbf{Family of sets}\\
        $N^{From}$ & $N^s, N^m$ & $N^s, N^m, N^w$ & $N^s, N^m, N^w, N^d$\\
        $N^{To}$ & $N^m, N^r$ & $N^m, N^w, N^r$ & $N^m, N^w, N^d N^r$\\
        $N^{Inv}$ & $N^r$ & $N^w, N^r$ & $N^w, N^d, N^r$\\
        $N^{Invf}$ & $N^m$ & $N^m, N^w$ & $N^m, N^w, N^d$\\
        $N^{Invt}$ & $N^r,N^z$ & $N^w,N^r,N^z$ & $N^w,N^d,N^r,N^z$\\
        \bottomrule
    \end{tabularx}
\end{table}

\subsubsection{PPO and MORL/D Setup}
The PPO agent is from the library Stable Baselines 3~\citep{stable-baselines3} while the MORL/D agent is adopted from MORL Baselines 1.0.0~\citep{felten_toolkit_2023}. Both are trained in the SC environments developed using the \textit{Messiah} simulator. Some normalisation procedures are performed on both agents to enhance the learning, as suggested in the Stable Baseline 3 documentation. The action spaces are normalised to the values between $-1$ and $1$, and the observation spaces are transformed into values between $0$ and $1$ when training the MORL and RL agents based on each minimum and maximum value.

Since PPO is a single-objective algorithm, the source code does not have a built-in reward normalisation method. When combined with the weighted sum to extend it into a multi-objective method, we normalise the rewards to ensure a balanced preference among them (see Section~\ref{PPO}).~\citet{schulman_proximal_2017} used 7 environments with 3 random seeds in each environment. In this study, upon 21 weight combinations, 10 random seeds are taken for each weight set, resulting in a total of 210 runs for each SC environment. Applying non-dominated sorting as per Equation~\eqref{eq:pareto_front} to solution sets yields 10 total PF approximation sets from all runs, each set comprising 21 solution points in an SC problem. The generalised advantage estimation (GAE) method is employed to achieve low variance while preserving low bias estimates. The GAE is calculated using the weighted average temporal difference residuals over the time steps. The advantage function is normalised during training to maintain stability and improve learning by preventing policy updates from being dominated by excessively large advantage values. The agent employs a multi-layer perceptron neural network architecture commonly used in actor-critic algorithms such as PPO.

The MORL/D algorithm code is built with a reward normalisation wrapper to ensure that all values in the reward vector are presented within the same scale. It is optimised according to the scalarised expected returns criterion~\citep{hayes_practical_2022} where the expectation value is first obtained from multiple runs and then used to calculate the utility. In this paper, the MORL/D algorithm uses MOSAC~\citep{chen_combining_2020}, which operates in two steps: initially applying the MORL/D method to learn a set of policies, subsequently using the evolutionary strategy for further search, based on previous policies. In training MORL/D, there are four scenarios applied: 1)~without weight adaptation method and the SB (MORL/D); 2)~without weight adaptation method, with SB (MORL/D (SB)); 3)~with PSA weight adaptation method, without SB (MORL/D (PSA)); 4)~with PSA weight adaptation method and with SB (MORL/D (SB\&PSA). In a previous study, ~\citet{felten_multi-objective_2023} applied 10 random seeds to determine statistical robustness in each set of experiments. In this study, each scenario runs with 10 random seeds, resulting in 40 runs in an SC environment.

\begin{table}[h]
    \centering
    \begin{minipage}[t]{0.48\textwidth}
        \centering
        \captionof{table}{Hyperparameters of PPO Algorithm}
        \label{tab:hyperparameter_ppo}
        \begin{tabularx}{\textwidth}{ll}
            \toprule
            Hyperparameters & Values \\
            \midrule
            Learning rate & $3 \times 10^{-4}$ \\
            Number of steps & 2048 \\
            Minibatch size & 64 \\
            Number of epochs & 10 \\
            Discount factor & 0.99 \\
            GAE Weight & 0.95 \\
            Clipping range & 0.2 \\
            Entropy coefficient & 0 \\
            Value function coefficient & 0.5 \\
            Max gradient clipping & 0.5 \\
            Episode per iteration & 50 \\
            Steps per episode & 100 \\
            Total time steps & $5 \times 10^{5}$ \\
            Activation function & Tanh \\
            \bottomrule
        \end{tabularx}
    \end{minipage}%
    \hspace{0.03\textwidth}
    \begin{minipage}[t]{0.48\textwidth}
        \centering
        \captionof{table}{Hyperparameters of MORL/D Algorithm with MOSAC Policy}
        \label{tab:hyperparameter_MORLD}
        \begin{tabularx}{\textwidth}{ll}
            \toprule
            Hyperparameters & Values \\
            \midrule
            \textbf{MORL/D} & \\
            Discount factor & 0.995 \\
            Population size & 6 \\
            Exchange every & $5 \times 10^4$ \\
            Total time steps & $5 \times 10^5$ \\
            Neighbourhood size & 1 \\
            Update passes & 10 \\
            Initial weight dist. & uniform \\
            \midrule
            \textbf{MOSAC} & \\
            Buffer size & $10^6$ \\
            Discount factor & 0.99 \\
            Target smoothing coef. & 0.005 \\
            Batch size & 128 \\
            Steps before learning & $10^3$ \\
            Hidden neurons & [256; 256] \\
            Actor LR & $3 \times 10^{-4}$ \\
            Critic LR & $10^{-3}$ \\
            Actor training freq. & 2 \\
            Target training freq. & 1 \\
            Activation function & ReLU \\
            \bottomrule
        \end{tabularx}
    \end{minipage}
\end{table}

The hyperparameters listed in Tables~\ref{tab:hyperparameter_ppo} and~\ref{tab:hyperparameter_MORLD} are mainly based on the default settings of the RL libraries employed, selected to ensure stable training in all SC scenarios. For PPO, learning and stability parameters, such as learning rate, update steps, and clipping range, follow widely validated defaults to maintain reliable policy updates. Similarly, MORL/D used standard evolutionary settings for population size and update intervals, while the embedded MOSAC policy applied typical actor-critic configurations for off-policy learning. These settings supported consistent convergence and effective PF approximation across varying network complexities.

\subsubsection{NSGA-II Setup}
It is challenging to compare the RL algorithms and the optimisation-based methods due to their inherently different nature. For instance, in RL, algorithms are iterated over time steps to find the best policies that map actions and cumulative returns. Meanwhile, in NSGA-II, the algorithm looks for solutions directly over generations based on the objective value without the need to define policies. Nevertheless, the comparison is vital to examine the performance of the RL algorithms relative to commonly used methods in the multi-objective SC domain. Accordingly, the comparison is regulated on the best possible parameters that promote convergence with consideration of computational resources. We employ NSGA-II from the Pymoo package~\citep{blank_pymoo_2020} to solve our SC problem.

Table~\ref{tab:nsga_parameters} provides a summary of the hyperparameters utilised for experiments incorporating NSGA-II. The choice of distribution indices $\eta$ has a significant impact on the algorithm's search dynamics: lower $\eta$ values increase diversity by allowing larger movements in the solution space, whereas higher values focus on local searches by producing offspring closer to their parents~\citep{deb_fast_2002}. Likewise, raising the mutation probability boosts exploration but can destabilise convergence, while a higher crossover probability (as applied here) facilitates wide genetic recombination, advantageous for high-dimensional spaces. Although increasing population sizes might potentially improve solution quality, extensive empirical testing showed that values exceeding 300 consistently resulted in system-level memory problems, leading to premature termination of the Python kernel. Thus, a population size of 300 is the maximum feasible configuration within our computational constraints.

\begin{table}[h]
    \centering
    \caption{Hyperparameters of NSGA-II to solve the SC problem across three network complexities: simple, moderate, and complex.}
    \label{tab:nsga_parameters}
    \begin{tabularx}{7cm}{ll}
        \toprule
        Hyperparameters & Values\\
        \midrule
        Population size & 300\\
        Offspring number & 30\\
        Cross-over operator & binary (SBX)\\
        Cross-over probability & 90\%\\
        Cross-over $\eta$ & 15\\
        Mutation method & polynomial\\
        Mutation $\eta$ & 20\\
        \bottomrule
    \end{tabularx}
\end{table}

\subsection{Results \& Discussion} \label{sec:results}
This section presents the results of the experiments and discusses their implications. The analysis is broken down into overall performance in accordance with multi-objective optimisation performance indicators, visual examination of the PF approximation sets, and operational behaviour of implemented solutions.

\subsubsection{Overall Performance}
\label{sec:overall_performance}

\begin{figure}[H]
    \centering
    \begin{subfigure}{0.32\textwidth}
        \includegraphics[width=\textwidth]{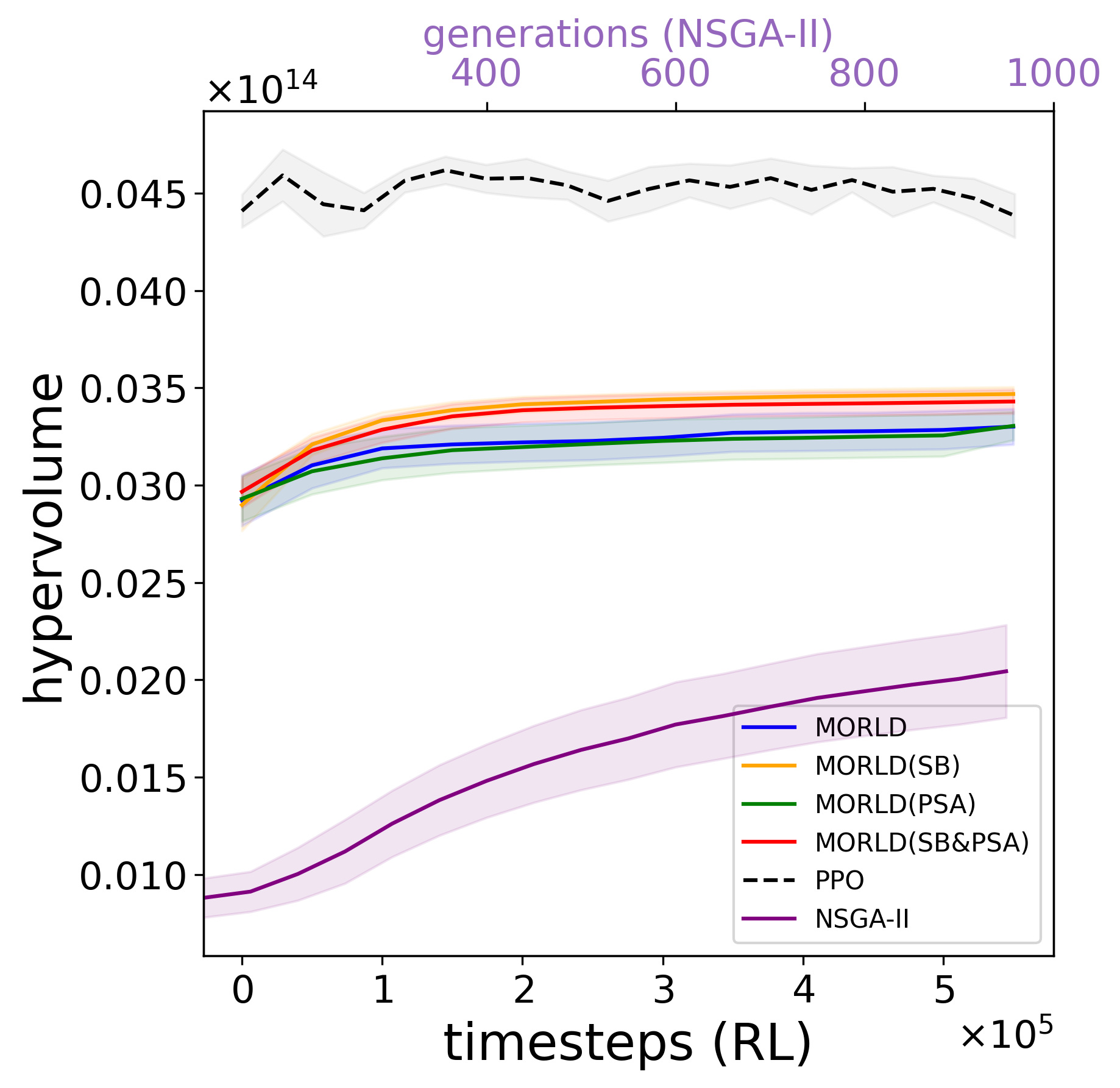}
        \caption{Hypervolume-Simple SC (max)}
        \label{fig:hypervolume_simple}
    \end{subfigure}
    %\hfill
    \begin{subfigure}{0.32\textwidth}
        \includegraphics[width=\textwidth]{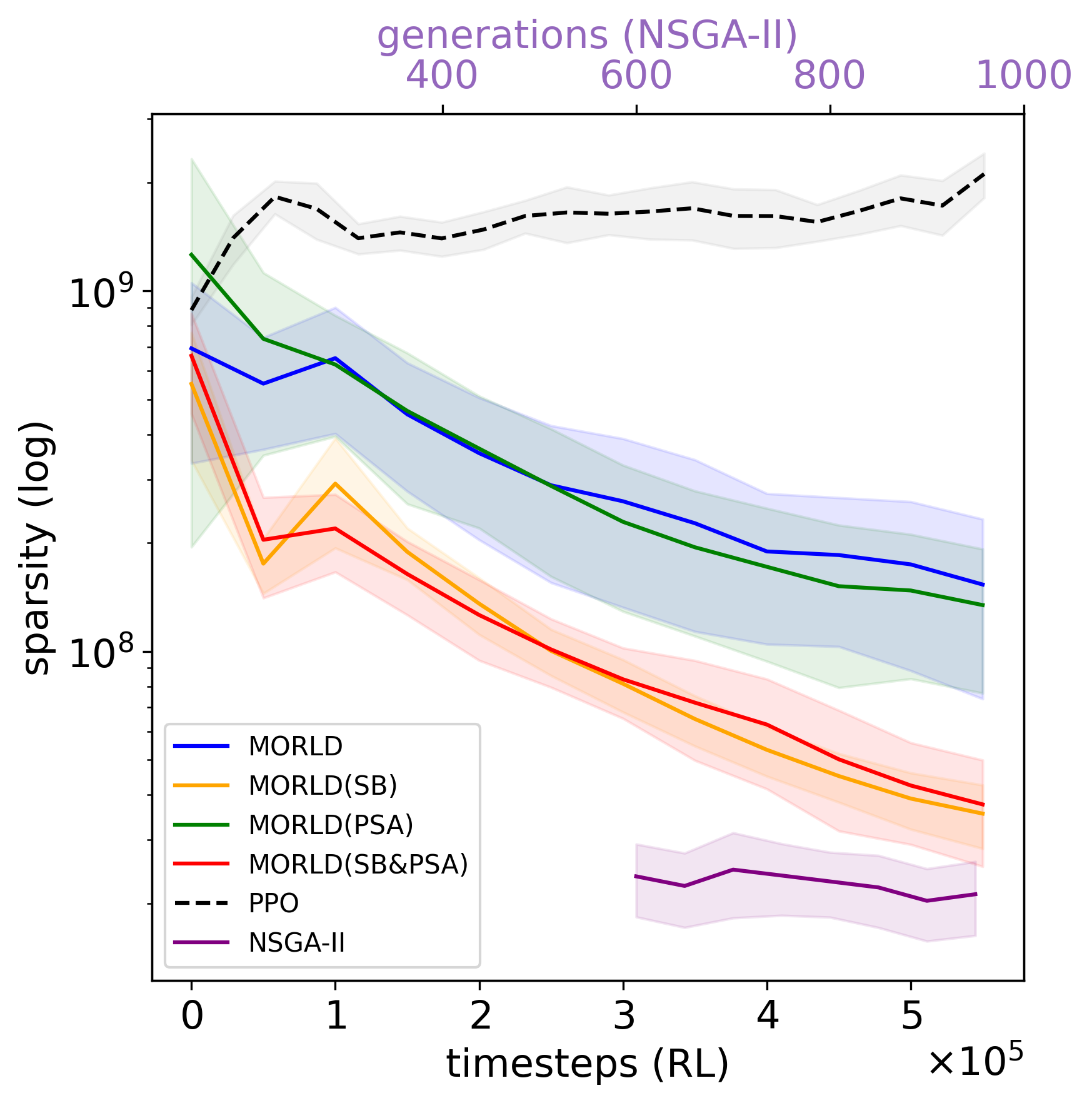}
        \caption{Sparsity-Simple SC (min)}
        \label{fig:sparsity_simple}
    \end{subfigure}
    %\hfill
    \begin{subfigure}{0.32\textwidth}
        \includegraphics[width=\textwidth]{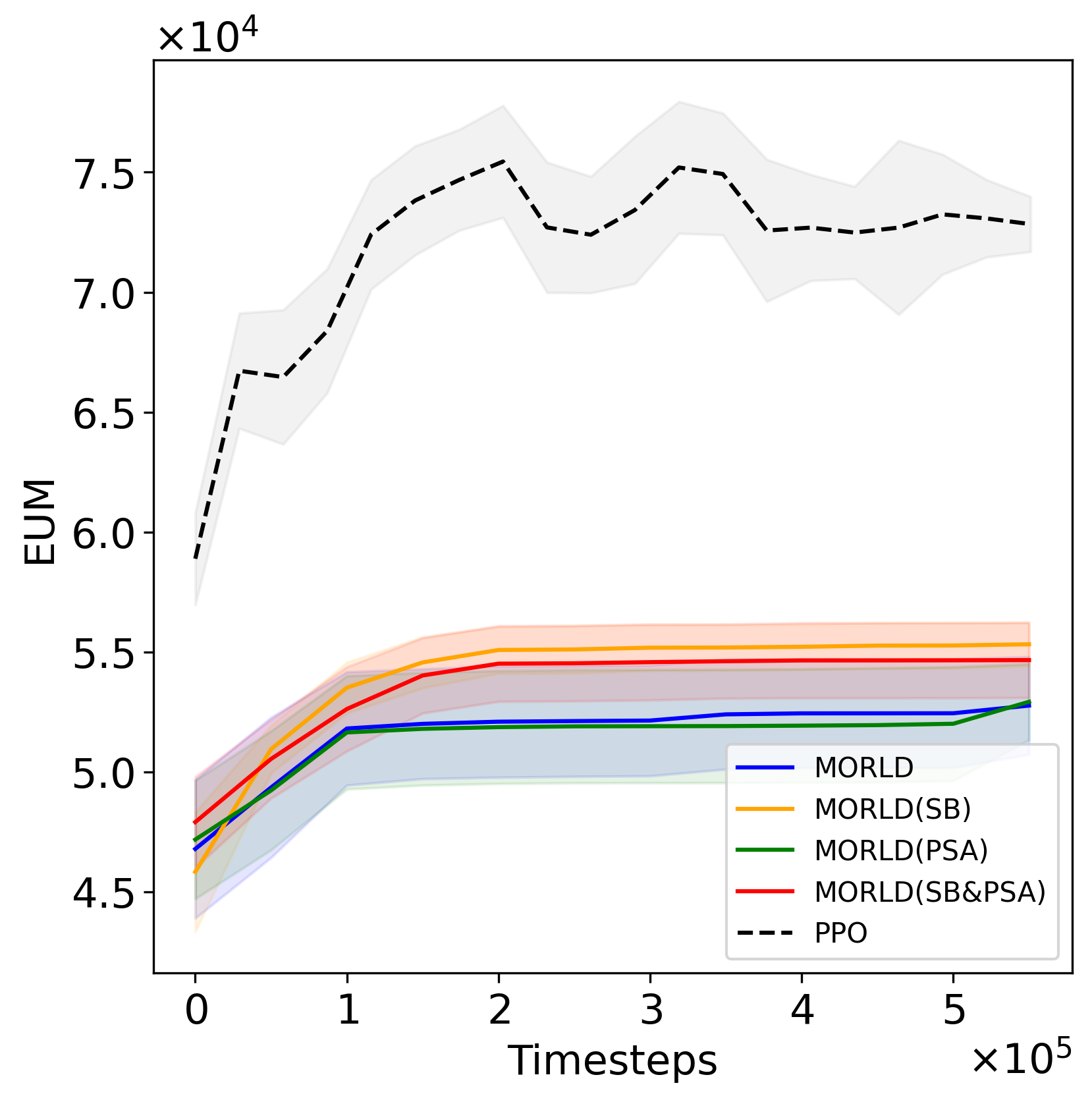}
        \caption{EUM-Simple SC (max)}
        \label{fig:eum_simple}
    \end{subfigure}

    %\hfill
    %\vskip\baselineskip
    
    \begin{subfigure}{0.32\textwidth}
        \includegraphics[width=\textwidth]{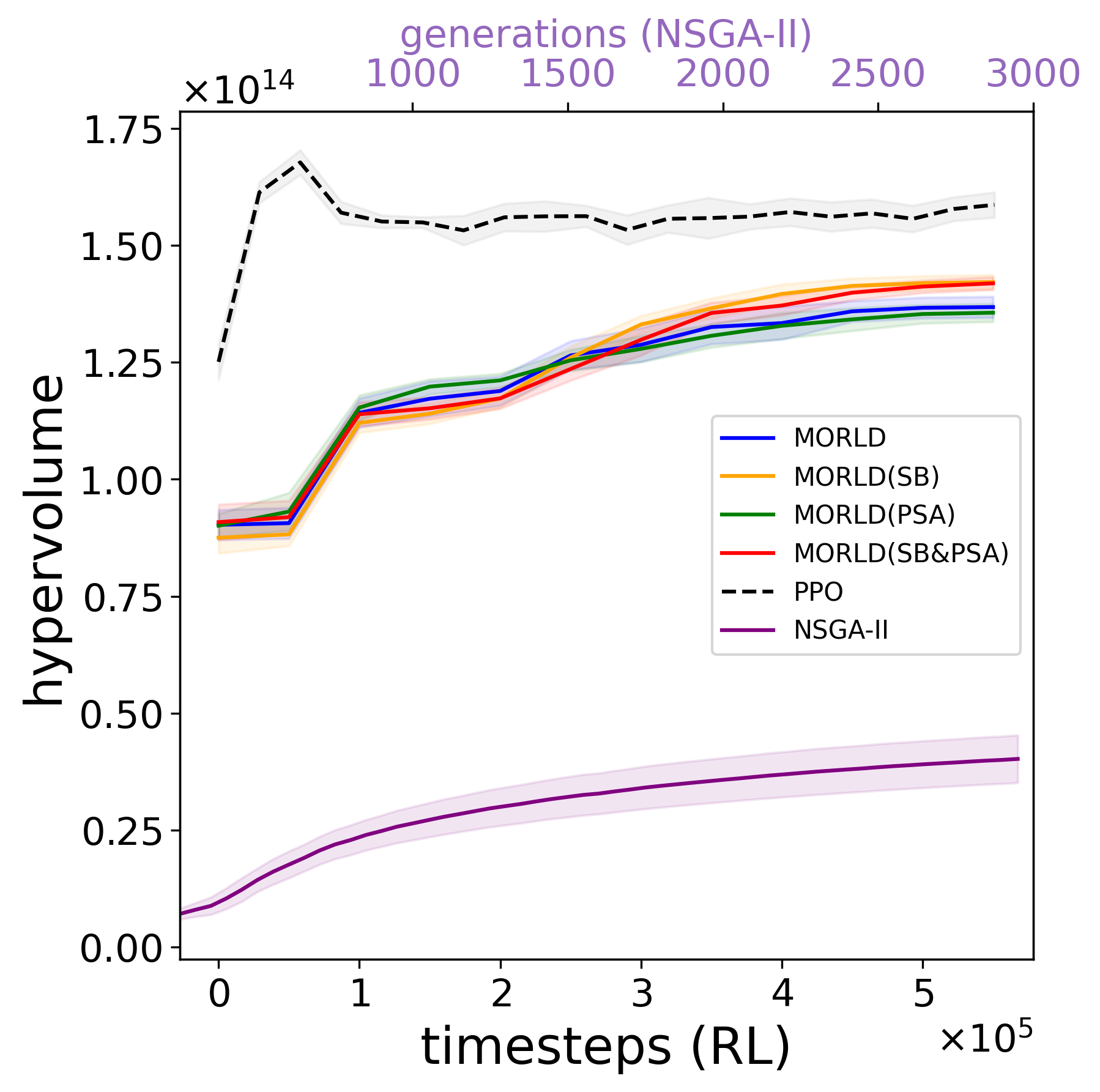}
        \caption{Hypervolume-Moderate SC (max)}
        \label{fig:hypervolume_moderate}
    \end{subfigure}
    %\hfill
    \begin{subfigure}{0.32\textwidth}
        \includegraphics[width=\textwidth]{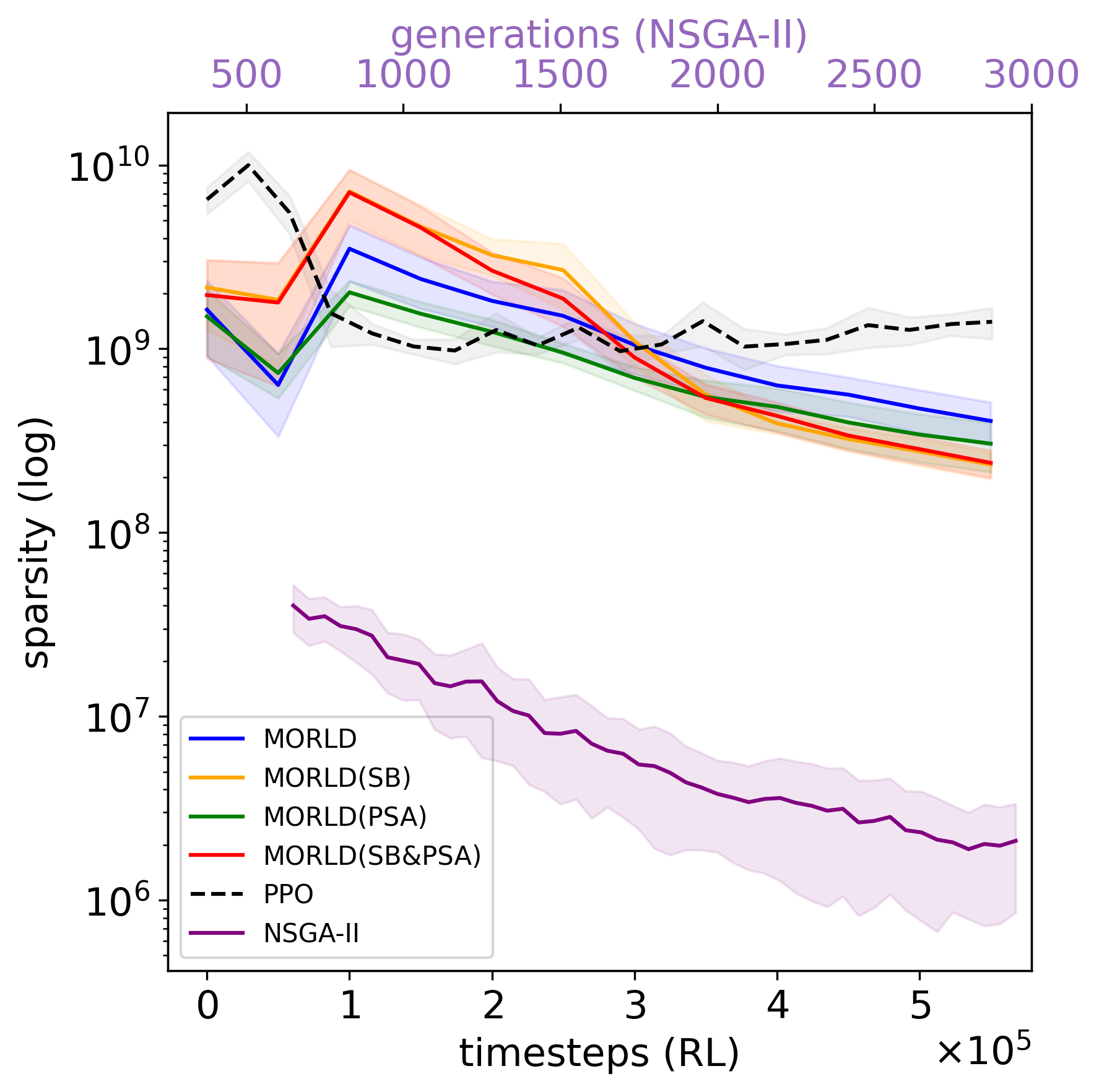}
        \caption{Sparsity-Moderate SC (min)}
        \label{fig:sparsity_moderate}
    \end{subfigure}
    %\hfill
    \begin{subfigure}{0.32\textwidth}
        \includegraphics[width=\textwidth]{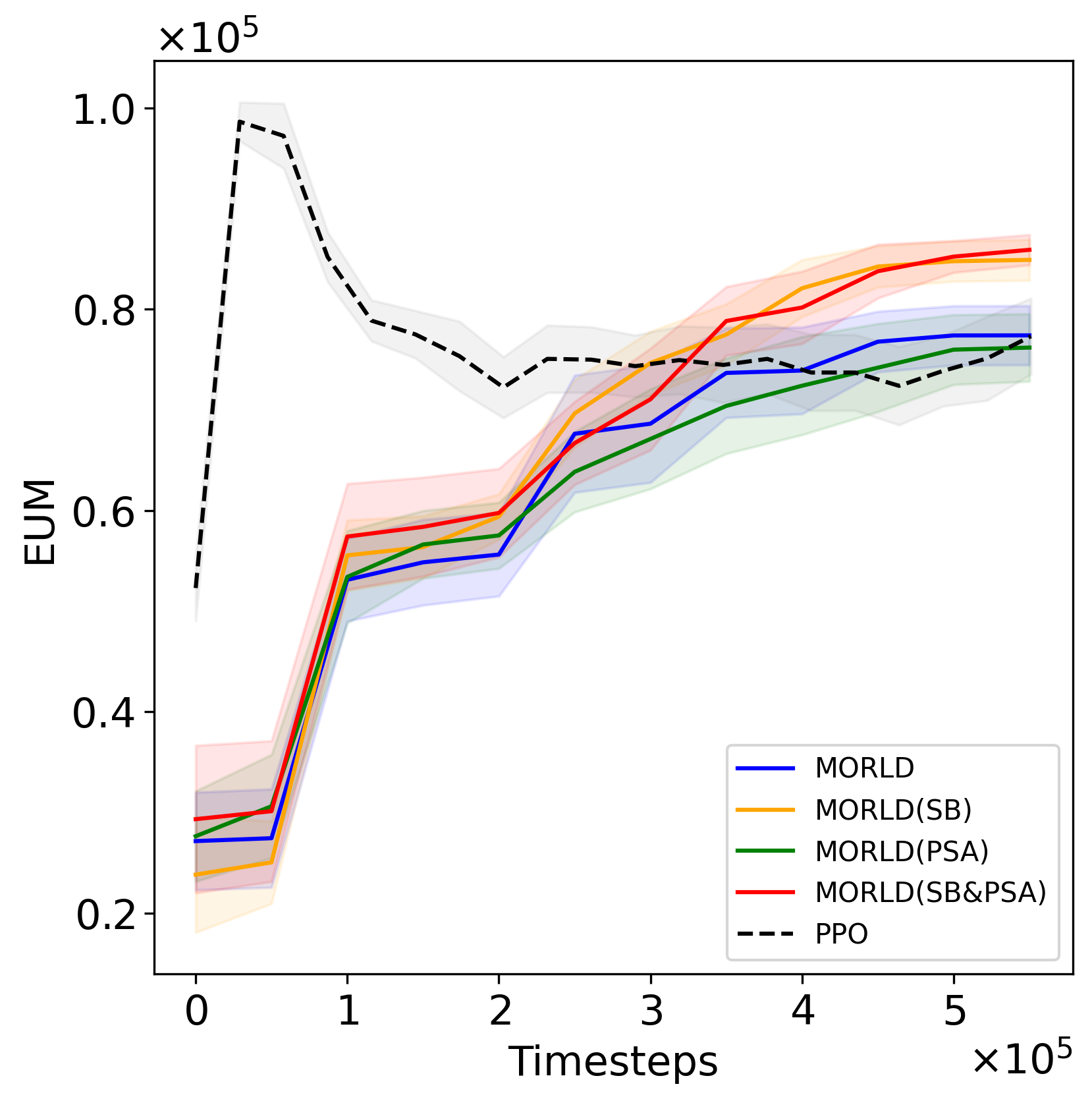}
        \caption{EUM-Moderate SC (max)}
        \label{fig:eum_moderate}
    \end{subfigure}

    \begin{subfigure}{0.32\textwidth}
        \includegraphics[width=\textwidth]{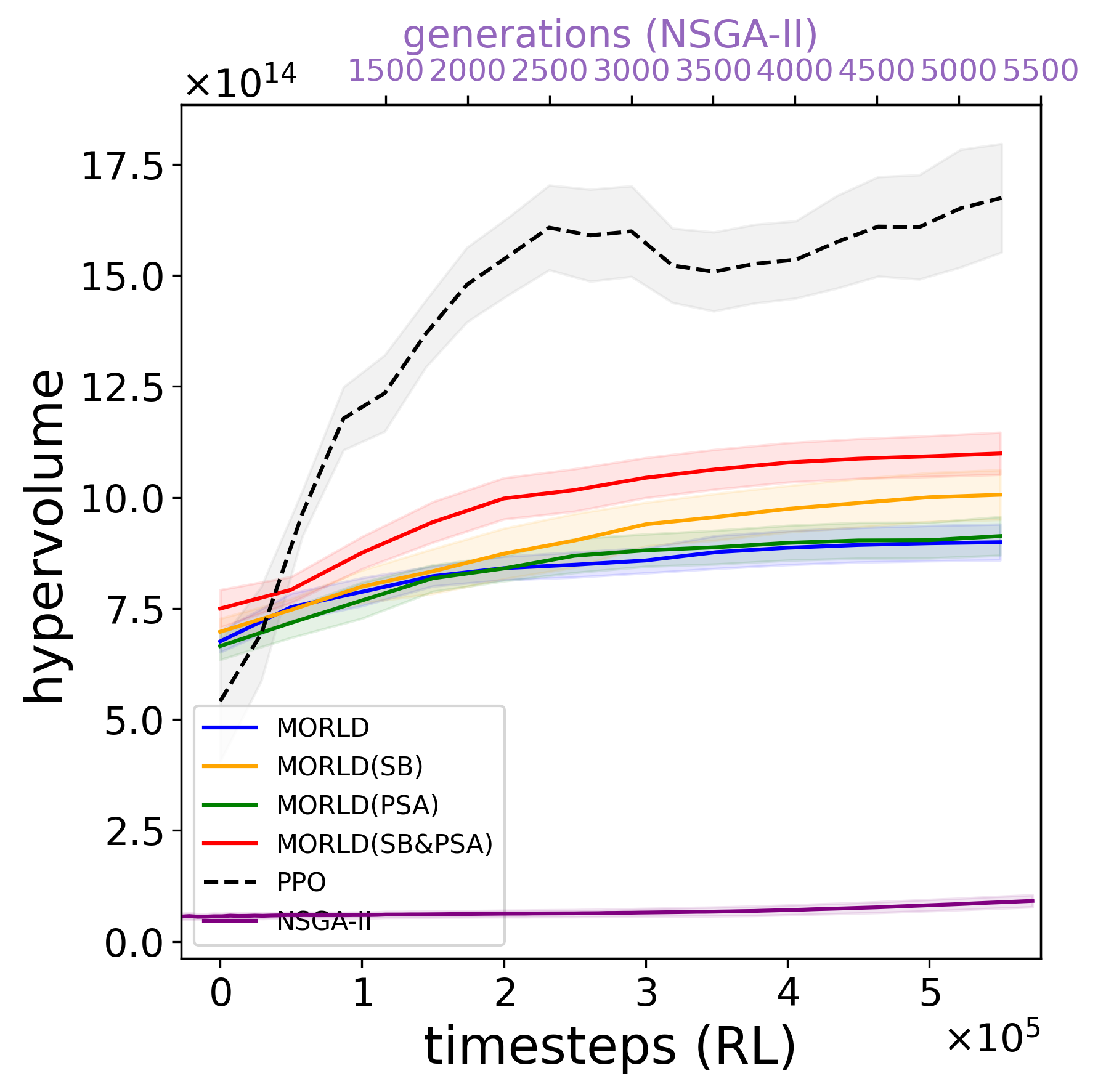}
        \caption{Hypervolume-Complex SC (max)}
        \label{fig:hypervolume_complex}
    \end{subfigure}
    %\hfill
    \begin{subfigure}{0.32\textwidth}
        \includegraphics[width=\textwidth]{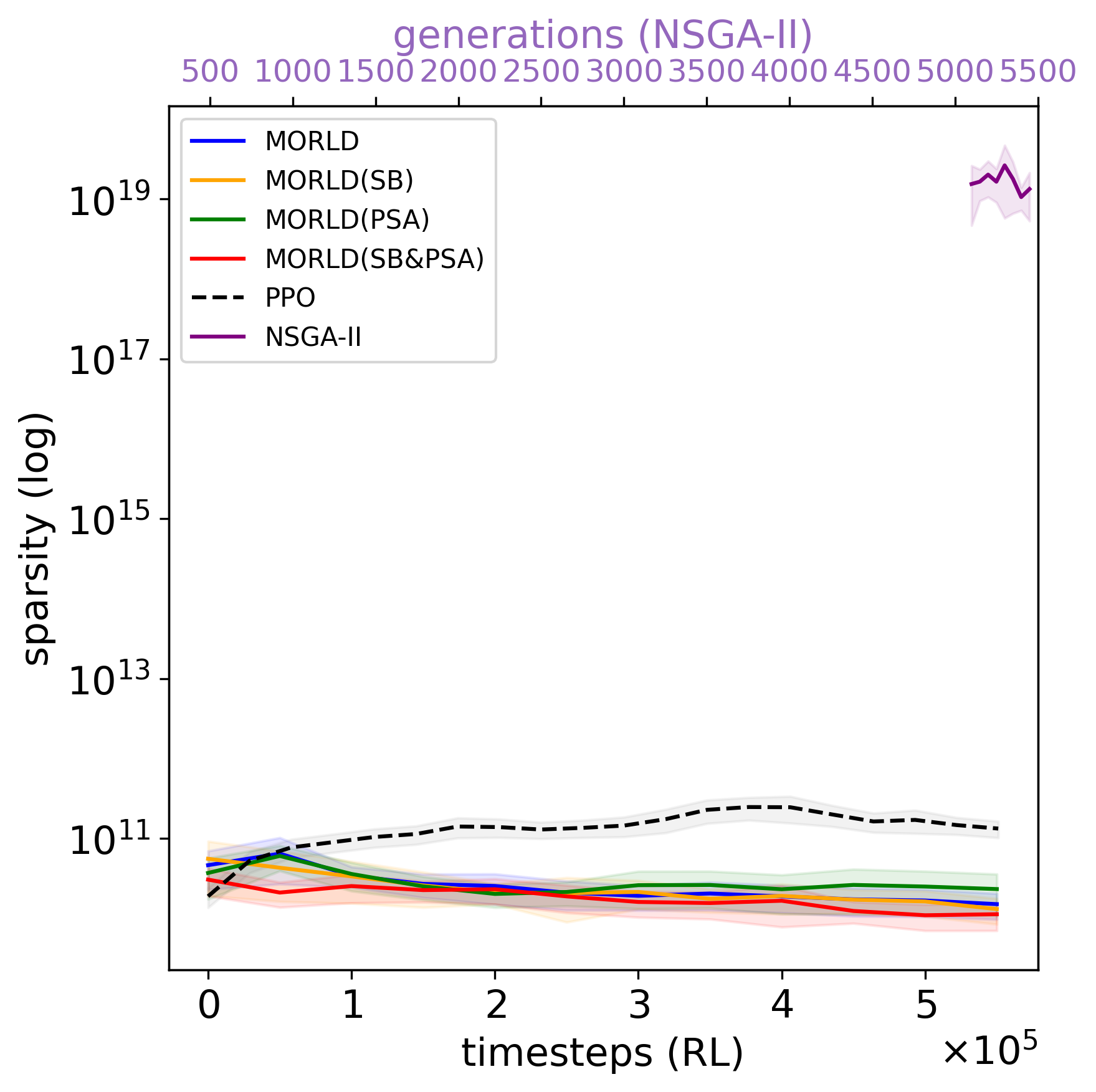}
        \caption{Sparsity-Complex SC (min)}
        \label{fig:sparsity_complex}
    \end{subfigure}
    %\hfill
    \begin{subfigure}{0.32\textwidth}
        \includegraphics[width=\textwidth]{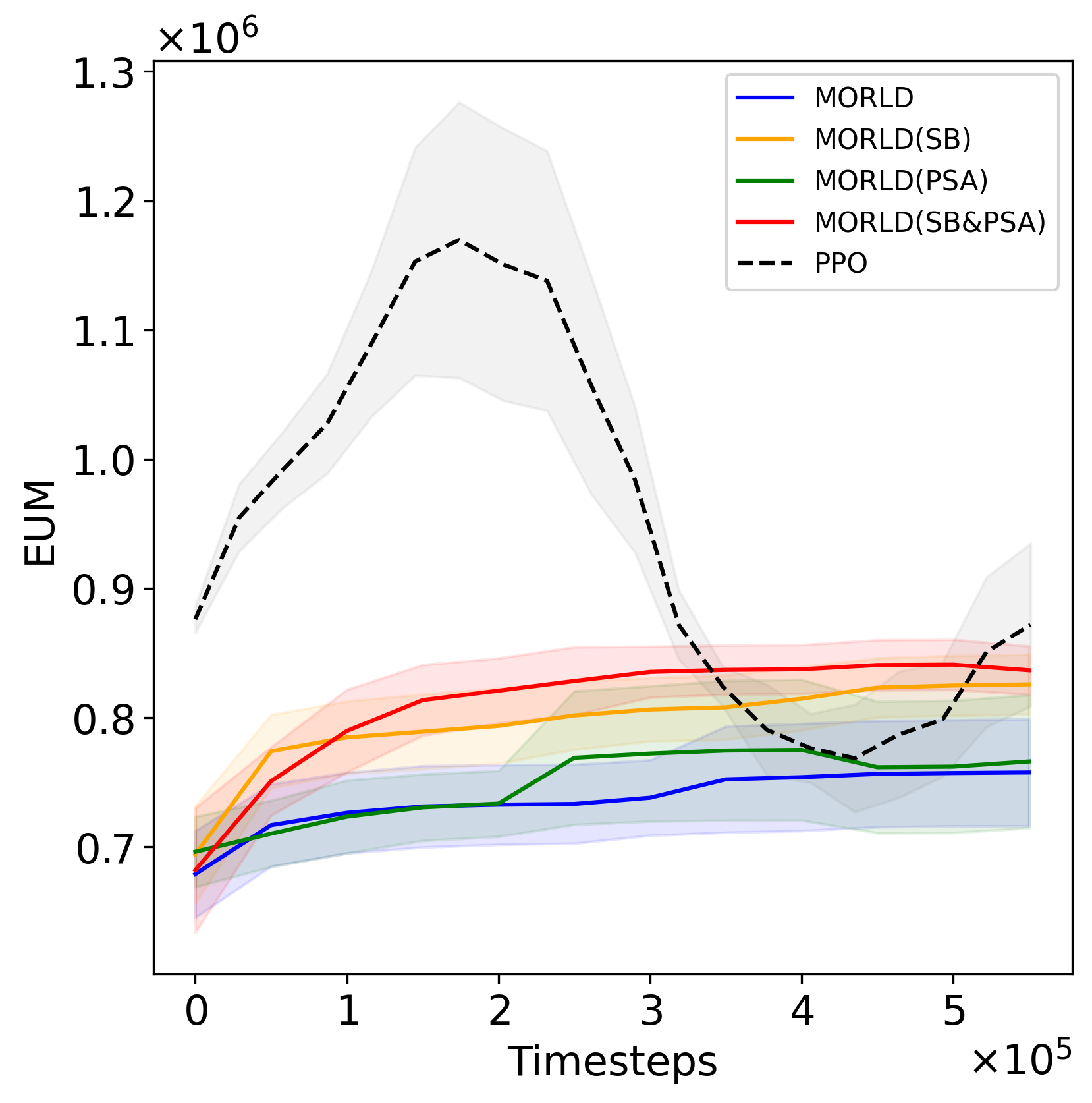}
        \caption{EUM-Complex SC (max)}
        \label{fig:eum_complex}
    \end{subfigure}
    \caption{The overall performance of the three methods. PPO consistently shows the highest hypervolume in simple, moderate, and complex SC environments, while MORL/D presents modest values and NSGA-II shows the least values (\ref{fig:hypervolume_simple},~\ref{fig:hypervolume_moderate},~\ref{fig:hypervolume_complex}). However, PPO's sparsity values remain much higher than MORL/D (\ref{fig:sparsity_simple},~\ref{fig:sparsity_moderate},~\ref{fig:sparsity_complex}). Additionally, the former's EUM is getting more unstable along the increased problem complexity(\ref{fig:eum_simple},~\ref{fig:eum_moderate},~\ref{fig:eum_complex}).}
    \label{fig:overall_performance}
\end{figure}

This section evaluates the performance of all results based on various metrics for multi-objective optimisation. Figure~\ref{fig:overall_performance} displays the hypervolume and sparsity for all algorithms and EUM specifically for RL-based techniques\footnote{EUM is pertinent solely to utility-based methods (refer to Section~\ref{sec:performance_measurement})}. Despite achieving the highest hypervolume values in all three scenarios (Figures~\ref{fig:hypervolume_simple},~\ref{fig:hypervolume_moderate},~\ref{fig:hypervolume_complex}), the PPO algorithm exhibits significant EUM fluctuations over iterations (Figures~\ref{fig:eum_simple},~\ref{fig:eum_moderate},~\ref{fig:eum_complex}). Its EUM for the complex SC problem reveals fluctuations, indicating difficulties in effectively balancing the three objectives according to the assigned weights (Figure~\ref{fig:eum_complex}). Moreover, it presents the sparsest solutions among the three methods, indicating reduced model robustness due to significant gaps between solutions (Figures~\ref{fig:sparsity_simple},~\ref{fig:sparsity_moderate},~\ref{fig:sparsity_complex}).

While PPO remains a strong baseline and performs reasonably well across various metrics, it exhibits limitations in effectively managing trade-offs in multi-objective settings. Specifically, the lack of a dynamic objective balancing mechanism in its normalisation process can lead to uneven improvement across objectives. For example, a 1-unit increase in the normalised scale may correspond to disproportionately large changes in raw objectives, such as 50,000 units in profit, 10,000 in emissions, and only 1 unit in inequality, resulting in a high hypervolume score that may be skewed by a single dominant objective. In multi-objective contexts, conflicting objectives introduce gradient interference during training, which can cause PPO policy updates to fluctuate and slow convergence. As a single-objective algorithm, PPO is inherently designed to maximise cumulative rewards based on normalised values, without explicitly accounting for trade-offs. Consequently, even with a weighted sum approach, the algorithm may struggle to consistently align with the assigned objective preferences over time, leading to observable dips in EUM values. This challenge is compounded in high-dimensional combinatorial settings, where the enlarged action and policy spaces further complicate learning stable policies. Although PPO’s clipping mechanism helps stabilise updates in single-objective tasks, it appears less effective in managing the nuanced dynamics of multi-objective problems. Furthermore, the use of static weights throughout the training contrasts with the evolving nature of the agent's policy, which could contribute to instability. This is reflected in increasing sparsity values toward convergence, suggesting that the approximated Pareto front lacks density and may reduce the robustness of the resulting solutions.

Conversely, the hypervolume values for NSGA-II consistently rise across all problem scenarios as the algorithm enhances the non-dominating rank and crowding distance of its solution sets (Figures~\ref{fig:hypervolume_simple},~\ref{fig:hypervolume_moderate},~\ref{fig:hypervolume_complex}). However, it demonstrates the lowest hypervolume in all cases, indicating that the solution sets lack both optimality and diversity. Furthermore, in the complex SC problem, the increase in hypervolume becomes less notable. Constraints and the vast number of decision variables (e.g., 5,900 variables) restrict the solution search space, reflecting an imbalance between the exploration-exploitation strategy in high-dimensional problems. The algorithm predominantly exploits the present solution space, which heavily relies on the initial population as parents of future generations, leading to less diverse solutions (Figures~\ref{fig:sparsity_simple},~\ref{fig:sparsity_moderate},~\ref{fig:sparsity_complex}). Furthermore, it encounters difficulties in generating solution options within the complex SC problem, as evidenced by undefined sparsity values up to approximately generation 5,000 (Figure~\ref{fig:sparsity_complex}). This limitation underscores NSGA-II's incompatibility with large-dimensional problems given current computational limitations.

MORL/D showcases the most thoughtful balance among algorithms with respect to optimality, diversity, and density for all SC problems. The hypervolume (Figures~\ref{fig:hypervolume_simple},~\ref{fig:hypervolume_moderate},~\ref{fig:hypervolume_complex}) and EUM (Figures~\ref{fig:eum_simple},~\ref{fig:eum_moderate},~\ref{fig:eum_complex}) of MORL/D consistently increase across the three SC problems. Furthermore, a decrease in sparsity indicates an increase in density (Figures~\ref{fig:sparsity_simple},~\ref{fig:sparsity_moderate},~\ref{fig:sparsity_complex}), which contributes to a better model robustness~\citep{pmlr-v119-xu20h}. This algorithm exhibits the most consistent learning over time compared to the others. In the four scenarios of the MORL/D application, the SB mechanism augments both the hypervolume and the EUM values while decreasing the sparsity for all SC problems. Meanwhile, the implementation of PSA weight adaptation is advantageous in more complex SC environments. The SB mechanism facilitates knowledge transfer between subproblems to improve the agent's learning process. The PSA mechanism simultaneously refines the weight update throughout the learning process, resulting in a better policy.

\begin{figure}[h]
    \centering
    \begin{subfigure}{0.32\textwidth}
        \includegraphics[width=\textwidth]{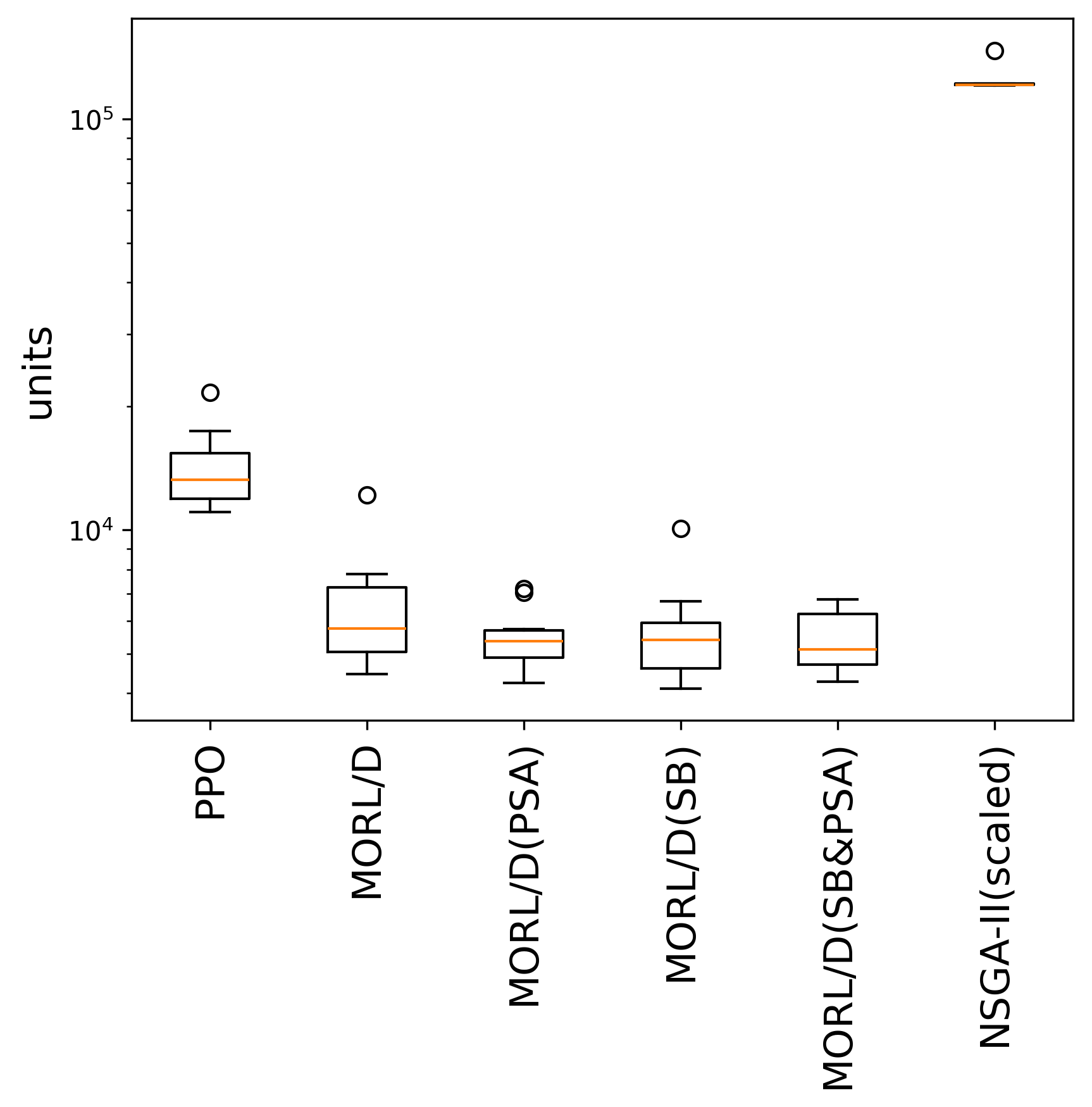}
        \caption{AHD-Simple SC}
        \label{fig:ahd_simple}
    \end{subfigure}
    \begin{subfigure}{0.32\textwidth}
        \includegraphics[width=\textwidth]{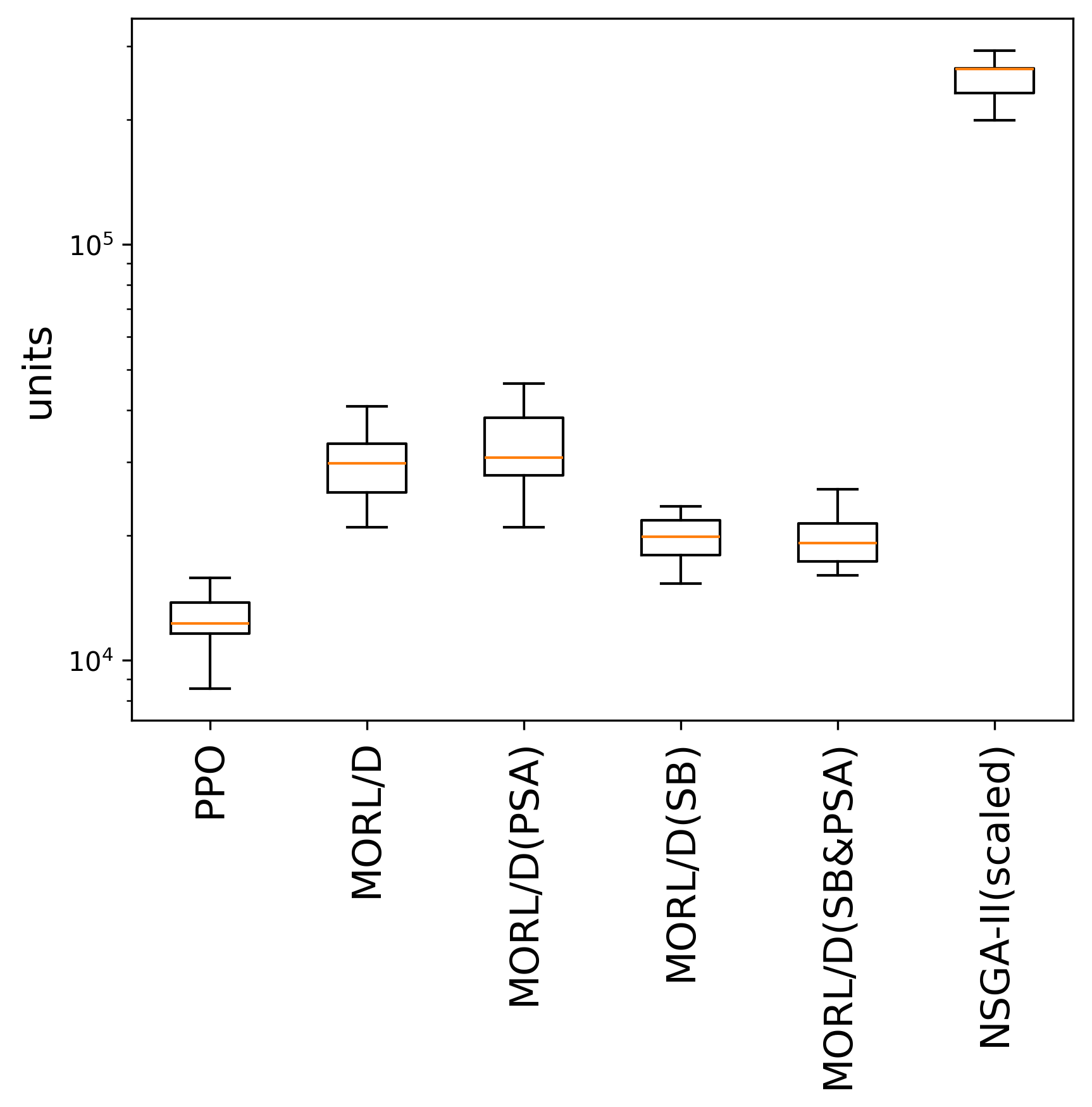}
        \caption{AHD-Moderate SC}
        \label{fig:ahd_moderate}
    \end{subfigure}
    \begin{subfigure}{0.32\textwidth}
        \includegraphics[width=\textwidth]{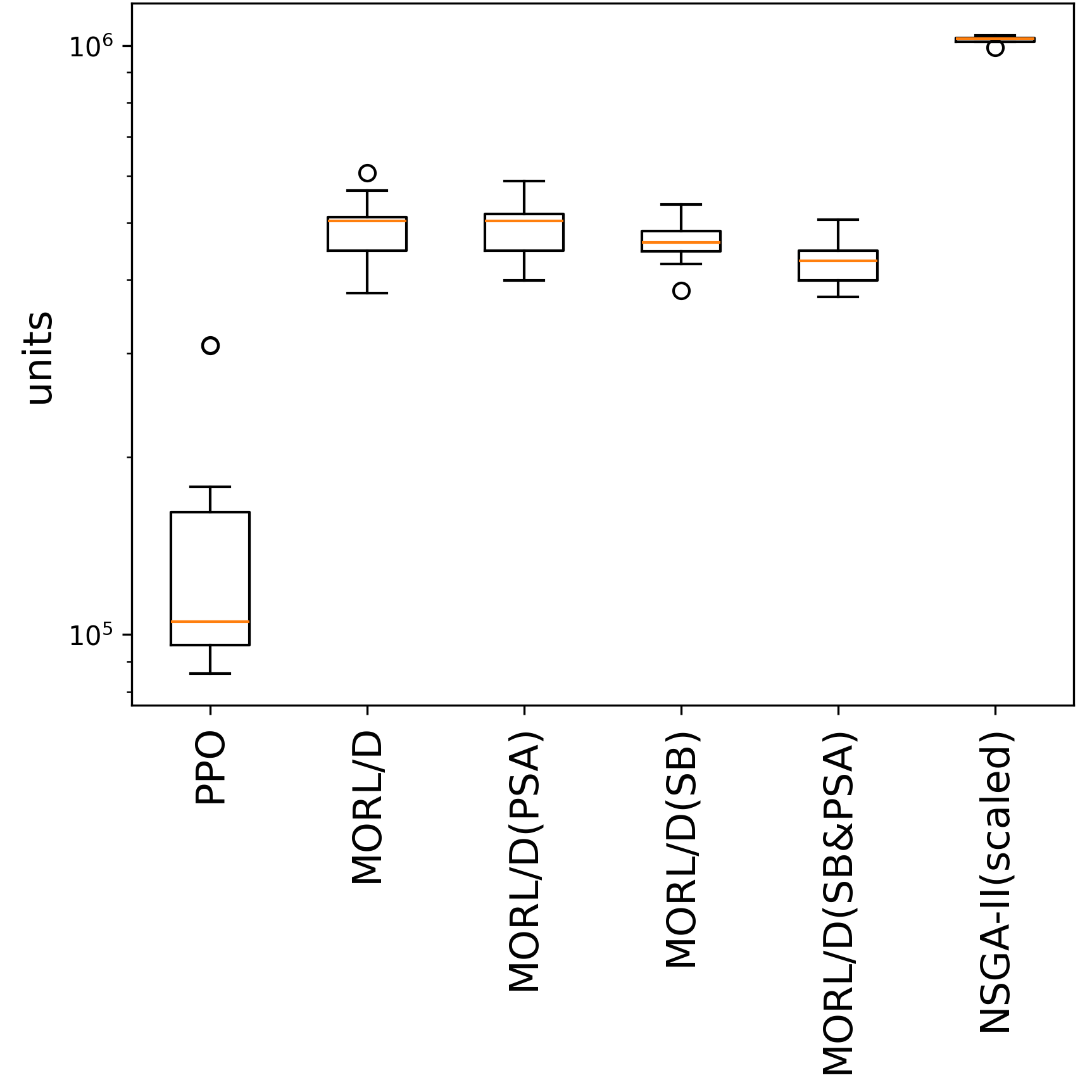}
        \caption{AHD-Complex SC}
        \label{fig:ahd_complex}
    \end{subfigure}
    \caption{AHD values of all algorithms expressed on logarithmic scales. MORL/D AHD exhibits the least values among the three algorithms in the simple SC problem (\ref{fig:ahd_simple}), while PPO AHD depicts the least values in the moderate and complex SC problems (\ref{fig:ahd_moderate} and~\ref{fig:ahd_complex}). Meanwhile, NSGA-II consistently shows the highest AHD values in all problems.}
    \label{fig:ahd}
\end{figure}

AHD is calculated to account for the varying number of solutions in the PF approximation sets, with lower values being preferable. Although PPO attains the maximum hypervolume in the simple SC (Figure~\ref{fig:ahd_simple}), MORL/D exceeds it in all other setups (Figures~\ref{fig:ahd_moderate} and~\ref{fig:ahd_complex}). PPO solutions appear to closely approximate the true PF approximation solutions with objective values that largely impact hypervolume metrics, i.e., higher profit values. However, these solutions are not well distributed across other segments near the frontier. As complexity increases, the gap between PPO and MORL/D AHD becomes more evident (Figure~\ref{fig:ahd_complex}). This phenomenon, similar to the hypervolume measure, may result from differences in the scales of the objectives. In the complex SC scenario, the profit scale is much larger compared to the other two objectives, making the latter's contributions minimal even when they are close to the actual PF set. In contrast, NSGA-II exhibits notably high AHD values, especially in more complex scenarios, indicating a reduction in efficiency as problem complexity increases.

\subsubsection{PF Solution Sets} \label{sec:pf_solution_sets}

\begin{figure}[H]
    \centering
    \begin{subfigure}{0.32\textwidth}
        \includegraphics[width=\textwidth]{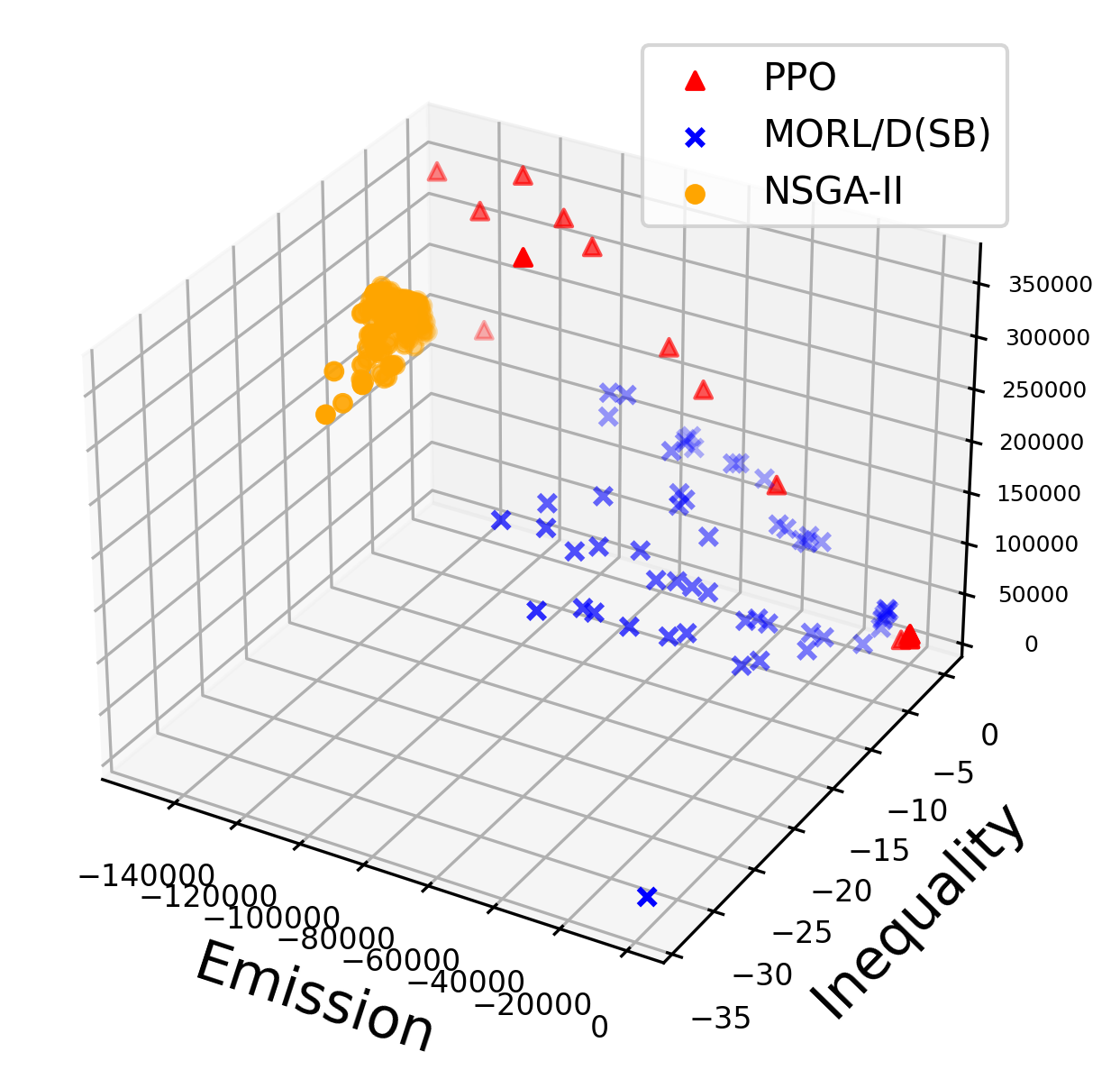}
        \caption{first 3D view-Simple SC}
    \end{subfigure}
    \hfill
    \begin{subfigure}{0.32\textwidth}
        \includegraphics[width=\textwidth]{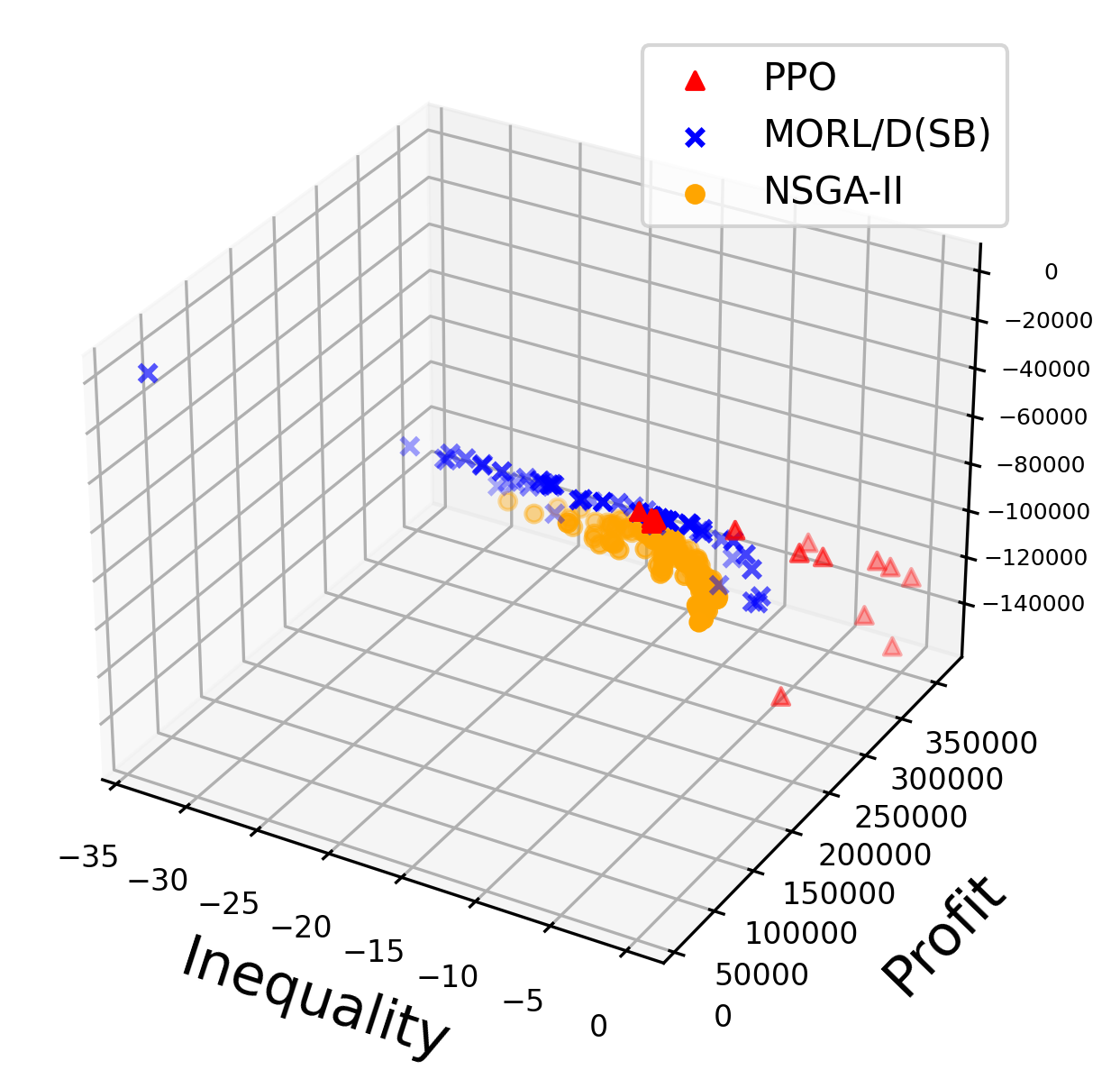}
        \caption{second 3D view-Simple SC}
    \end{subfigure}
    \hfill
    \begin{subfigure}{0.32\textwidth}
        \includegraphics[width=\textwidth]{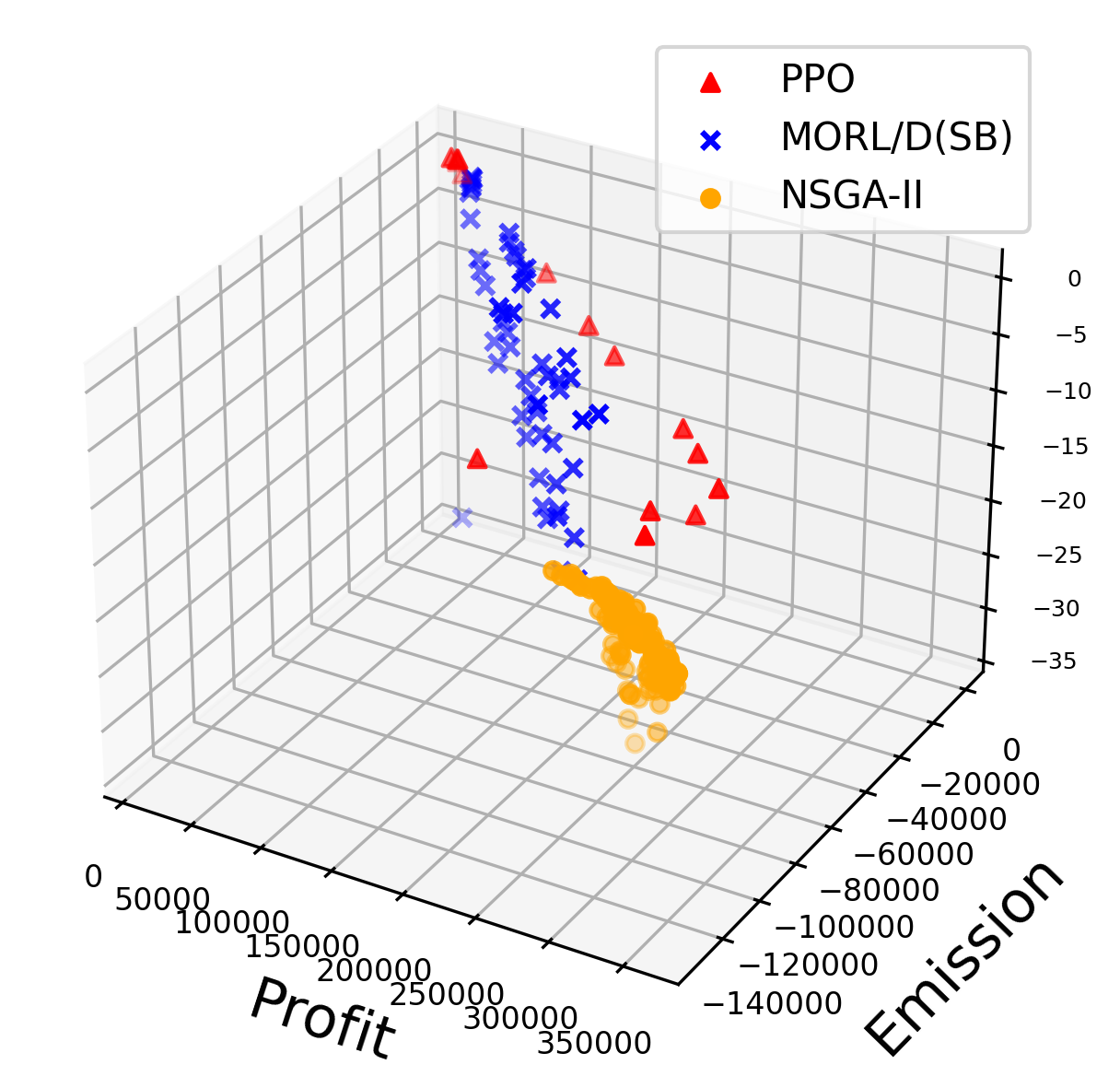}
        \caption{third 3D view-Simple SC}
    \end{subfigure}
    
    %\vskip\baselineskip
    
    \begin{subfigure}{0.32\textwidth}
        \includegraphics[width=\textwidth]{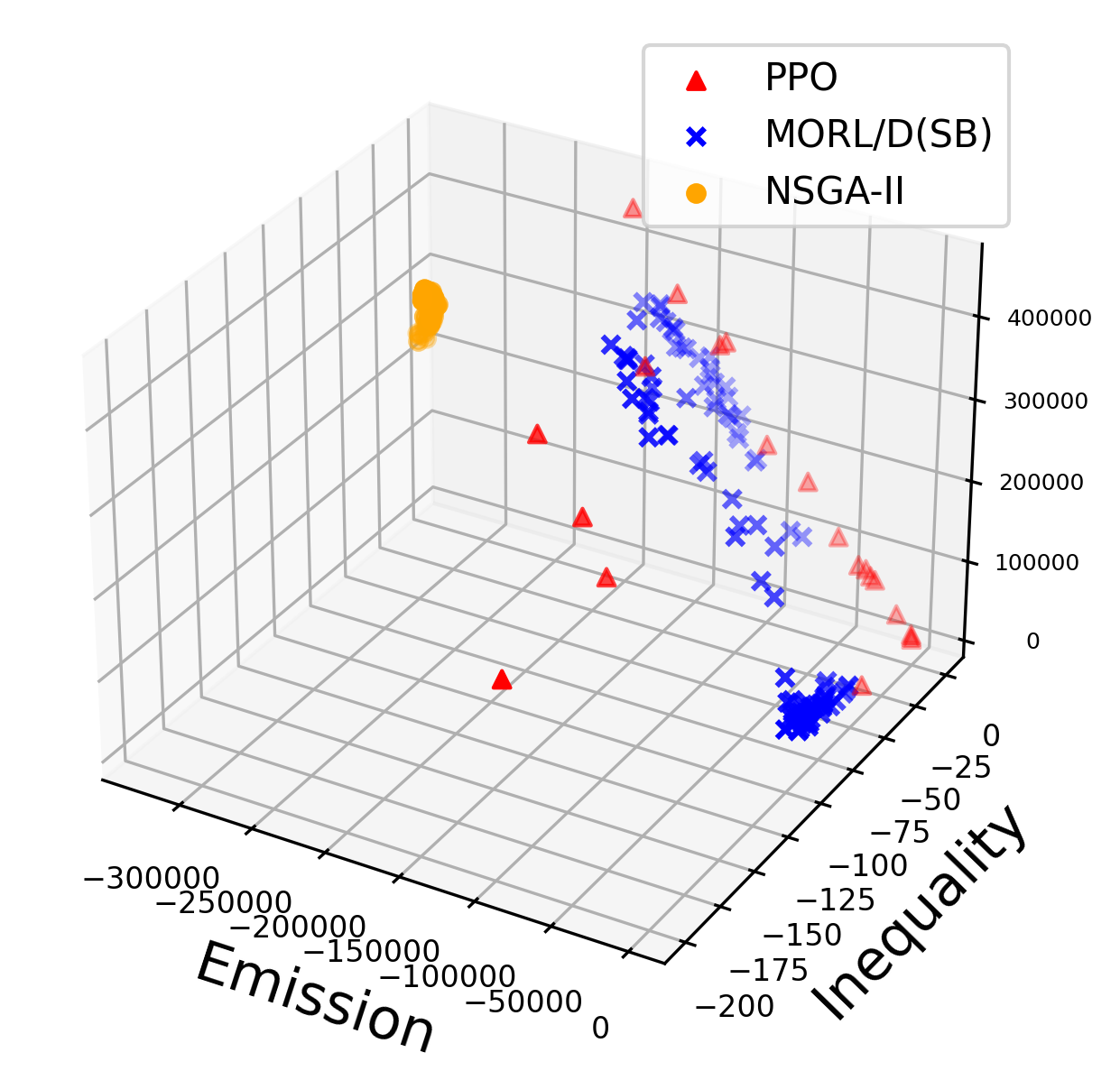}
        \caption{first 3D view-Moderate SC}
    \end{subfigure}   
    \hfill
    \begin{subfigure}{0.32\textwidth}
        \includegraphics[width=\textwidth]{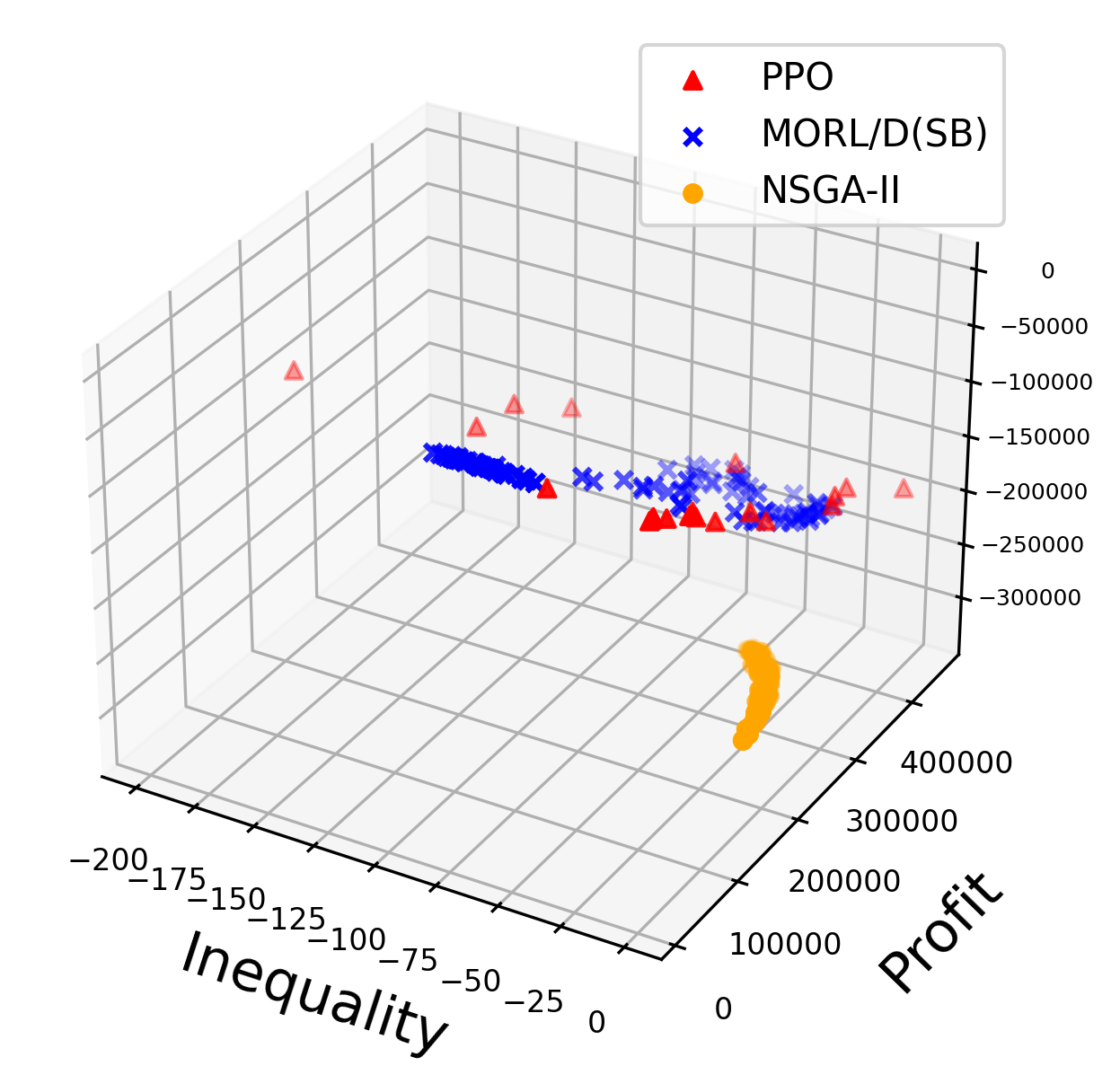}
        \caption{second 3D view-Moderate SC}
    \end{subfigure}
    \hfill
    \begin{subfigure}{0.32\textwidth}
        \includegraphics[width=\textwidth]{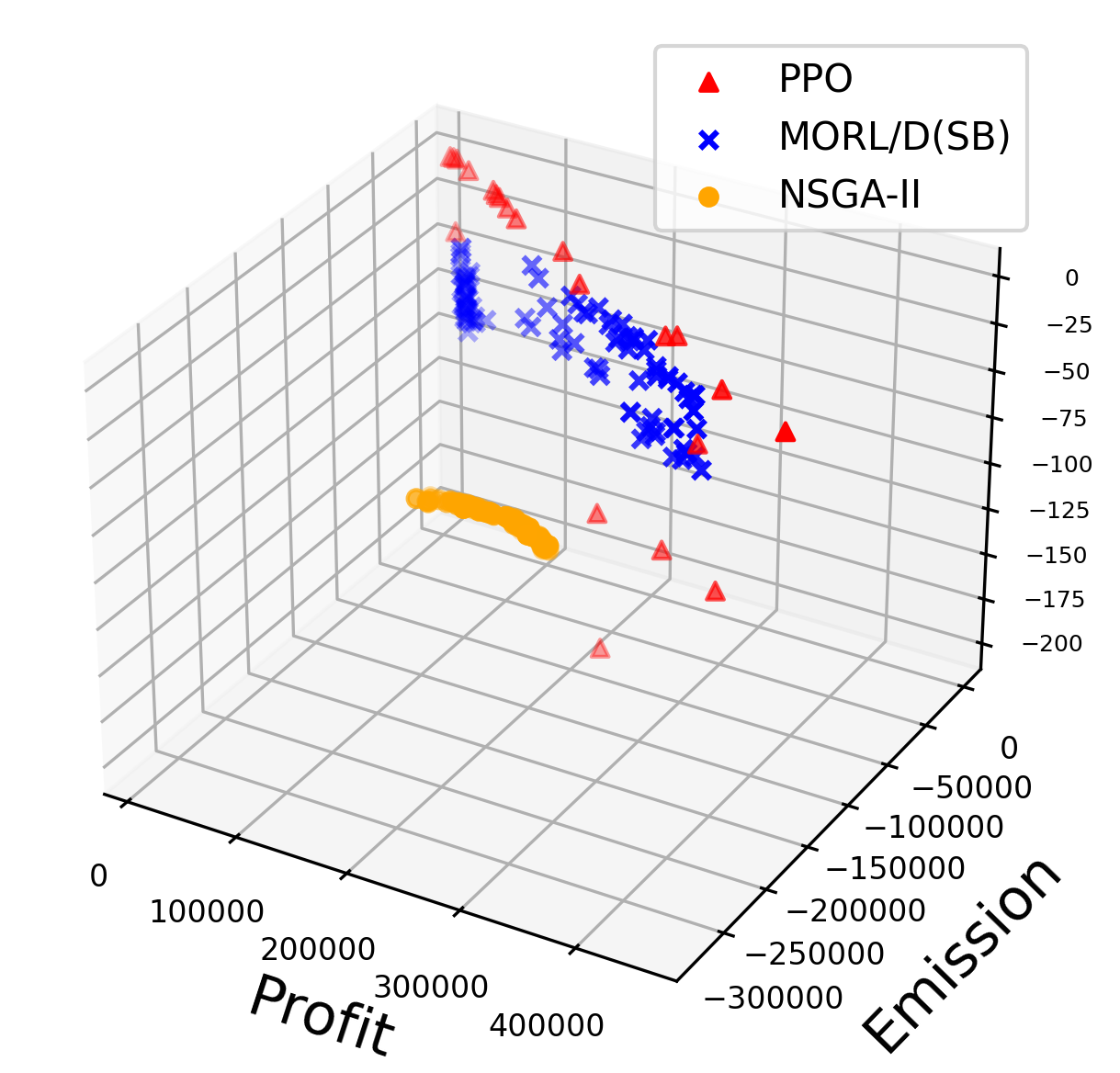}
        \caption{third 3D view-Moderate SC}
    \end{subfigure}

    \begin{subfigure}{0.32\textwidth}
        \includegraphics[width=\textwidth]{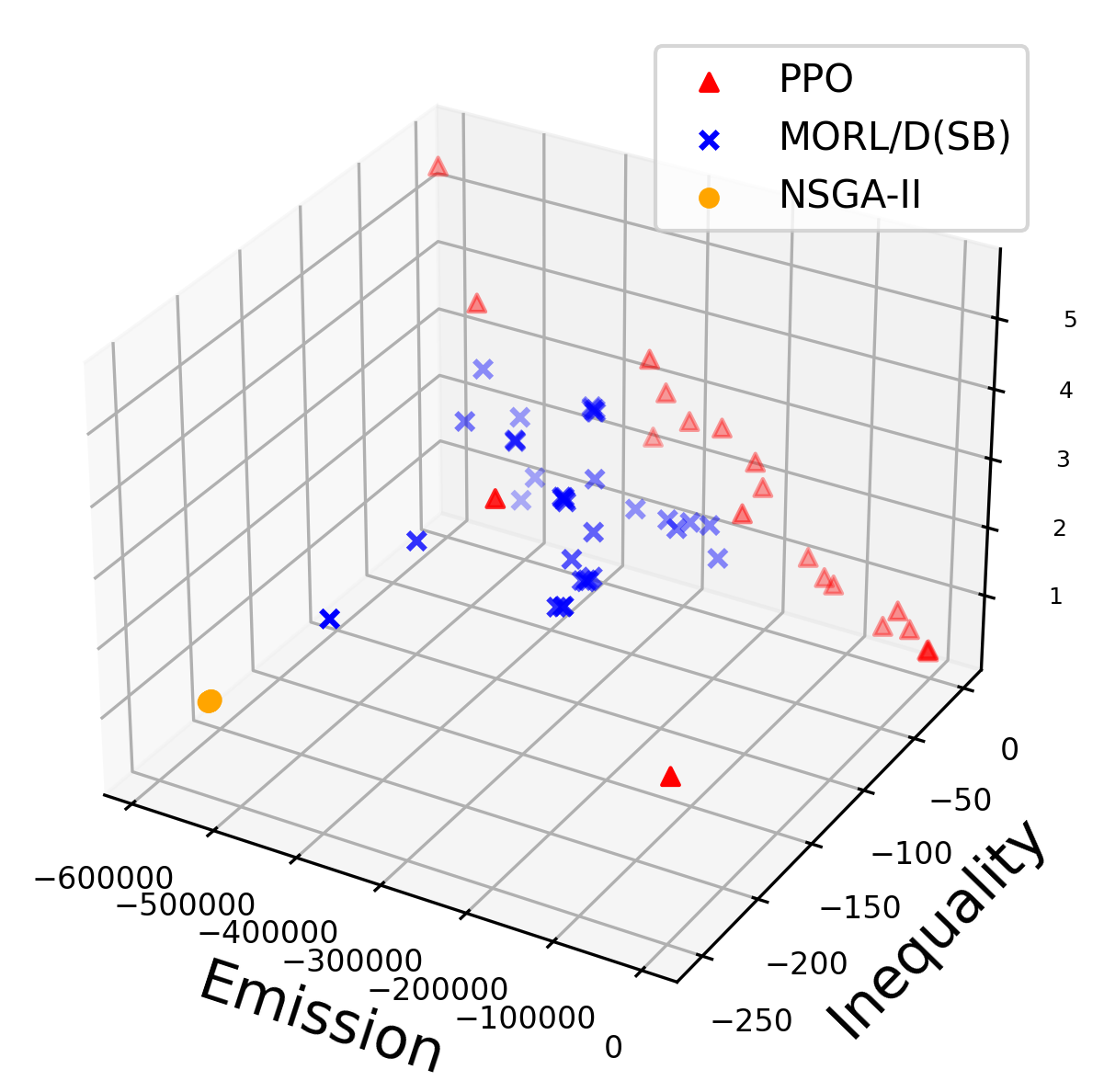}
        \caption{first 3D view-Complex SC}
        \label{fig:pf_complex1}
    \end{subfigure}   
    \hfill
    \begin{subfigure}{0.32\textwidth}
        \includegraphics[width=\textwidth]{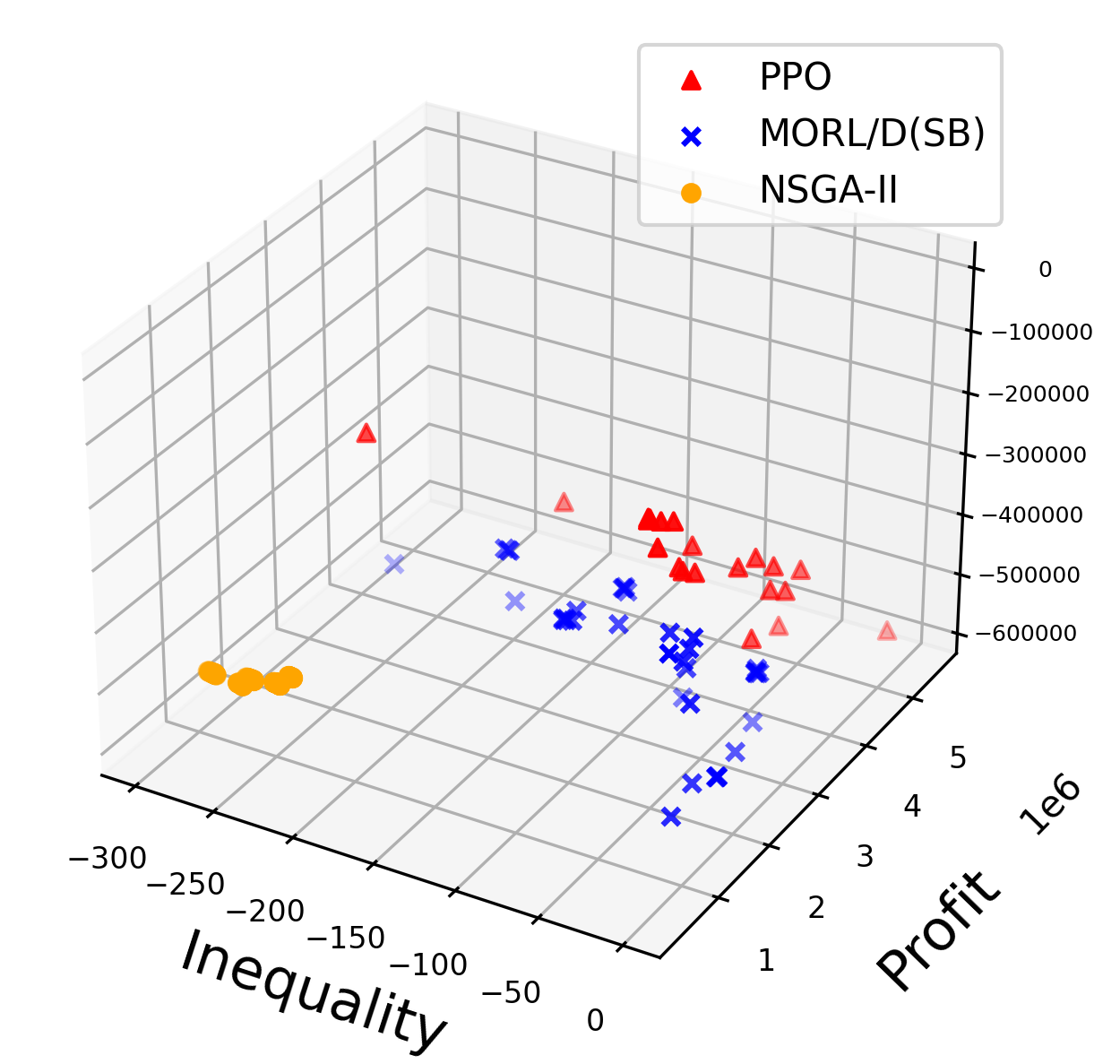}
        \caption{second 3D view-Complex SC}
        \label{fig:pf_complex2}
    \end{subfigure}
    \hfill
    \begin{subfigure}{0.32\textwidth}
        \includegraphics[width=\textwidth]{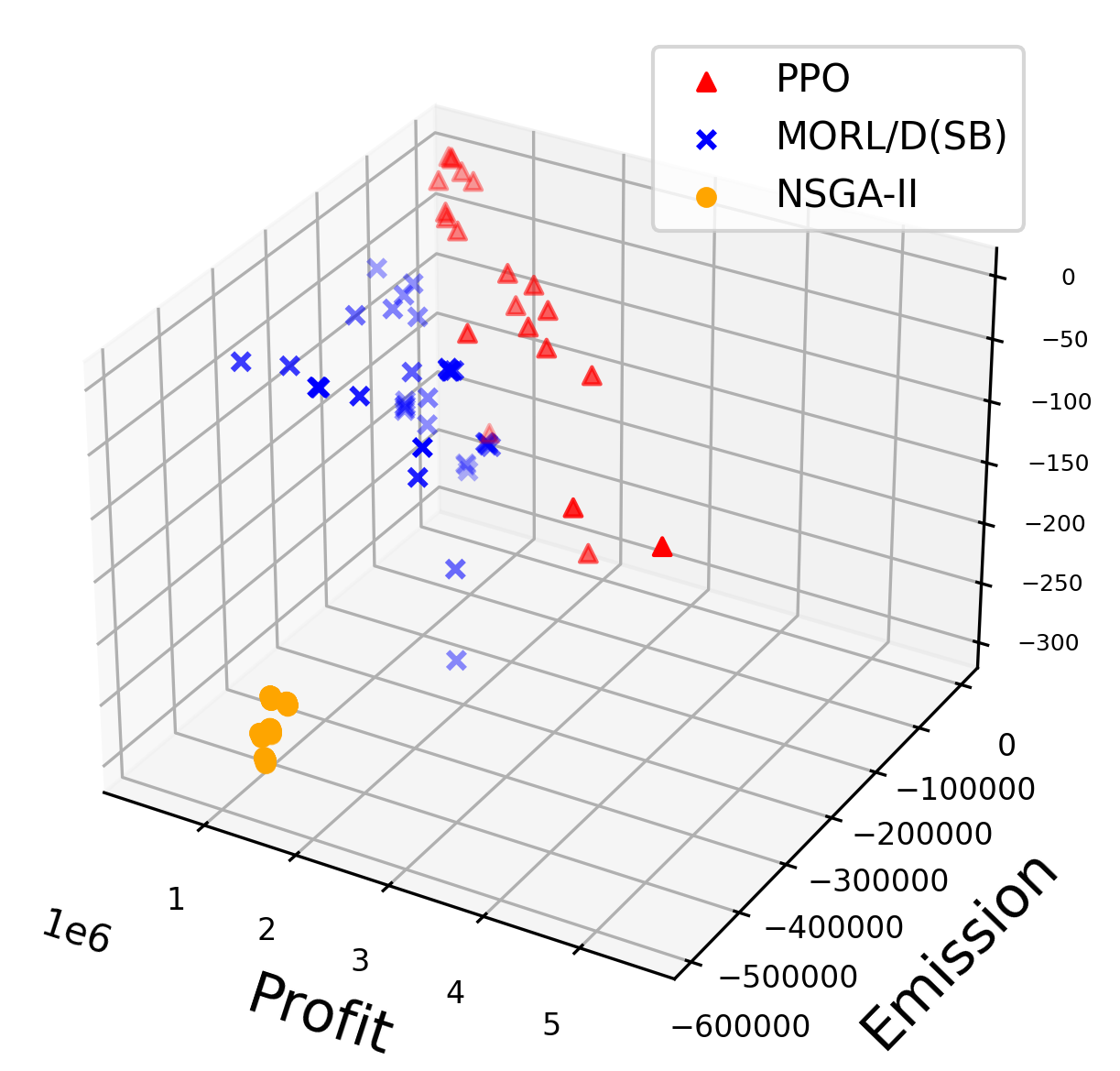}
        \caption{third 3D view-Complex SC}
        \label{fig:pf_complex3}
    \end{subfigure}
    \caption{The resulting PF approximation sets of the three methods. MORL/D shows the most balance between diversity, optimality, and density. A few PPO solution points are close to the 'ideal point', resulting in a high hypervolume value. Nonetheless, the solutions are sparse, diminishing the robustness of the model. Meanwhile, NSGA-II solutions are concentrated in a narrow solution space.}
    \label{fig:pf_all}
\end{figure}

The experiments result in 10, 40, and 20 PF approximation sets for PPO, MORL/D, and NSGA-II in each problem instantiation. Among all evaluated sets, we select the optimal PF approximation sets for each approach based on their hypervolume analysis. The PF approximation sets consist of non-dominated solutions, representing the best trade-offs found by each algorithm to approximate the true PF. We also examine all resulting PF approximation sets from all runs, as detailed in the supplementary material, to ensure visually that these PF approximation sets cover similar regions in the objective spaces and confirm that the samples discussed here are adequately representative of the overall results. Figure~\ref{fig:pf_all} illustrates the distribution of solutions for each optimal PF. The PF approximation set derived from PPO is notably the sparsest, aligning with its sparsity metric. In every problem scenario, there exists a clear inverse relationship between profit and GHG emission values, where increases in profit coincide with higher emissions. In contrast, the connections between the SL inequality and the other two objectives are less pronounced, since changes in GHG emissions or profit do not directly impact the SL inequality. This subtlety stems from an indirect relationship between SL inequality and profit or GHG emissions, as well as the narrower range used for SL inequality compared to the other objectives. Consequently, the PPO agent does not effectively encourage variability in the SL inequality values. It is crucial to note that the solution quality from PPO is heavily reliant on pre-defined weights, which remain fixed throughout the learning process.

In contrast, the NSGA-II PF approximation set exhibits a tendency toward concentration. As problem complexity increases, both the profit and the SL inequality deteriorate, specifically, profit declines while the SL inequality rises. This inadequate performance is particularly noticeable in complex SC issues (Figures~\ref{fig:pf_complex1},~\ref{fig:pf_complex2},~\ref{fig:pf_complex3}), where 38 of the generated solutions exhibit visually indistinguishable proximal values within unfavourable regions marked by low profitability, high GHG emissions, and increased SL inequality. This phenomenon reveals the possibility of the action search being trapped in local optima and corroborates the plausibility of exploration-exploitation imbalance as previously discussed in Section~\ref{sec:overall_performance}. Furthermore, this also aligns with the observations of~\citet{seyyedabbasi_hybrid_2021}, who stated that RL-based techniques surpass metaheuristic methods in exploring new search areas.

In line with the hypervolume metric, PPO surpasses MORL/D overall solutions in optimality, where some solutions generated by MORL/D are dominated by those of PPO. However, the MORL/D agent provides the most diversified and densely populated solutions among the compared approaches. Unlike PPO, MORL/D weights are not predefined but are adaptively updated during training based on the results of each policy update. This adaptability improves the diversity and density of solutions. In all SC problems, solutions extend in all directions toward the PF. Furthermore, the spread of solutions showcases a balanced representation of three objectives across the solution space, indicating the efficacy of MORL/D's normalisation mechanism in achieving equilibrium among objectives, regardless of their value ranges. Consequently, the MORL/D PF approximation set emerges as the most sensible solution in the multi-objective context due to its extensive range, dense dispersion, and balanced integration of all objectives.

\subsubsection{Operational Behaviour} \label{sec:operational_behaviour}
\begin{figure}[H]
    \centering
    \begin{subfigure}{0.32\textwidth}
        \includegraphics[width=\textwidth]{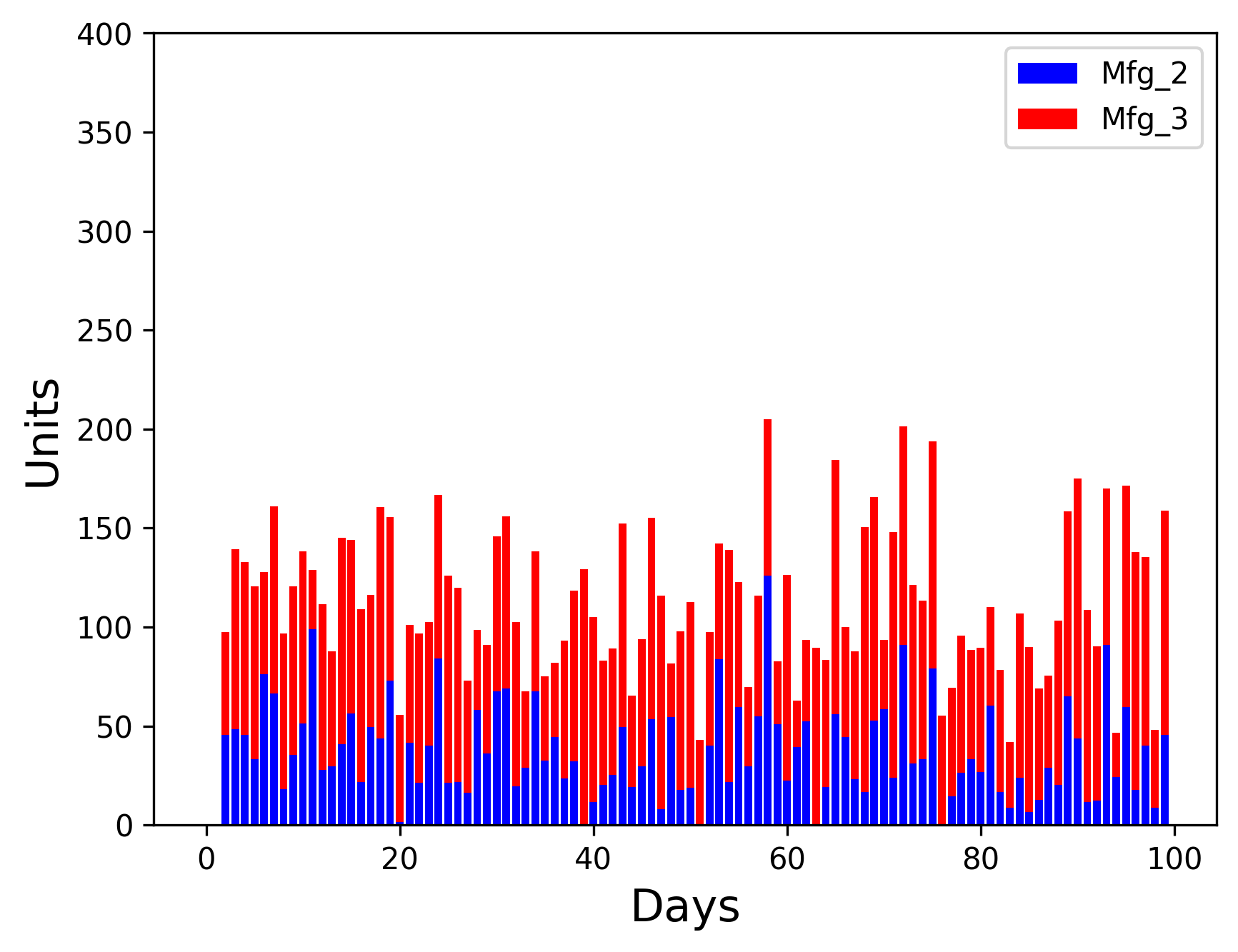}
        \caption{PPO-Simple SC}
        \label{fig:mfg_ppo_simple}
    \end{subfigure}
    \hfill
    \begin{subfigure}{0.32\textwidth}
        \includegraphics[width=\textwidth]{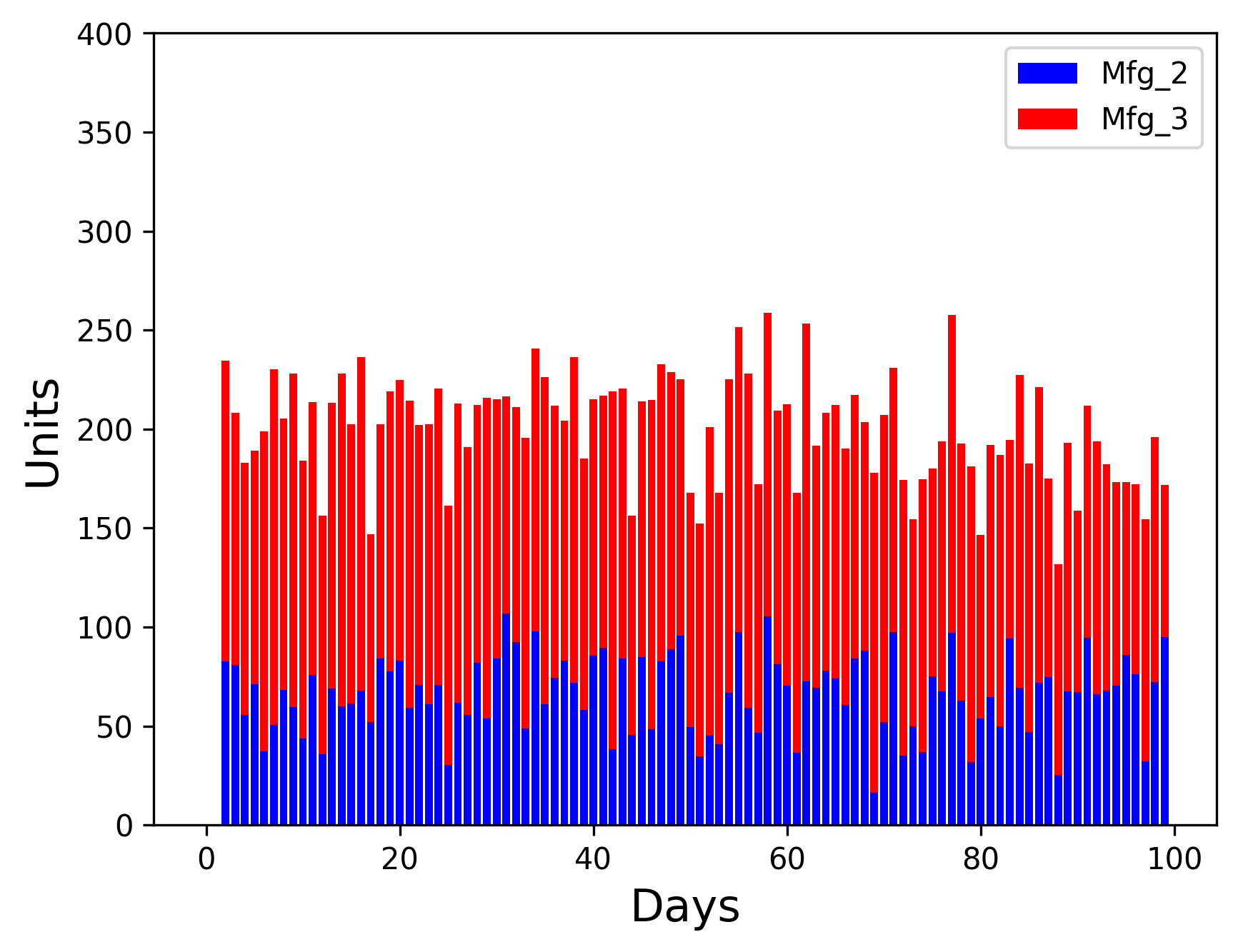}
        \caption{MORL/D-Simple SC}
        \label{fig:mfg_morld_simple}
    \end{subfigure}
    \hfill
    \begin{subfigure}{0.32\textwidth}
        \includegraphics[width=\textwidth]{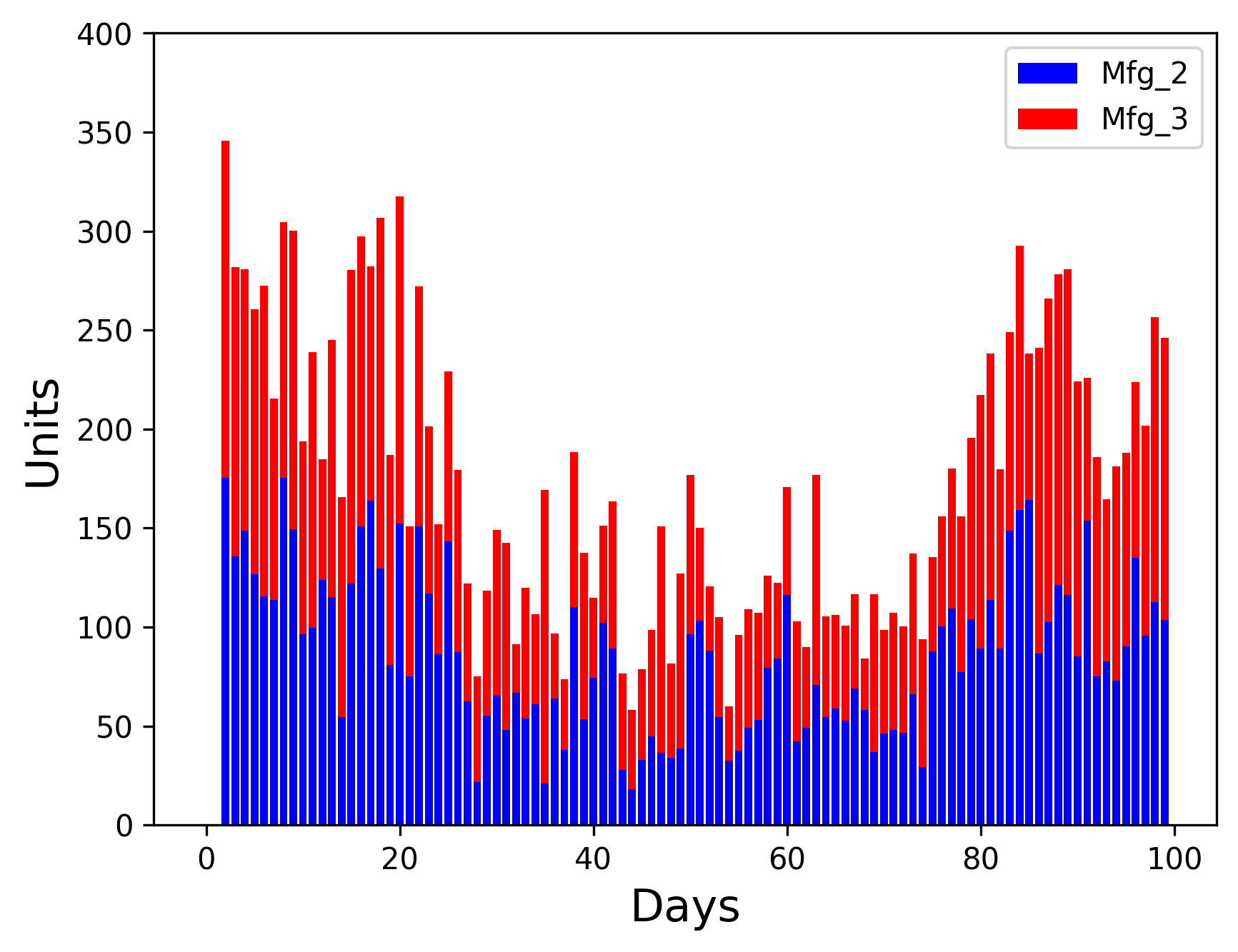}
        \caption{NSGA-Simple SC}
        \label{fig:mfg_nsga_simple}
    \end{subfigure}

        \begin{subfigure}{0.32\textwidth}
        \includegraphics[width=\textwidth]{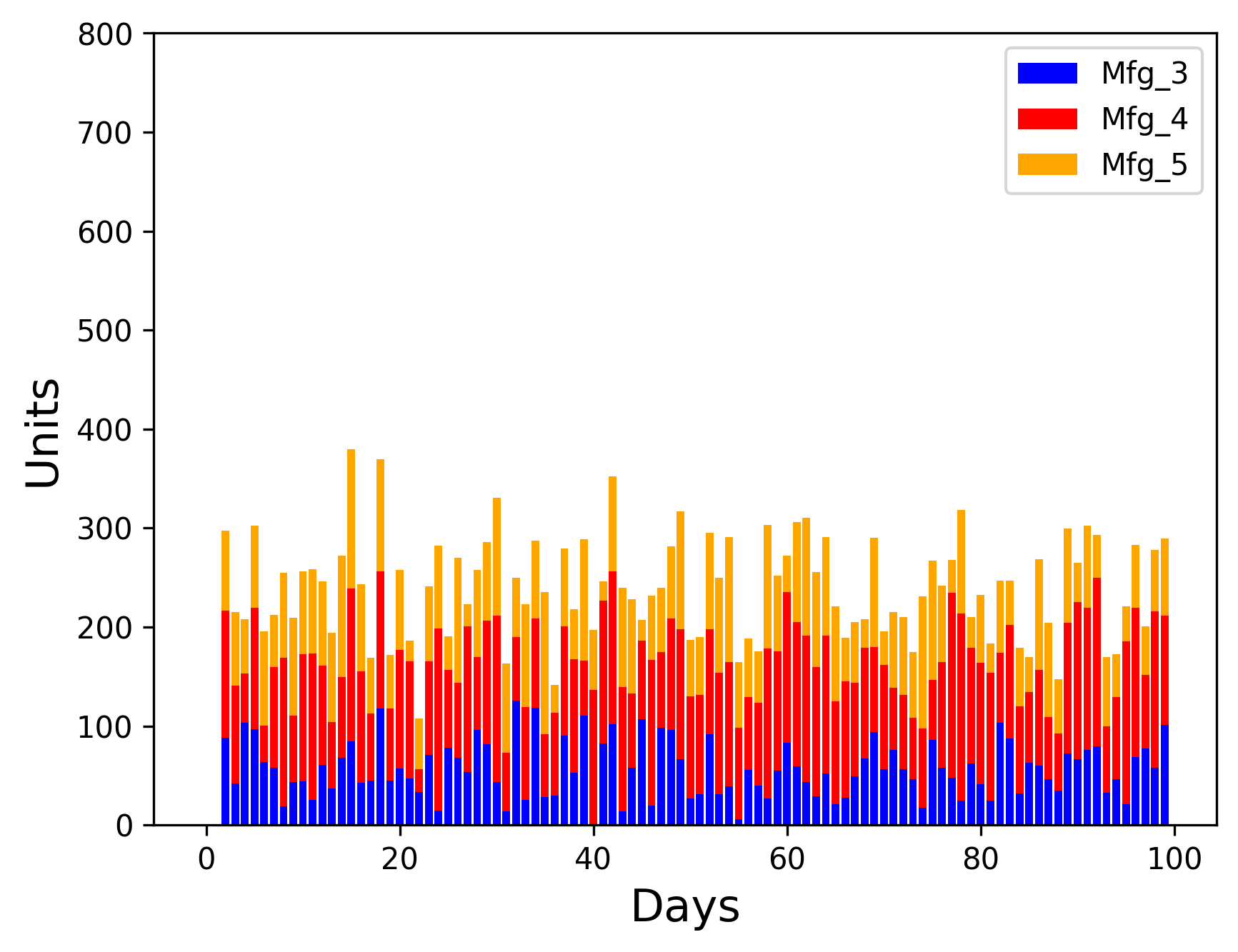}
        \caption{PPO-Moderate SC}
        \label{fig:mfg_ppo_moderate}
    \end{subfigure}
    \hfill
    \begin{subfigure}{0.32\textwidth}
        \includegraphics[width=\textwidth]{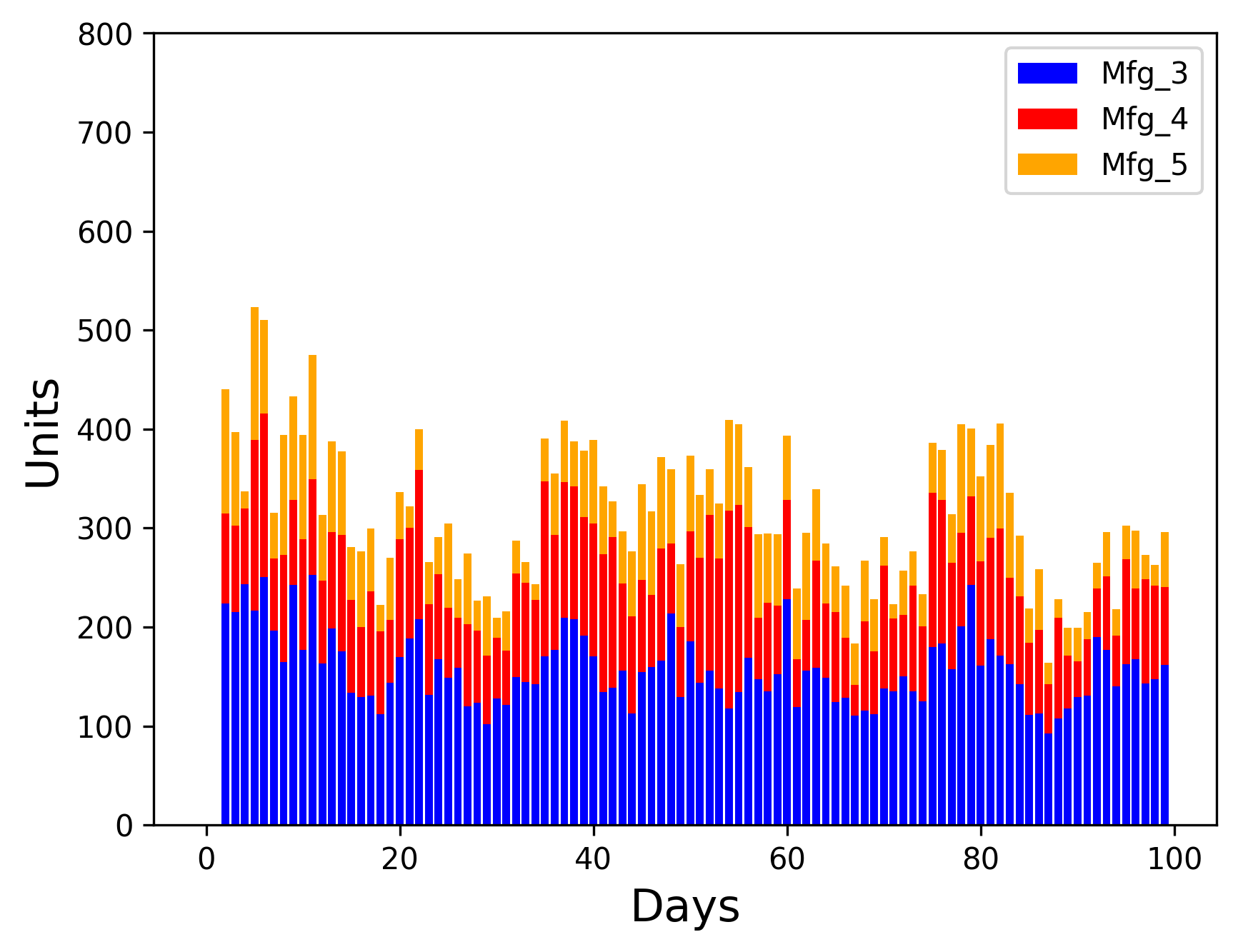}
        \caption{MORL/D-Moderate SC}
        \label{fig:mfg_morld_moderate}
    \end{subfigure}
    \hfill
    \begin{subfigure}{0.32\textwidth}
        \includegraphics[width=\textwidth]{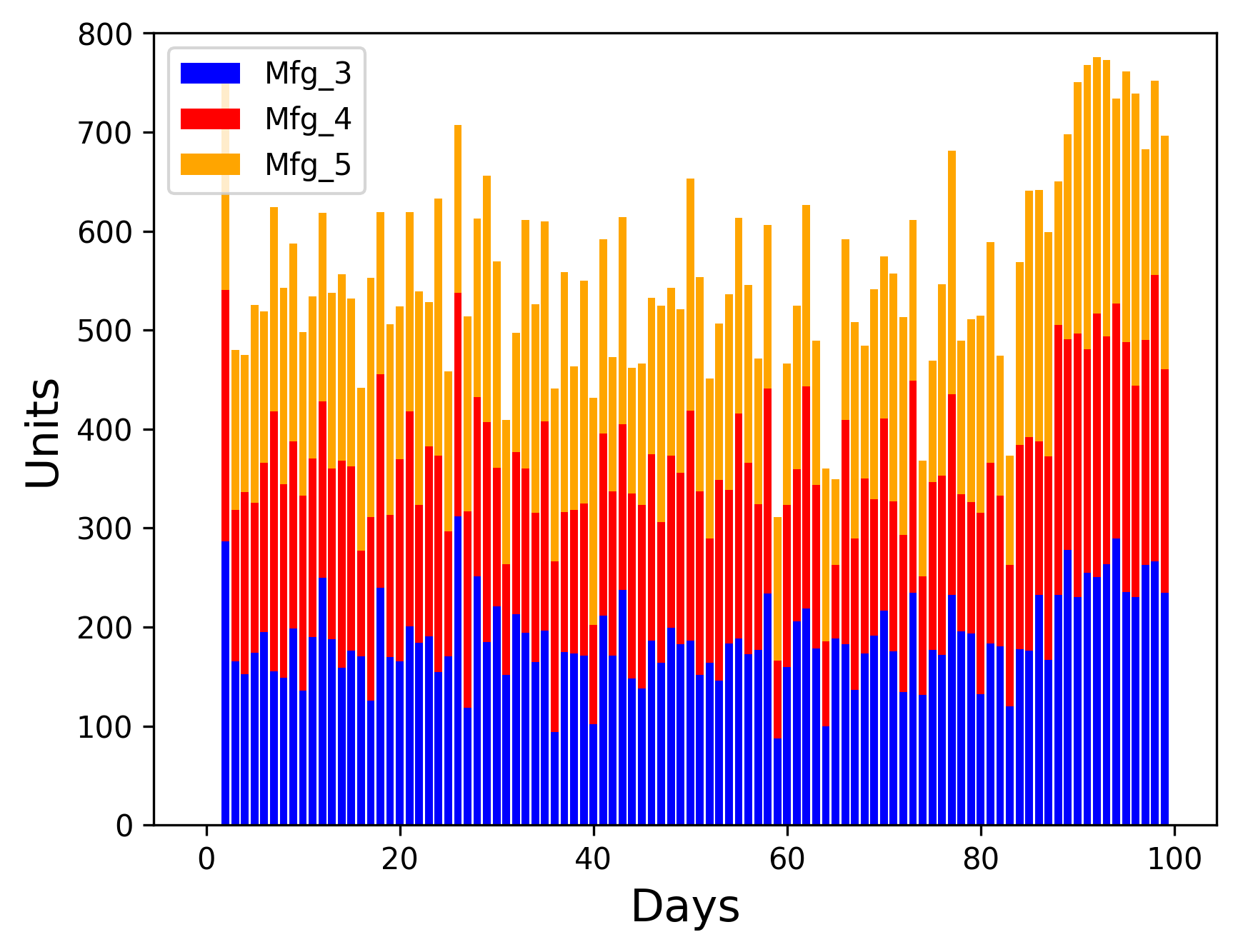}
        \caption{NSGA-Moderate SC}
        \label{fig:mfg_nsga_moderate}
    \end{subfigure}
            \begin{subfigure}{0.32\textwidth}
        \includegraphics[width=\textwidth]{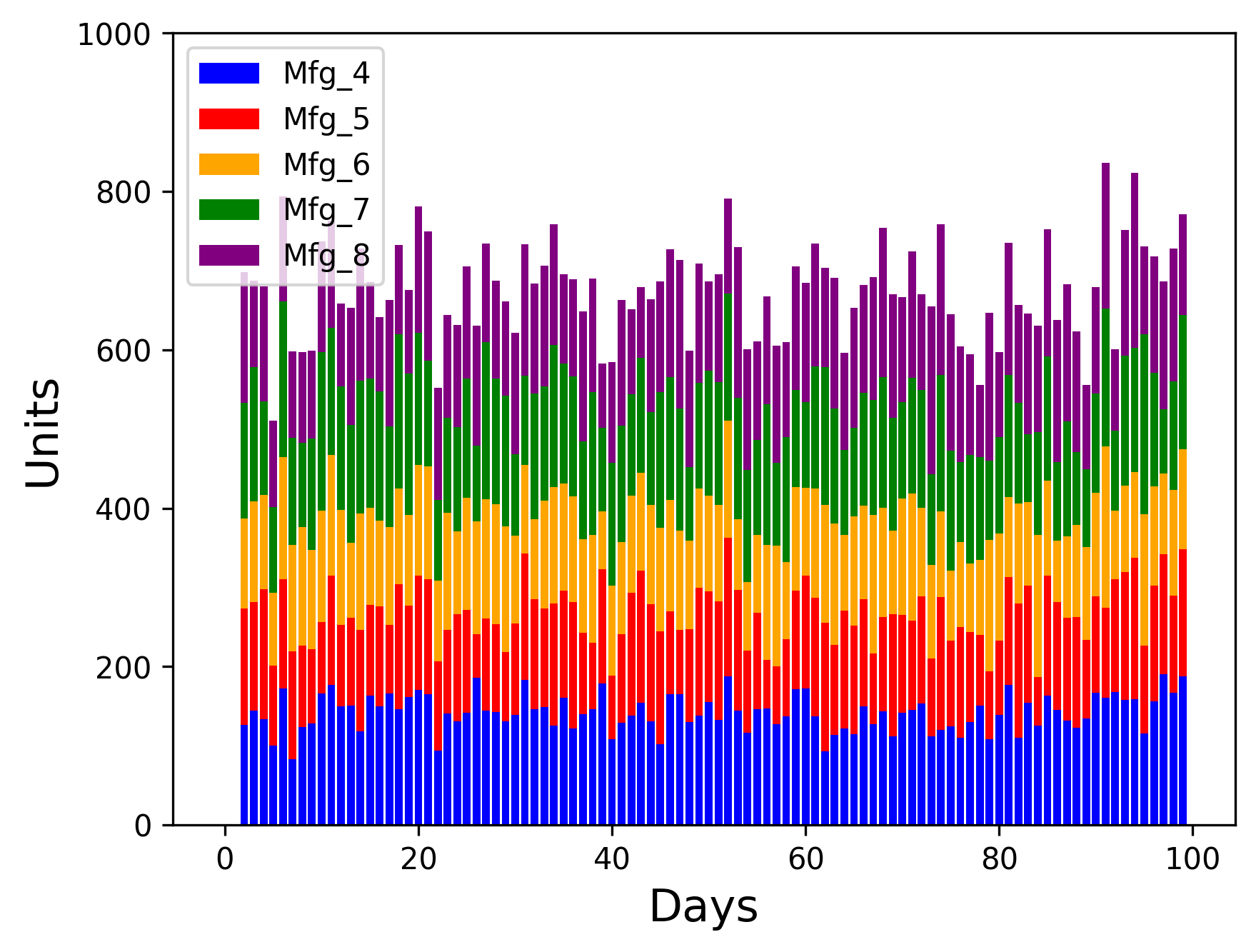}
        \caption{PPO-Complex SC}
        \label{fig:mfg_ppo_complex}
    \end{subfigure}
    \hfill
    \begin{subfigure}{0.32\textwidth}
        \includegraphics[width=\textwidth]{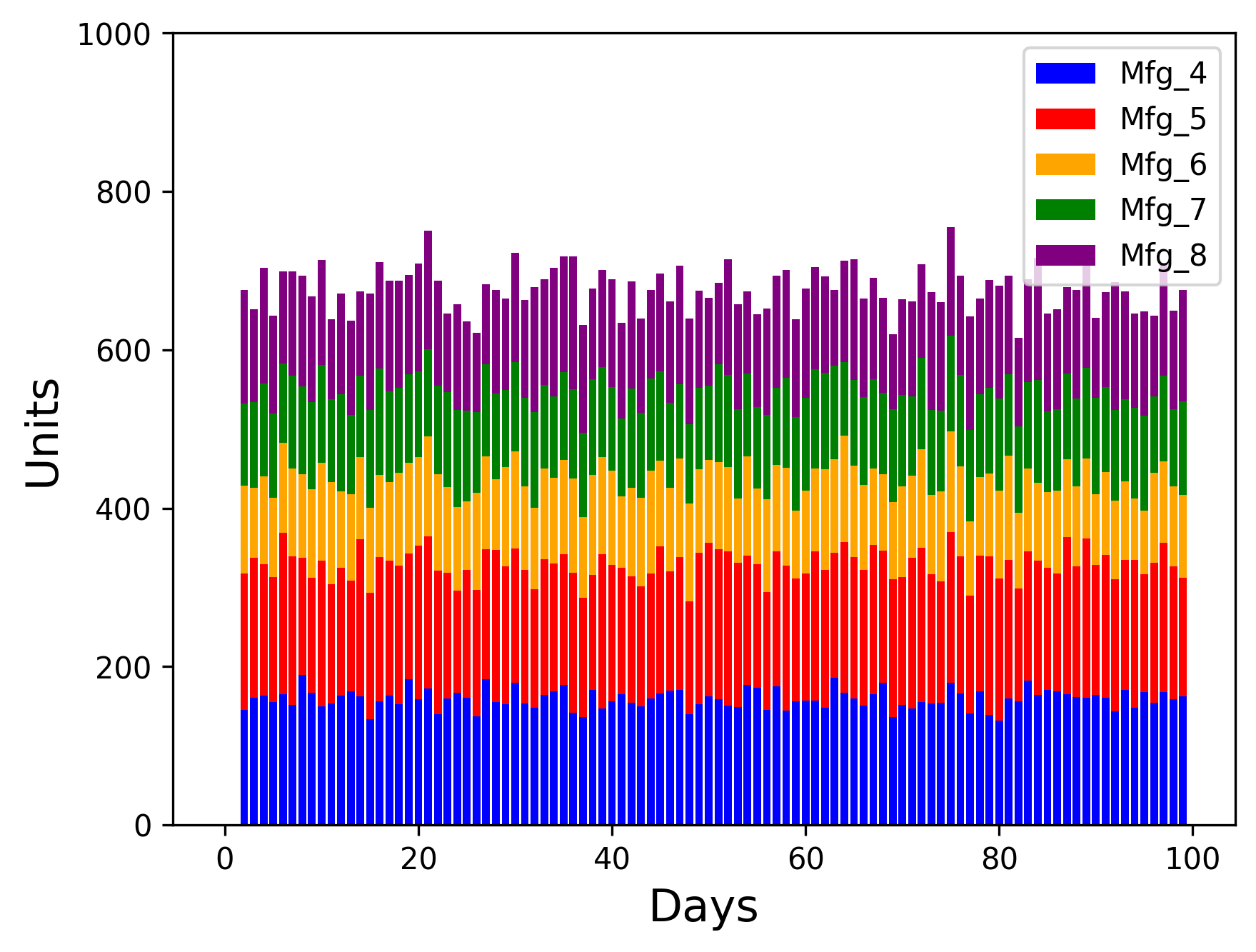}
        \caption{MORL/D-Complex SC}
        \label{fig:mfg_morld_complex}
    \end{subfigure}
    \hfill
    \begin{subfigure}{0.32\textwidth}
        \includegraphics[width=\textwidth]{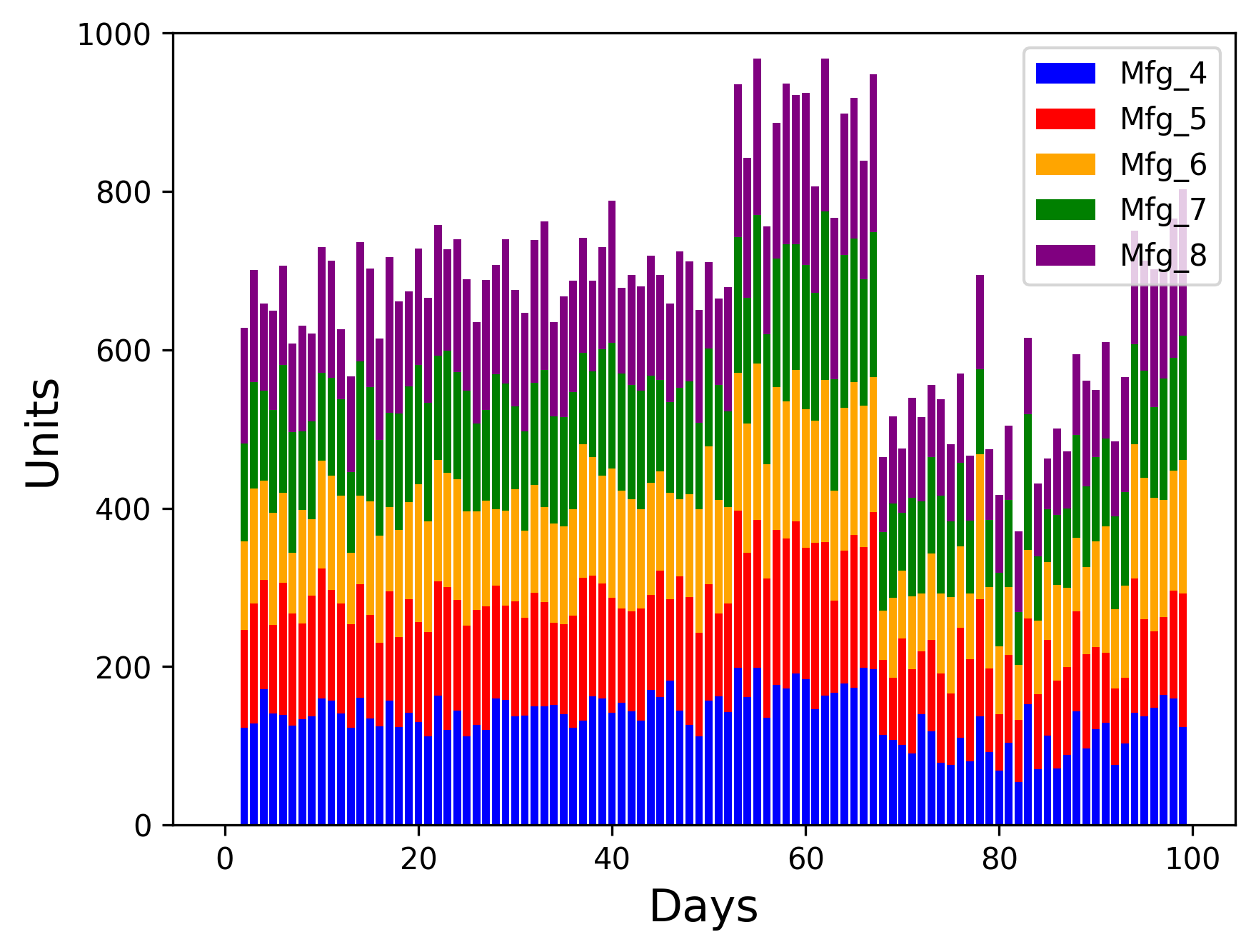}
        \caption{NSGA-Complex SC}
        \label{fig:mfg_nsga_complex}
    \end{subfigure}
    \caption{The comparison of manufacturing quantities resulted from the three algorithms in each SC problem. MORL/D displays relatively more stable production levels with moderate production quantity along the periods (\ref{fig:mfg_morld_simple},~\ref{fig:mfg_morld_moderate},~\ref{fig:mfg_morld_complex}). Its manufacturing quantity is less affected by the demand fluctuation. Meanwhile, PPO exhibits relatively stable yet low production levels (\ref{fig:mfg_ppo_simple},~\ref{fig:mfg_ppo_moderate},~\ref{fig:mfg_ppo_complex}). Moreover, NSGA-II generates more fluctuating production during periods (\ref{fig:mfg_nsga_simple},~\ref{fig:mfg_nsga_moderate},~\ref{fig:mfg_nsga_complex}).}
    \label{fig:mfg}
\end{figure}

\begin{figure}[H]
    \centering
    \begin{subfigure}{0.32\textwidth}
        \includegraphics[width=\textwidth]{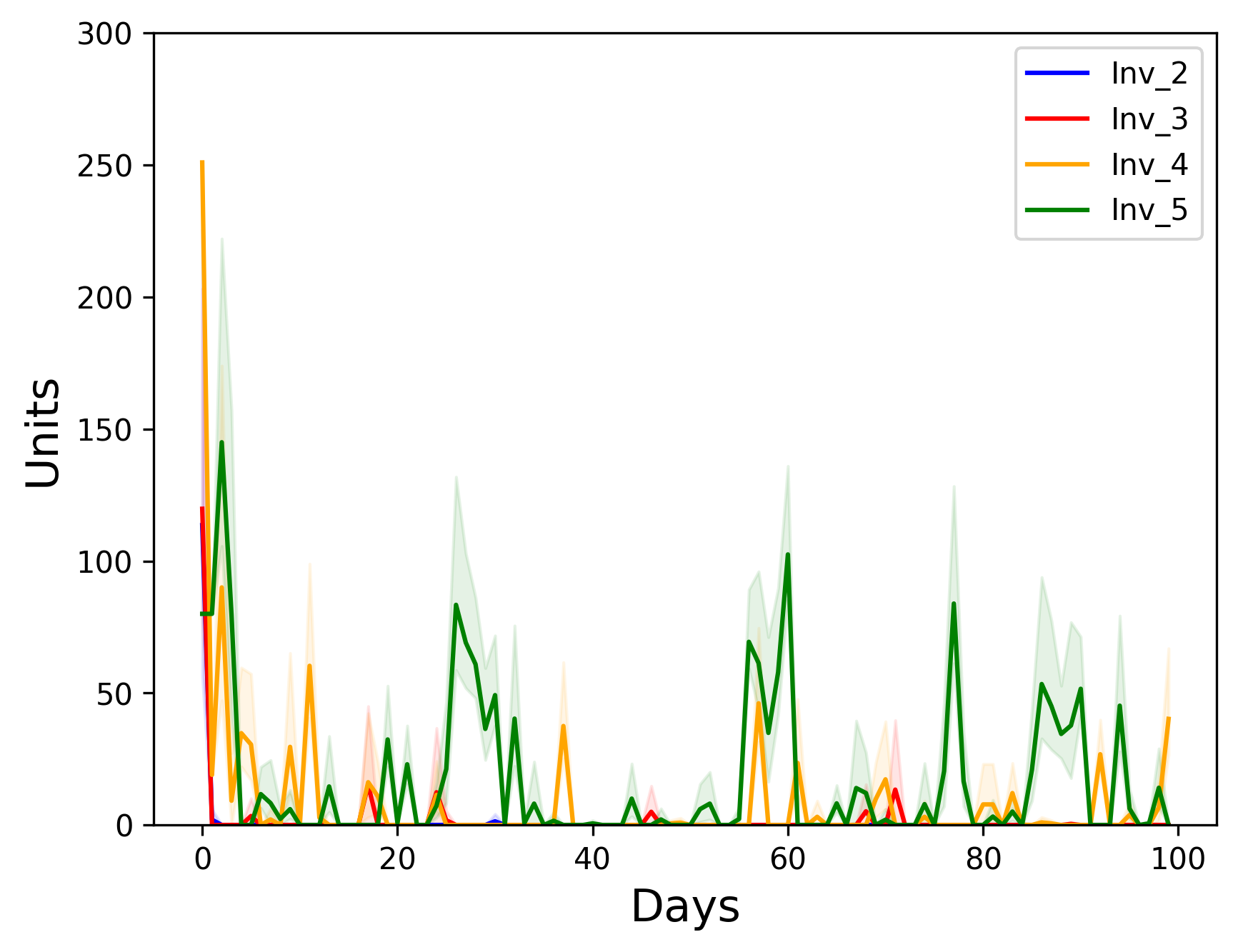}
        \caption{PPO-Simple SC}
        \label{fig:inv_ppo_simple}
    \end{subfigure}
    \hfill
    \begin{subfigure}{0.32\textwidth}
        \includegraphics[width=\textwidth]{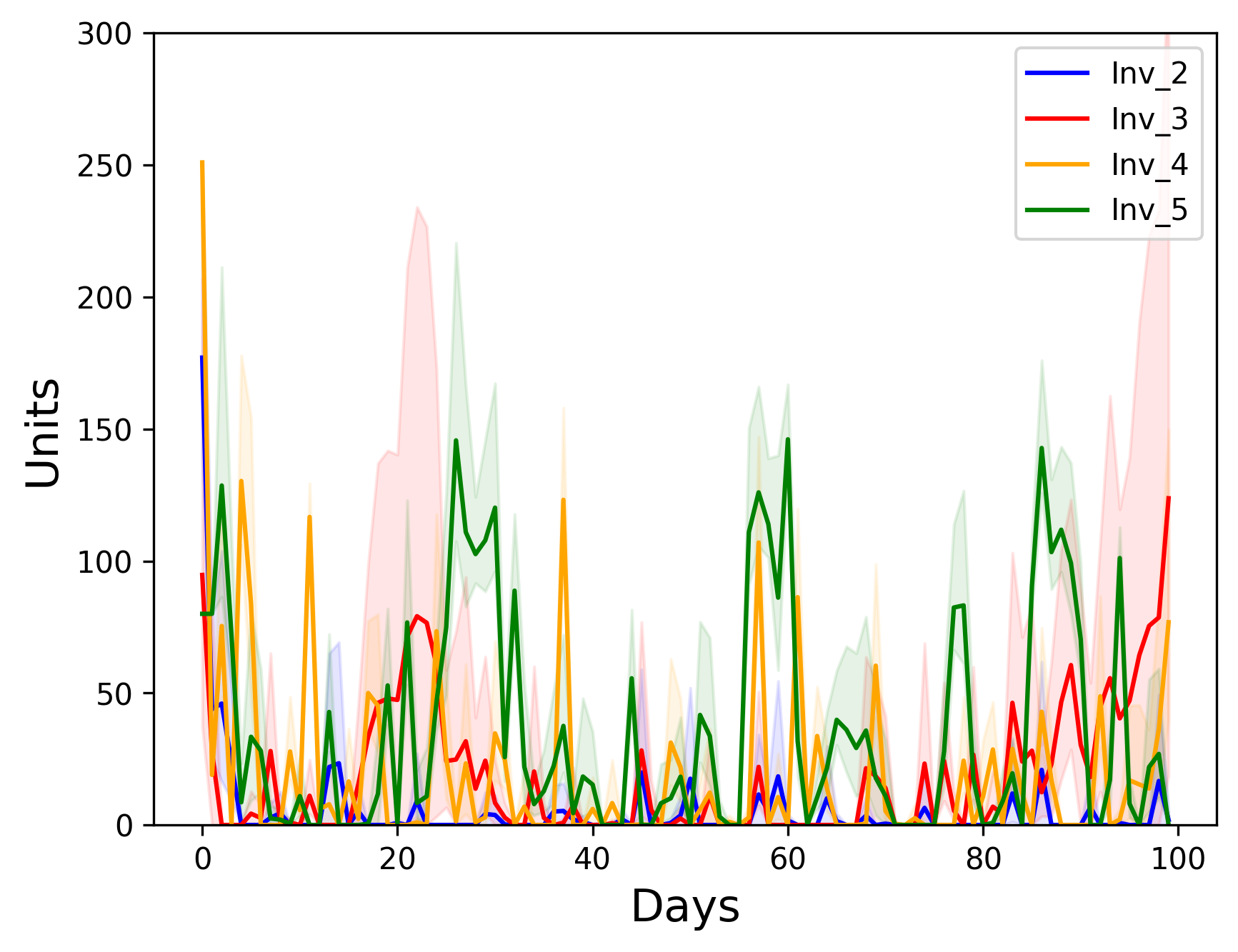}
        \caption{MORL/D-Simple SC}
        \label{fig:inv_morld_simple}
    \end{subfigure}
    \hfill
    \begin{subfigure}{0.32\textwidth}
        \includegraphics[width=\textwidth]{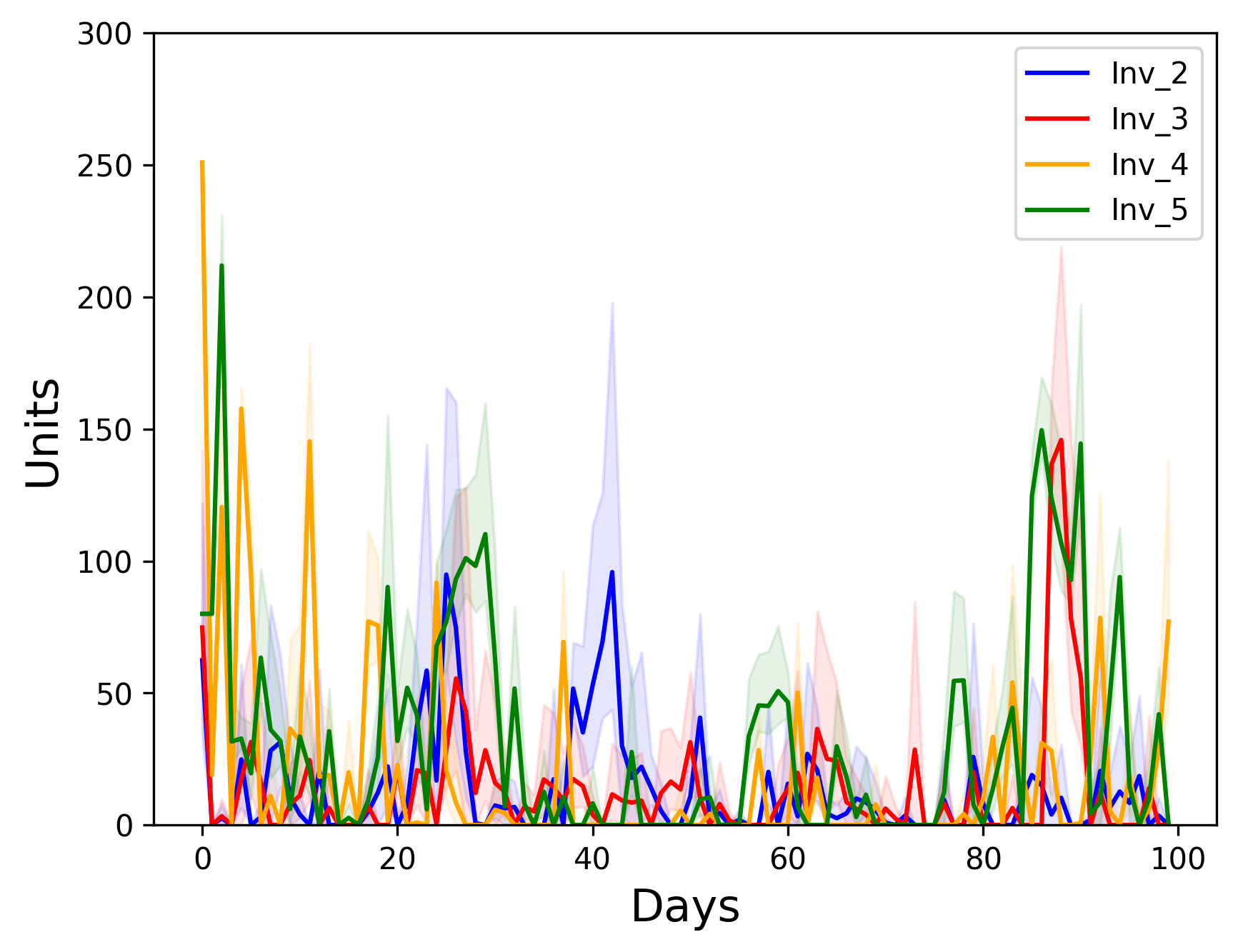}
        \caption{NSGA-II-Simple SC}
        \label{fig:inv_nsga_simple}
    \end{subfigure}

    \begin{subfigure}{0.32\textwidth}
        \includegraphics[width=\textwidth]{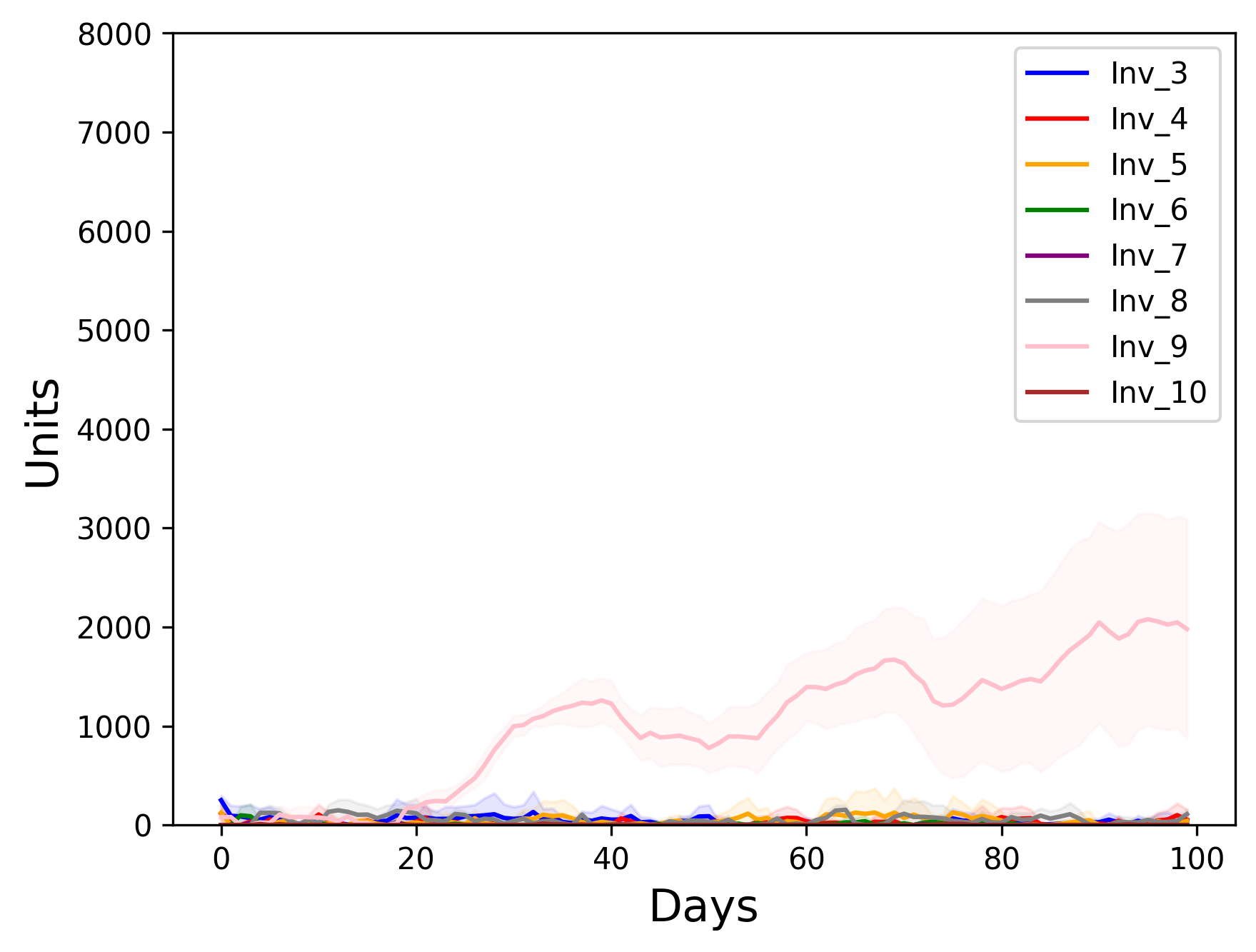}
        \caption{PPO-Moderate SC}
        \label{fig:inv_ppo_moderate}
    \end{subfigure}
    \hfill
    \begin{subfigure}{0.32\textwidth}
        \includegraphics[width=\textwidth]{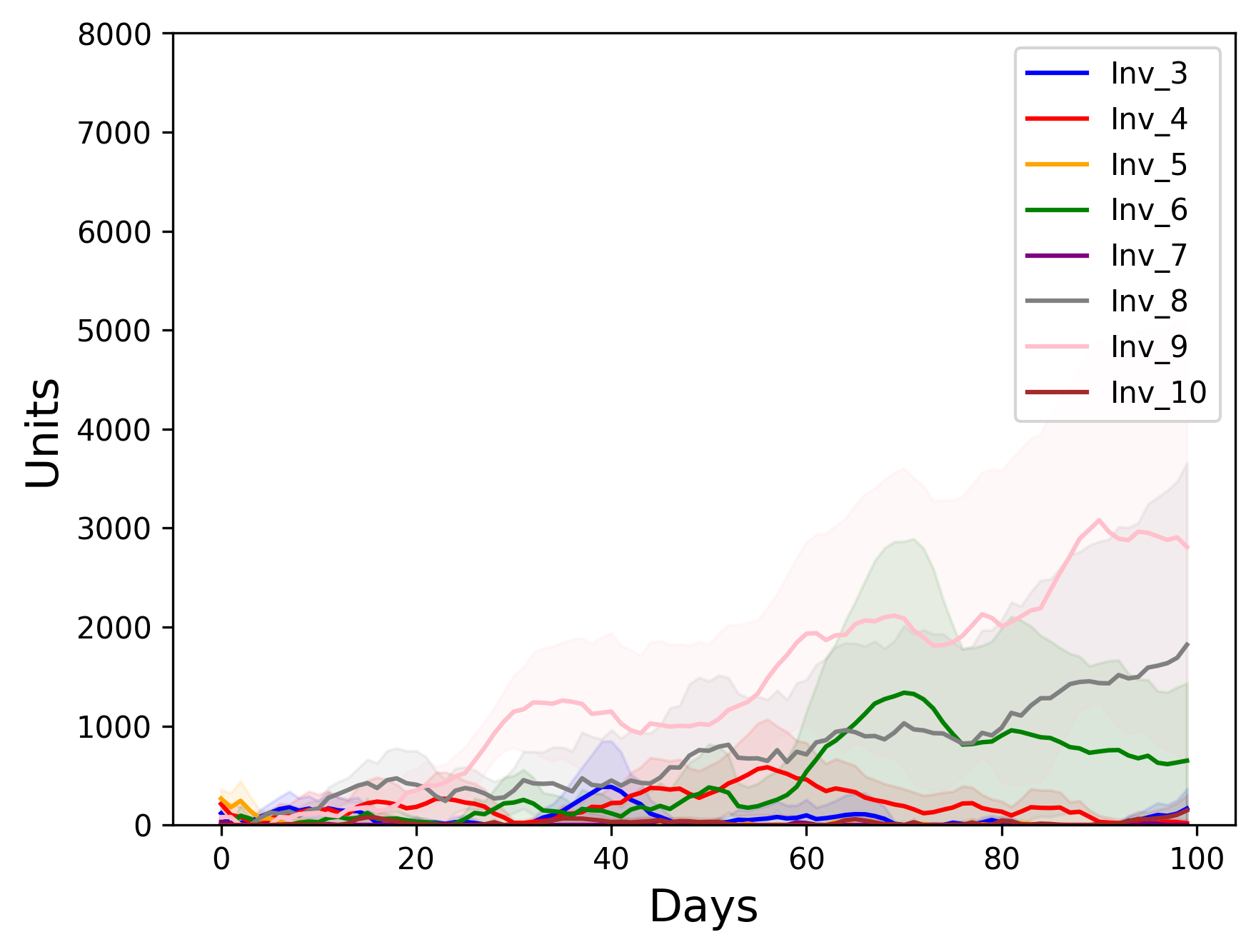}
        \caption{MORL/D-Moderate SC}
        \label{fig:inv_morld_moderate}
    \end{subfigure}
    \hfill
    \begin{subfigure}{0.32\textwidth}
        \includegraphics[width=\textwidth]{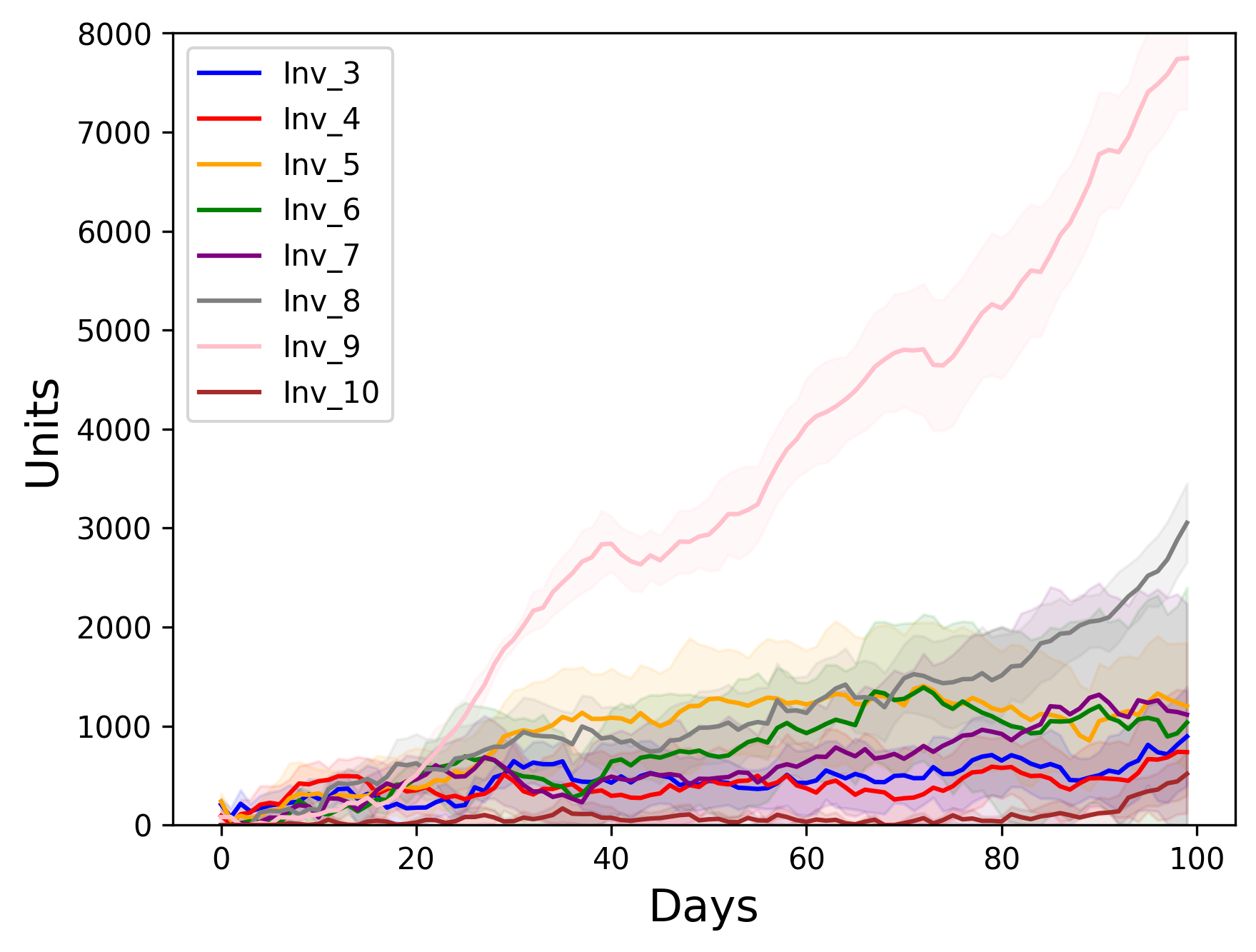}
        \caption{NSGA-II-Moderate SC}
        \label{fig:inv_nsga_moderate}
    \end{subfigure}

    \begin{subfigure}{0.32\textwidth}
        \includegraphics[width=\textwidth]{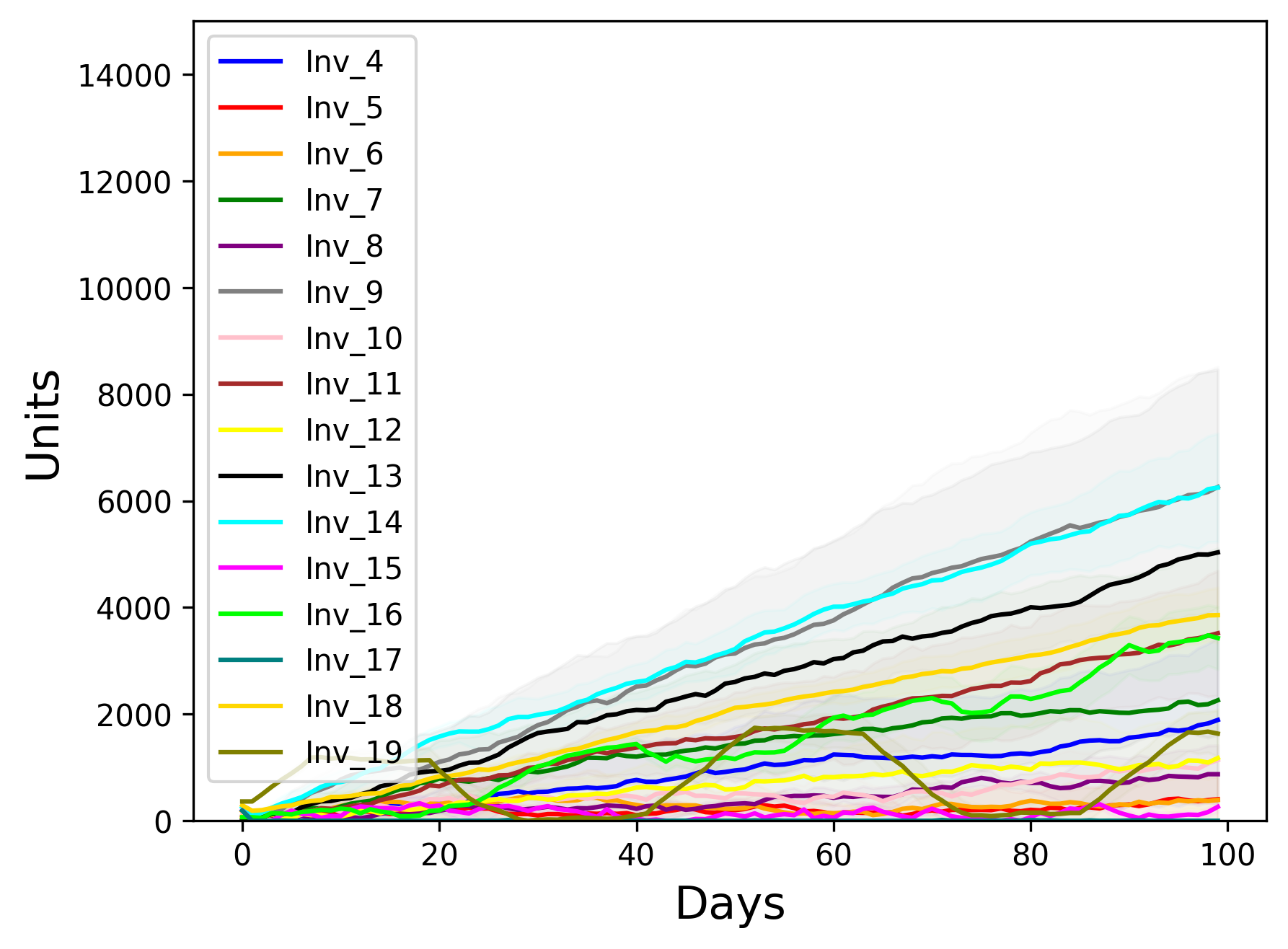}
        \caption{PPO-Complex SC}
        \label{fig:inv_ppo_complex}
    \end{subfigure}
    \hfill
    \begin{subfigure}{0.32\textwidth}
        \includegraphics[width=\textwidth]{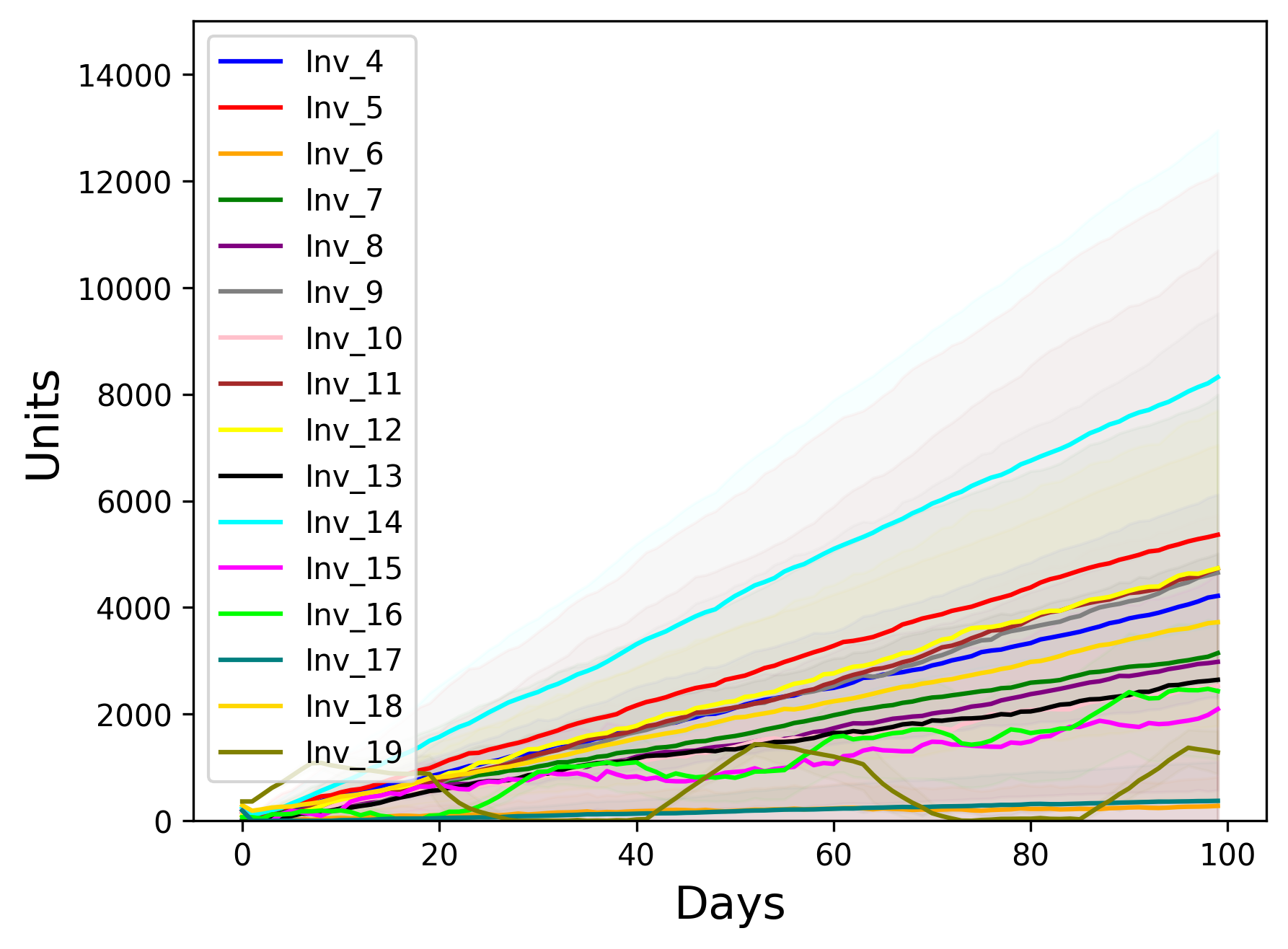}
        \caption{MORL/D-Complex SC}
        \label{fig:inv_morld_complex}
    \end{subfigure}
    \hfill
    \begin{subfigure}{0.32\textwidth}
        \includegraphics[width=\textwidth]{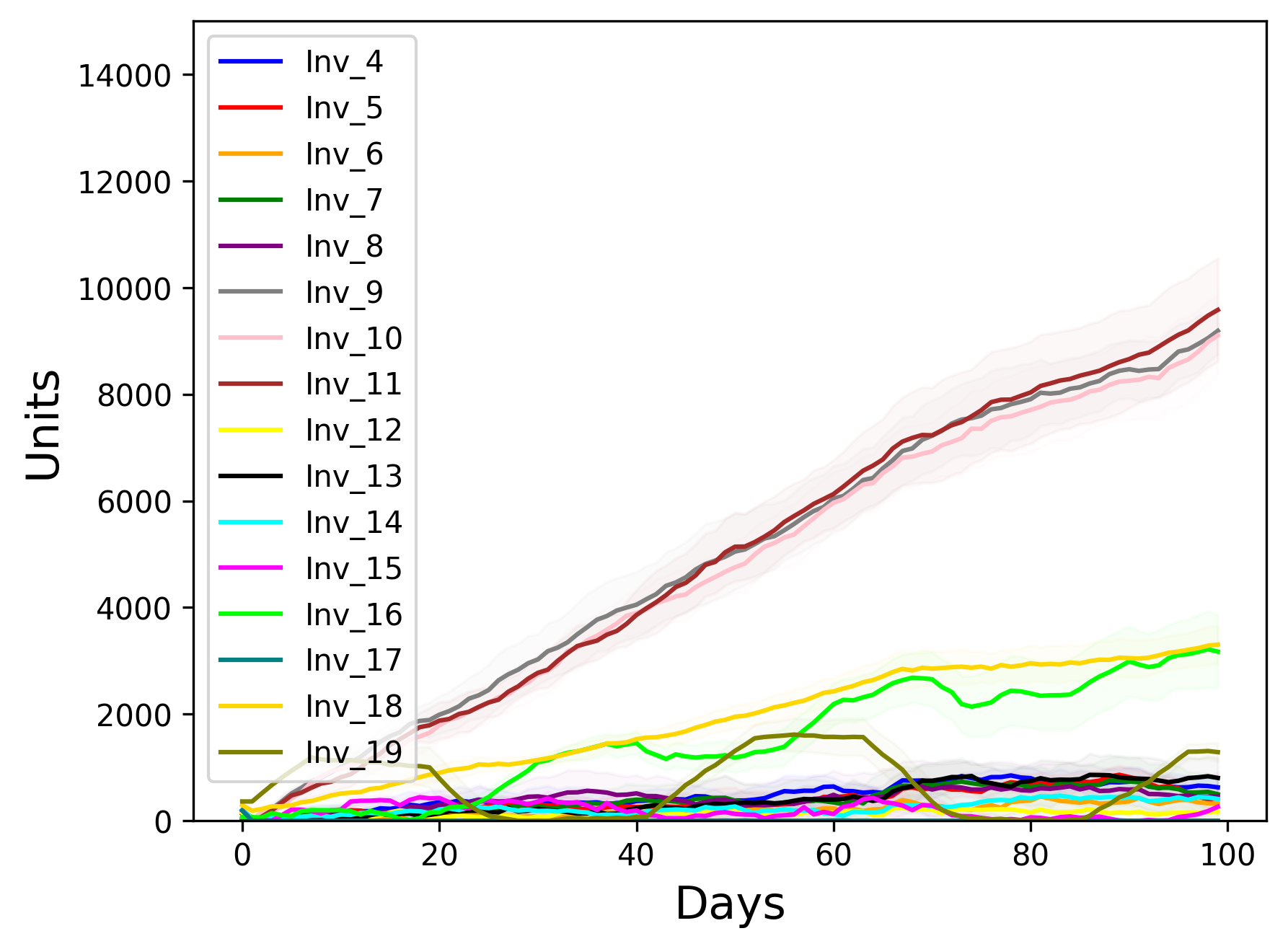}
        \caption{NSGA-II-Complex SC}
        \label{fig:inv_nsga_complex}
    \end{subfigure}
    \caption{Inventory build-up coincides with increasing complexity across all algorithms. In simpler environments, RL-based agents maintain relatively stable inventory levels (\ref{fig:inv_ppo_simple},~\ref{fig:inv_morld_simple}). The NSGA-II approach results in inventory imbalance and accumulation impacting multiple facilities within moderate and complex environments (\ref{fig:inv_nsga_moderate},~\ref{fig:inv_nsga_complex}).}
    \label{fig:inv}
\end{figure}

\begin{figure}[h]
    \centering
    \begin{subfigure}{0.32\textwidth}
        \includegraphics[width=\textwidth]{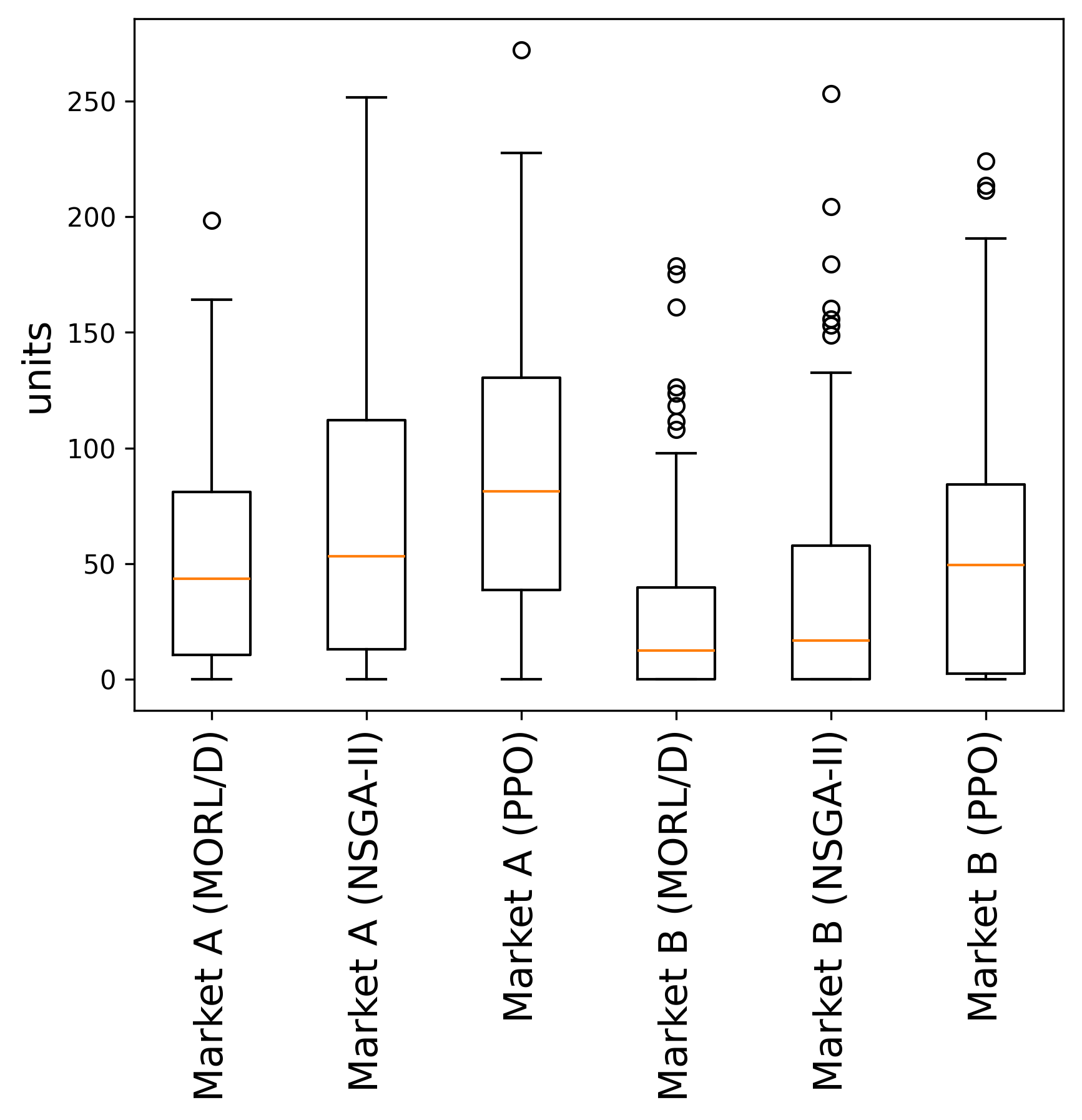}
        \caption{Simple SC}
        \label{fig:loss_simple}
    \end{subfigure}
    \begin{subfigure}{0.32\textwidth}
        \includegraphics[width=\textwidth]{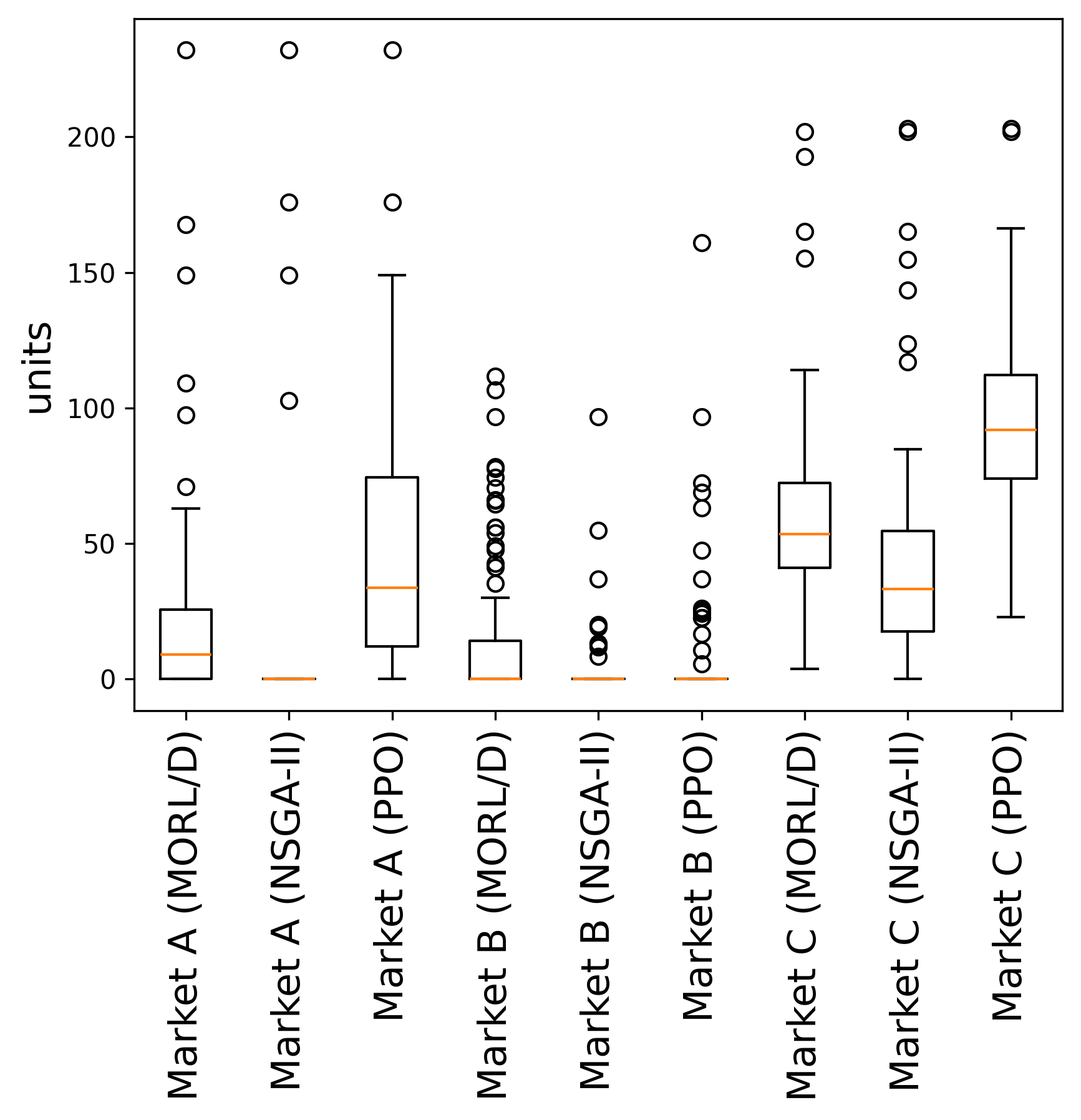}
        \caption{Moderate SC}
        \label{fig:loss_moderate}
    \end{subfigure}
    \begin{subfigure}{0.32\textwidth}
        \includegraphics[width=\textwidth]{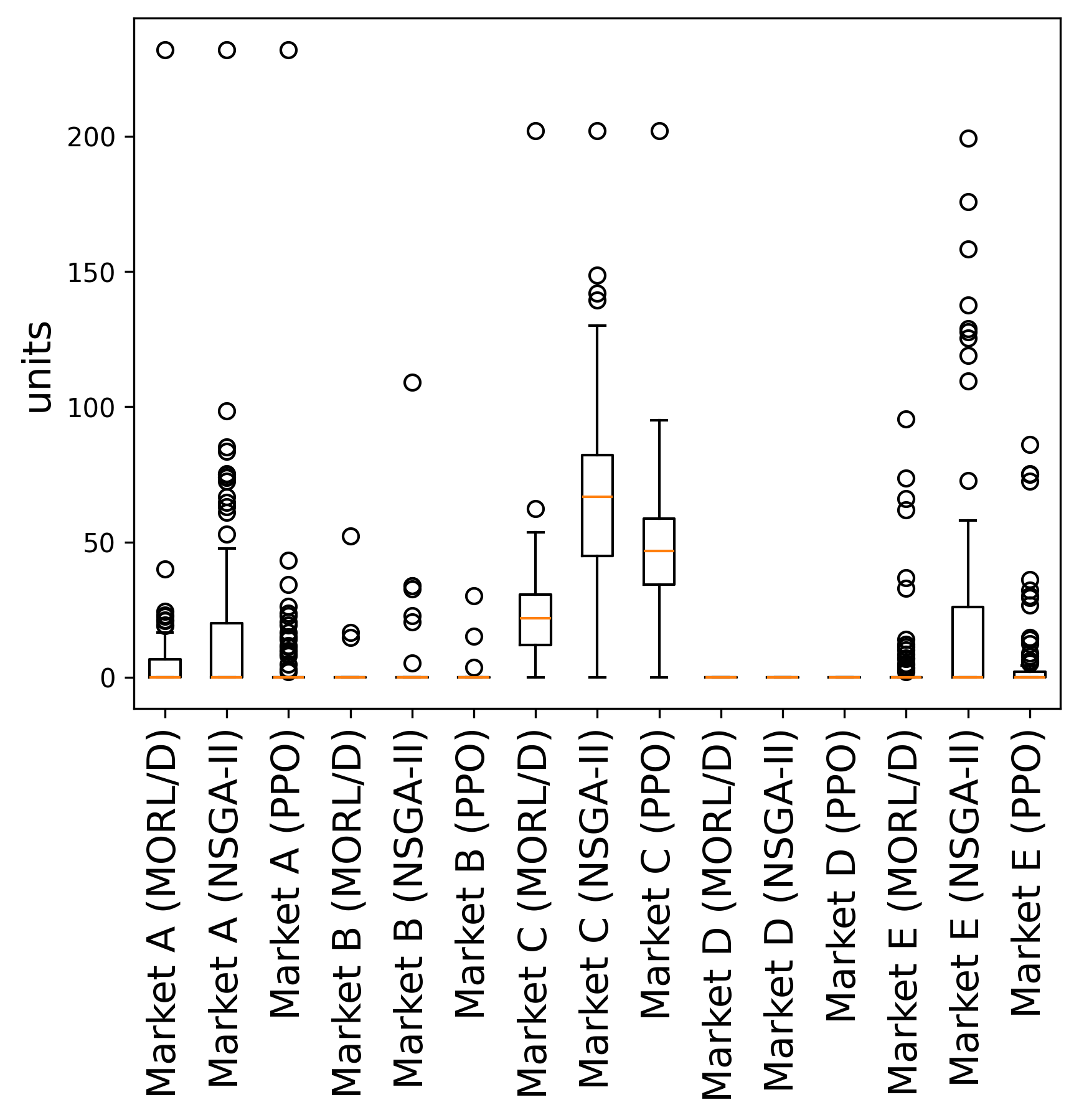}
        \caption{Complex SC}
        \label{fig:loss_complex}
    \end{subfigure}
    \caption{NSGA-II outperforms the other two in terms of demand fulfilment. MORL/D exhibits a moderate demand loss, while PPO experienced the most demand loss due to its lean inventory.}
    \label{fig:loss}
\end{figure}

To delve deeper into the operational characteristics, solution samples were chosen from five distinct runs, guided by the similarity of weights and/or objective values. Figure~\ref{fig:mfg} illustrates the differences in production between the three manufacturers pertaining to each algorithm. The PPO agent generally sustains low production volumes with slight variations over time (Figures~\ref{fig:mfg_ppo_simple},~\ref{fig:mfg_ppo_moderate},~\ref{fig:mfg_ppo_complex}). This pattern may elevate the potential for demand loss, as depicted in Figure~\ref{fig:loss}. Here, the PPO method incurs the maximum demand loss in both simple and moderate SC scenarios (Figures~\ref{fig:loss_simple},~\ref{fig:loss_moderate}), and the second highest in the complex SC scenario (Figure~\ref{fig:loss_complex}). Low production levels also lead to lean inventory in simple and moderate SC cases, resulting in inventory levels hovering around zero daily, except for a few days, as shown in Figures~\ref{fig:inv_ppo_simple} and~\ref{fig:inv_ppo_moderate}. In contrast, the complex SC problem yields moderate accumulation of inventories in several facilities (Figure~\ref{fig:inv_ppo_complex}).

In contrast, NSGA-II production is generally more volatile and reactive to changes in demand across the three SC problem configurations, as demonstrated in Figures~\ref{fig:mfg_nsga_simple},~\ref{fig:mfg_nsga_moderate},~\ref{fig:mfg_nsga_complex}. This results in daily inventory levels that are considerably less predictable, as shown in Figures~\ref{fig:inv_nsga_simple},~\ref{fig:inv_nsga_moderate},~\ref{fig:inv_nsga_complex}. An excessive inventory build-up is observed at retailer 9 for the moderate SC problem (Figure~\ref{fig:inv_nsga_moderate}), indicating insufficient market absorption. In the complex SC problem (Figure~\ref{fig:inv_nsga_complex}), inventory build-up is particularly significant compared to the RL-based approach in three facilities. Nonetheless, inventory levels demonstrate a significantly lower standard error than the other two methods. This reduced standard error corresponds to the highly concentrated solution points within the PF approximation set due to a narrow search area, as discussed in Section~\ref{sec:pf_solution_sets}. In addition to these concentrated solutions, the results appeared fairly consistent across separate runs, which, to some extent, is advantageous, as it indicates model robustness. However, in complex SC problems, where more diversity is anticipated, exceptionally low variability might be a consequence of strictly limited constraints because, in NSGA-II, constraints are treated as hard constraints. In addition, the high-dimensional problem exacerbates the exploration-exploitation imbalance by concentrating solutions in small feasible regions, reducing the algorithm's ability to explore the broader solution space. This issue is often addressed by employing a larger population. However, given current computational resource constraints, the chosen hyperparameter setup represents an optimal balance between feasibility and performance. Furthermore, despite its sensitivity to demand changes, the demand loss level remains moderate in simple and moderate SCs and is relatively high in the complex SC problem, as depicted in Figure~\ref{fig:loss}. This indicates the ineffectiveness of NSGA-II in handling complex high-dimensional problems.

Figures~\ref{fig:mfg_morld_simple},~\ref{fig:mfg_morld_moderate},~\ref{fig:mfg_morld_complex} reveal that with an increase in model complexity, the sensitivity of the MORL/D production quantity to demand is reduced, resulting in a fairly stable production quantity. This consistent MORL/D production level acts as a double-edged sword. Its reduced sensitivity to demand fluctuations can result in less precise manufacturing quantity decisions, compromising the efficiency of lean inventory practices and the ability to adapt to market shifts. However, in practical manufacturing, this stability can be advantageous in maintaining daily operational stability and simplifying the inventory management system. Figures~\ref{fig:inv_morld_simple} and~\ref{fig:inv_morld_moderate} demonstrate that MORL/D inventories remain more stable in less complex SC scenarios. As complexity increases, the agent begins to accumulate stocks. In the complex SC problem, inventories increase continuously, similar to other algorithms, albeit with a relatively larger standard error (Figure~\ref{fig:inv_morld_complex}). This significant error arises because of the variability in the weights between the selected solution samples, unlike in PPO, where these weights are consistent. In MORL/D, the agent assigns dynamic weights, resulting in discrepancies among the chosen solutions and contributing to varied outcomes. Without a penalty for unmet demand, the algorithms enjoy greater flexibility in determining production levels and product distribution. For example, by reducing production, the algorithm can decrease GHG emissions without incurring penalties for unmet demand. However, among all algorithms, MORL/D generally leads to a lower unmet demand, as evidenced by Figure~\ref{fig:loss}. This indicates that the MORL/D agent tends to be more cautious in meeting demand requirements. Although unmet demand is not directly penalised, negative inventory carries a penalty in the reward calculation, prompting the agent to avoid underproduction. A similar penalty is applied in PPO, yet it appears to be more effective in MORL/D.

\begin{table}[H]
    \centering
    \caption{The time required for experiments with each algorithm varies, with PPO and MORL/D reaching convergence at around $5\times10^5$ time steps, while NSGA-II requires 1000, 3000, and 5500 generations for simple, moderate, and complex problems, respectively. Compared to NSGA-II, PPO and MORL/D are more demanding in terms of computational resources. Furthermore, PPO necessitates several iterations to develop a set of solutions.}
    \label{tab:run_time}
    \begin{tabularx}{15cm}{@{}XXXX@{}}
    \toprule
    Algorithm & Time per Run & Time for a solution set & Number of solutions in a set\\
    \midrule
    PPO & 10-18 minutes & 210-378 minutes & 21 points\\
    MORL/D & 84-111 minutes & 84-111 minutes & 30 to 70 points.\\
    NSGA-II & 10-47 minutes & 10-47 minutes & 50 to 130 points\\
    \bottomrule
    \end{tabularx}
\end{table}

Table~\ref{tab:run_time} shows the computational time using current resources. The running time of the PPO increases proportionally to the number of solutions yielded in a PF approximation set. To illustrate, producing 50 solution points relative to the range of other methods requires approximately 8 hours and 18 minutes, up to 15 hours of runtime. In addition, additional effort will be required to specify the weight before starting the training. Meanwhile, the MORL/D and NSGA-II methods must only run once to result in a PF approximation set. RL methods are known to be computationally expensive due to the relatively long learning phase. However, they demonstrate a high success rate in balancing between exploration and exploitation, particularly in high-dimensional problems compared to metaheuristic methods~\citep{seyyedabbasi_hybrid_2021}.

\section{Conclusions}\label{sec:conclusion}
The study examines the efficiency and efficacy of implementing the MORL algorithm, specifically MORL/D, in tackling a multi-objective SC problem, outperforming PPO and NSGA-II. The explored environments include three instantiations of the SC problem, encompassing simple, moderate, and complex networks with various echelons, facilities, and markets subject to non-stationary demands. These problems involve action spaces with 8, 21, and 59 dimensions and observation spaces with dimensions of 20, 49, and 131, respectively, for simple, moderate, and complex SC scenarios.

The experiment results indicate that while PPO achieves the highest hypervolume across all SC setups, it provides the sparsest solution distribution and unstable EUM over time. The instability of PPO is potentially due to the gradient interference between conflicting objectives during training, the inherent policy update mechanism that is designed for single-objective problems, and the complexity of the problems that cause the relationship between states, rewards, and actions to become too intricate without an objective balancing mechanism. On the other hand, NSGA-II attains the lowest hypervolume and the highest AHD, with very low sparsity, indicating that its solutions are highly concentrated within a narrow search space, underscoring a disproportion between exploration and exploitation. In contrast, the MORL/D's hypervolume and EUM steadily rise over the time steps with decreasing sparsity, demonstrating the algorithm's stability across varying SC complexities. The MORL/D results converge to moderate values relative to the other algorithms. The SB mechanism improves hypervolume, EUM, and AHD while preserving density.

An in-depth analysis of operational behaviour reveals that MORL/D offers greater stability in production and inventory levels and experiences less demand loss compared to the other two algorithms. It indicates a lower sensitivity to demand fluctuations, which presents both advantages and drawbacks. On the positive side, it ensures production stability and simplifies operational and inventory management. In contrast, it may lead to inventory accumulation and reduced market adaptability. Although PPO achieves a higher hypervolume and lower AHD than MORL/D, the latter is noted to provide more balanced solutions and significantly reduce computation time, requiring only about one-sixth to one-eighth of that needed by the former. Future research in this field could concentrate on exploring more intricate SC settings. Moreover, employing sophisticated techniques like meta-learning may enhance model generalisation and significantly reduce the time required for training on new tasks.

% Use the style of numbering in square brackets.
% If nothing is used, default style will be taken.
%\begin{enumerate}[a)]
%\item 
%\item 
%\item 
%\end{enumerate}  

%\begin{itemize}
%\item 
%\item 
%\item 
%\end{itemize}  

%\begin{description}
%\item[]
%\item[] 
%\item[] 
%\end{description}  

% Uncomment and use as the case may be
%\begin{theorem} 
%\end{theorem}

% Uncomment and use as the case may be
%\begin{lemma} 
%\end{lemma}


\vspace{20pt}

\noindent\textbf{Acknowledgements}
The authors acknowledge the funding granted by the Economic and Social Research Council of UK Research and Innovation, the University of Manchester, and Peak AI, Ltd, with grant reference number ES/T002085/1. Research is carried out using computational resources from The University of Manchester and Peak AI, Ltd. We also acknowledge that the Peak AI, Ltd. Data Science team is involved in \textit{Messiah} development.
%\clearpage
%\newpage
\bibliographystyle{cas-model2-names-no-url}

\bibliography{references}



\end{document}